\documentclass{article} 

\usepackage{iclr2027_conference, times}

\usepackage{amsmath,amsfonts,bm}

\def\eqref#1{equation~\ref{#1}}
\def\Eqref#1{Equation~\ref{#1}}

\def\1{\bm{1}}

\DeclareMathAlphabet{\mathsfit}{\encodingdefault}{\sfdefault}{m}{sl}
\SetMathAlphabet{\mathsfit}{bold}{\encodingdefault}{\sfdefault}{bx}{n}

\newcommand{\Var}{\mathrm{Var}}

\newcommand{\Cov}{\mathrm{Cov}}

\usepackage{hyperref}
\usepackage{url}
\usepackage[utf8]{inputenc} 
\usepackage[T1]{fontenc} 
\usepackage{hyperref} 
\usepackage{url} 
\usepackage{booktabs} 
\usepackage{amsfonts} 
\usepackage{nicefrac} 
\usepackage{microtype} 
\usepackage{xcolor} 
\usepackage[most]{tcolorbox}

\usepackage{xcolor}
\usepackage[section]{placeins}
\usepackage{tabularx}
\usepackage{booktabs}
\usepackage{makecell}
\usepackage{wrapfig}
\usepackage{graphicx}
\usepackage{tikz} 
\usepackage{subcaption} 
\usepackage{colortbl}
\usepackage{minted}
\usepackage{circledsteps}
\usepackage{anyfontsize}
\usepackage{verbatim}
\RecustomVerbatimCommand{\verbatiminput}{VerbatimInput}%
{
 breaklines=true,
 breakanywhere=true,
 breakautoindent=false,
 breaksymbol={},
 fontsize=\fontsize{4pt}{5pt}\selectfont,
 formatcom=\fontsize{4pt}{5pt}\selectfont
}
\usepackage{multirow}
\usepackage{fontawesome5}
\usepackage{pifont} 
\usepackage{algorithm}
\usepackage{algorithmic}
\usepackage{epigraph}
\usepackage{xfp}
\usepackage{xstring}
\usepackage{enumitem}
\usepackage{adjustbox}
\usepackage{svg}
\usepackage{amsthm}
\newtheorem{proposition}{Proposition}[section]
\newtheorem{corollary}[proposition]{Corollary}

\newif\ifarxivversion      

\arxivversiontrue         

\ifarxivversion
    \iclrfinalcopy   
\fi

\makeatletter
\iffalse
\newcommand{\dimpp}[1]{\textcolor{blue}{[DP: #1]}}
\newcommand{\onur}[1]{\textcolor{magenta}{[OK: #1]}}
\newcommand{\desmond}[1]{\textcolor{orange}{[DE: #1]}}
\newcommand{\todo}[1]{\textcolor{red}{[TODO: #1]}}
\else
\newcommand{\todo}[1]{\@bsphack\@esphack}
\newcommand{\dimpp}[1]{\@bsphack\@esphack}
\newcommand{\onur}[1]{\@bsphack\@esphack}
\newcommand{\desmond}[1]{\@bsphack\@esphack}
\fi
\makeatother

\newcommand{\mypar}[1]{\vspace{0mm}
\noindent
\textbf{#1}}

\definecolor{myblue}{HTML}{3B6BA5}
\definecolor{mygreen}{HTML}{5B9759}
\definecolor{myorange}{HTML}{B1752D}

\iftrue
\newtcolorbox{takeaway}{ colback=gray!10, 
colframe=black, 
arc=0mm, 
boxrule=1pt, 
left=3pt, right=3pt, top=3pt, bottom=3pt, 
}
\else
\usepackage{environ}
\NewEnviron{takeaway}{} 
\fi
\newtcolorbox{prompt}[1][Prompt]{
title={#1}, enhanced, colback=myblue!5!white, 
colframe=myblue!90!white, 
colbacktitle=myblue!90!white,
coltitle=white, 
fonttitle=\bfseries, 
arc=0mm, 
boxrule=1pt, 
left=6pt, right=6pt, top=6pt, bottom=6pt, 
toptitle=4pt, 
bottomtitle=4pt, 
lefttitle=6pt, 
righttitle=6pt, 
fonttitle=\bfseries\footnotesize,
fontupper=\footnotesize,
float=tb,            
}

\newcommand{\promptcell}[4]{%
\begin{minipage}[t]{#1}
\begin{tcolorbox}[
    enhanced,
    width=\linewidth,
    height=#2,
    colback=white,
    boxrule=0pt,
    colframe=white,
    arc=1.5mm,
    outer arc=1.5mm,
    left=0mm,
    right=0mm,
    top=-4mm,
    bottom=1mm,
    title={#3},
    colbacktitle=black!7,
    coltitle=black,
    fonttitle=\bfseries\footnotesize,
    halign title=center,
    boxed title style={
        boxrule=0pt,
        arc=1.5mm,
        outer arc=1.5mm,
        height=5mm
    },
    attach boxed title to top={
        xshift=0mm,
        yshift=0mm
    },
    before upper={
        \scriptsize
        \raggedright
    }
]
#4
\end{tcolorbox}
\end{minipage}%
}

\newcommand{\deltanum}[1]{%
  \begingroup
  \tiny
  \IfSubStr{#1}{X.X}{%
    #1%
  }{%
    \IfStrEq{#1}{-.-}{%
      #1%
    }{%
      \ifdim\fpeval{#1}pt>0pt
        \textcolor{green!60!black}{#1}%
      \else
        \ifdim\fpeval{#1}pt<0pt
          \textcolor{red}{#1}%
        \else
          \textcolor{gray}{#1}%
        \fi
      \fi
    }%
  }%
  \endgroup
}
\newcommand{\normaldeltanum}[1]{%
  \begingroup
  \let\tiny\small
  \deltanum{#1}
  \endgroup
}

\newcommand{\sampleimage}[1]{%
    \adjustbox{max width=\linewidth, max height=2.5cm, center}{%
        \includegraphics{#1}%
    }%
}

\title{SIMIT: \underline{S}elf-\underline{I}mproving Vision-Language\\\underline{M}odels via \underline{I}magination at \underline{T}est-Time
\ifarxivversion
\vspace{-0.3em}
\fi
}
\author{Mehmet Onurcan Kaya$^{1,2}$ \quad Desmond Elliott$^{3,2}$ \quad Dim P. Papadopoulos$^{1,2}$\\
$^{1}$ Technical University of Denmark \quad $^{2}$ Pioneer Center for AI \quad $^{3}$
University of Copenhagen
\\
\texttt{monka@dtu.dk}, \quad \texttt{de@di.ku.dk}, \quad \texttt{dimp@dtu.dk}
\\[6pt]
\faGlobe\ \textbf{Project Page:} \href{https://monurcan.github.io/simit}{https://monurcan.github.io/simit}
\ifarxivversion
\vspace{-1.4em}
\fi
}

\begin{document}
  \maketitle
  
\begin{abstract}
We introduce \textbf{SIMIT}, a test-time self-improvement framework in which a vision-language model creates its own query-specific training data before answering an unlabeled test query. 
Our goal is 
to improve performance on a given target 
test set without annotations or external models.
To pursue this goal, some test-time methods improve answers through longer reasoning or repeated sampling but leave the model unchanged. Others update the model using rewards based on majority agreement among multiple candidate answers or the model's own judgments of correctness, but such rewards can reinforce existing mistakes.
SIMIT instead imagines similar problems whose solutions it controls: a model need not correctly solve the target query to construct useful practice examples. It synthesizes diverse (question, answer, image) triplets by specifying each answer before realizing its image.
To support diverse multimodal tasks, SIMIT combines native image generation with an extensible skill library for structured visuals, including documents, charts, and diagrams. Within this pipeline, multi-stage verification checks that images support their assigned answers, while difficulty filtering retains informative examples. Adaptive budgeting further targets this practice by allocating more samples to queries the model is less confident in answering.
The resulting data can improve the model 
in-context (\textbf{SIMIT-ICL}) or in-weight (\textbf{SIMIT-FT}).
Across 17 diverse benchmarks, SIMIT outperforms
existing self-improvement methods. 
Its best configuration achieves a
\textbf{+7.20\%} mean relative gain over 
BAGEL-7B, while training-free SIMIT-ICL achieves \textbf{+6.98\%}. \dimpp{ok, great results, but what's more? there should be something more we want to highlight. e.g., SIMIT vs LFRL or synthetic vs real}
\dimpp{the abstract misses the motivation part. Why synthesizing data is a good idea? (query-speficy data are extremely useful, but hard to obtain in practice?)}
\end{abstract} 
  \section{Introduction}
\label{sec:intro}


\begin{figure}[t]
    \vspace{-1.5ex}
    \centering
    \includegraphics[width=1.0\textwidth]{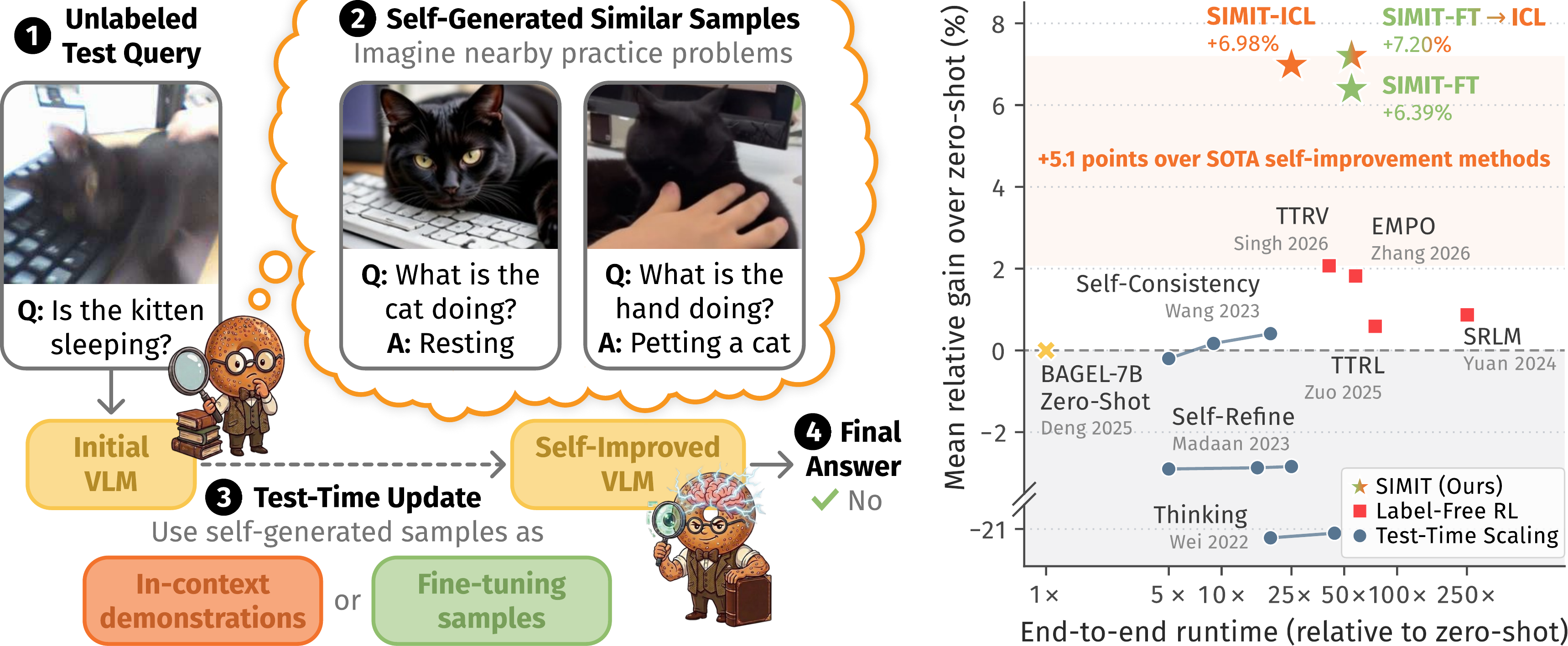}
    \caption{
    \textbf{(Left)} SIMIT. Given an unlabeled test query \textbf{\Circled{1}}, the VLM imagines similar samples \textbf{\Circled{2}}, uses them to improve itself via in-context learning or by fine-tuning \textbf{\Circled{3}}, and answers the original test query with the self-improved model \textbf{\Circled{4}}. No labels or external models are used. \textbf{(Right)} Mean relative gain versus runtime across 17 benchmarks.
    } 
    \label{fig:intro_fig}
\end{figure}

The success of vision-language models (VLMs) depends on expensive human-labeled data for general-purpose pretraining and task-focused post-training~\citep{dai2023instructblip,sun2024aligning}.
One issue with this training strategy is that models cannot anticipate every task, domain, or failure mode encountered in real-world use after deployment. 
In this paper, we study the problem of \emph{test-time self-improvement}, where a VLM can improve its performance on an \textit{unlabeled} dataset, without access to an external teacher or additional human-annotated data. 
Such a capability would enable a VLM to specialize its abilities on demand, using the input data itself to determine what it needs to learn.
Test-time scaling offers a partial solution to this problem. 
Longer thinking, 
repeated sampling, and majority voting can improve a frozen model’s predictions \citep{wei2022chainofthought,wang2022selfconsistency} without providing persistent parameter updates for subsequent queries. 
An alternative approach is test-time training, which updates model weights using pseudo-labels or rewards derived from confidence, consensus, or self-rewarding \citep{huang2023selfimprove,zuo2025ttrl,singh2026ttrv,yuan2024selfrewarding}. 
This enables cross-query transfer, but supervision comes from the response behavior being corrected, which can reinforce systematic errors \citep{wei2026mmupt,he2026howfar}. 
Furthermore, open-ended multimodal tasks present an additional problem: semantically equivalent answers may differ in wording, complicating voting-based solutions, while verification requires checking fine-grained visual evidence.
Retrieval-based methods, such as RICES~\citep{alayrac2022flamingo} and TTT-NN~\citep{hardt2024tttnn}, offer another source of supervision by adapting from human-annotated examples. 
However, they require access to a labeled retrieval set that is not available in our label-free problem setting.
Taken together, these limitations motivate the need for a source of supervision that requires neither correctly answering the target test query nor access to labeled data.

In this paper, we tackle the problem of test-time self-improvement by synthesizing input-relevant supervision samples using the underlying model itself 
(Fig.~\ref{fig:intro_fig}). 
Our approach aims to synthesize useful, accurate, and diverse 
question--answer--image triplets, conditioned on a given test query. 
SIMIT achieves these goals using unified multimodal models (UMMs), which have image understanding and generation capabilities that enable \emph{the same model} to synthesize and learn from its own multimodal samples without target labels or external models. 
But existing UMMs can only generate natural images and struggle with structured visuals, e.g., documents, charts, and diagrams~\citep{wei2026ggbench}.
Addressing this weakness, SIMIT can synthesize a variety of image types with an extensible library of structured rendering skills.
In order to minimize unnecessary computation, the synthesis budget is allocated adaptively, based on the confidence of the UMM in its zero-shot response.
The generated samples are checked for internal consistency and filtered based on their expected difficulty.

The synthesized samples can be used in two 
modes for self-improvement: 
\textbf{SIMIT-FT} uses them for finetuning, while 
\textbf{SIMIT-ICL} uses them as in-context demonstrations without model parameter access.
With access to a single test query, we use SIMIT-ICL or SIMIT-FT (Single-Query). 
But with
access to
multiple test
queries, SIMIT-FT (Test-Set) allows the performance gained for one query to benefit others. 
\textbf{SIMIT-FT$\rightarrow$ICL} combines both 
using the finetuned model with few-shot prompting. 
In summary, 
our main
contributions 
are as follows:
\begin{itemize}[leftmargin=*]
    \item \textbf{Imagination-based self-supervision.} The same model creates its own query-conditioned 
    training samples by imagining similar problems, without predicting labels for the original test inputs.

    \item \textbf{General multimodal realization.} We combine the model's native image
    generation with structured
    rendering skills, multi-stage verification,
    difficulty filtering, and confidence-adaptive budget allocation to build 
    query-conditioned synthetic data 
    spanning natural and structured visual tasks. 

    \item \textbf{Flexible adaptation and broad evaluation.}
    We can use the self-generated samples for in-context learning and single-query or test-set finetuning.
    Across 17 benchmarks with BAGEL-7B \citep{deng2025bagel}, mean relative gains over zero-shot are $6.98\%$ for SIMIT-ICL, $1.69\%$ for single-query SIMIT-FT, $6.39\%$ for test-set SIMIT-FT, and $7.20\%$ for SIMIT-FT$\rightarrow$ICL, outperforming prior work.
     Our synthetic data can even outperform human-annotated data on some benchmarks.
    \dimpp{not clear what single-query or the other variations are} 
\end{itemize} 
  \section{Related Work}
\label{sec:relatedwork}

\dimpp{it feeels very thin. I don't think you need to extend the text but try to include more papers by grouping them. e.g., why don't you mention your own paper?} 
\dimpp{Also, what I miss is what is really on LLMs only and what is on VLMs. One important aspect of this work is that it's multimodal and as we discussed applying something from text-only to multimodal setting is not trivial.}

\mypar{Test-Time Scaling (TTS)} improves answers by spending more
computation when solving a query. 
\emph{Sequential} methods think step-by-step (chain-of-thought, \citealp{wei2022chainofthought}) or
repeatedly critique and revise an answer (Self-Refine, \citealp{madaan2023selfrefine}).
\emph{Parallel} methods generate multiple candidate solutions, then
choose the majority-voted answer (Self-Consistency, \citealp{wang2022selfconsistency}) or score the candidates
and select the best one (best-of-$K$)
\citep{cobbe2021training,snell2025scaling}.
However, approaches that score using external 
reward models fall outside our self-contained problem setting. 
But self-contained aggregation can 
reinforce the model's 
biases 
and lack meaningful consensus on open-ended tasks such as captioning.
Also, conventional TTS leaves model parameters unchanged. Computation spent on one query does not 
benefit subsequent queries. SIMIT instead converts test-time compute into query-conditioned
practice 
data, enabling in-context or in-weight adaptation. 

\mypar{Label-Free Reinforcement Learning (LFRL)}
updates the model parameters using intrinsic rewards. TTRL rewards majority agreement
\citep{zuo2025ttrl}, TTRV uses response frequencies and entropy regularization
\citep{singh2026ttrv}, and EMPO uses semantic clustering
\citep{zhang2026empo}, which may require an external model. SRLM instead uses
self-judging rewards \citep{yuan2024selfrewarding}, limited by evaluation
capability and vulnerable to reward hacking \citep{gao2023scaling}.
Cross-query transfer and online reward updates can improve pseudo-labels beyond
the initial majority vote \citep{zuo2025ttrl}, but isolated single query adaptation lacks
this transfer. MM-UPT additionally synthesizes questions around existing images
while retaining majority-derived supervision \citep{wei2026mmupt}. SIMIT
instead specifies answers before constructing and verifying new visual inputs. 

\mypar{Test-Time Training (TTT)}
adapts parameters using either retrieved neighbors \citep{hardt2024tttnn},
supervision from augmented demonstrations or test inputs 
\citep{akyurek2025tttarcagi, kaya2026efficient}, or likelihood objectives on unlabeled inputs
\citep{hu2025tlm}. Closest to SIMIT, TT-SI \citep{acikgoz2026ttsi} finetunes on self-generated
tool-use examples, validating calls and selecting function labels by majority
agreement. SIMIT constructs complete multimodal
examples without a labeled retrieval corpus or supplied demonstrations.
Extended related work can be found in
Appendix~\ref{sec:appendix_extendedrelatedwork}. 
  \section{Methods}
\label{sec:methods}

\subsection{Problem Formulation: Test-Time Self-Improvement}



Given a vision-language model \(\pi_\theta\) and an \textit{unlabeled target test set}
\(\mathcal{Q}=\{(I_i,q_i)\}_{i=1}^N\), we aim to improve performance on the target distribution without its ground-truth answers. We study a \textit{self-contained} setting: $\pi_\theta$ is the only learned model available at test time, with no auxiliary teacher, verifier, retriever, or generator. Deterministic tools and renderers may execute model-produced specifications, but all semantic decisions and supervision originate from $\pi_\theta$.
This problem setting measures 
whether the model can reorganize its existing latent capabilities for self-improvement. 

The setting is \textit{label-free} and \textit{target-conditioned}. Unlike conventional SFT, RLVR, and RLHF, \dimpp{do you define these?} supervision comes from the target model $\pi_\theta$ rather than ground-truth targets, externally specified verifiable rewards, or human preferences. Moreover, unlike untargeted zero-data self-improvement 
\citep{he2026visplay,huang2026rzero,zhao2026absolute}, observed target queries determine what the model should practice in our problem setting, focusing self-improvement on a specific target.

\begin{figure}[t]
    \vspace{-1.5ex}
    \centering
    \includegraphics[width=1.0\textwidth]{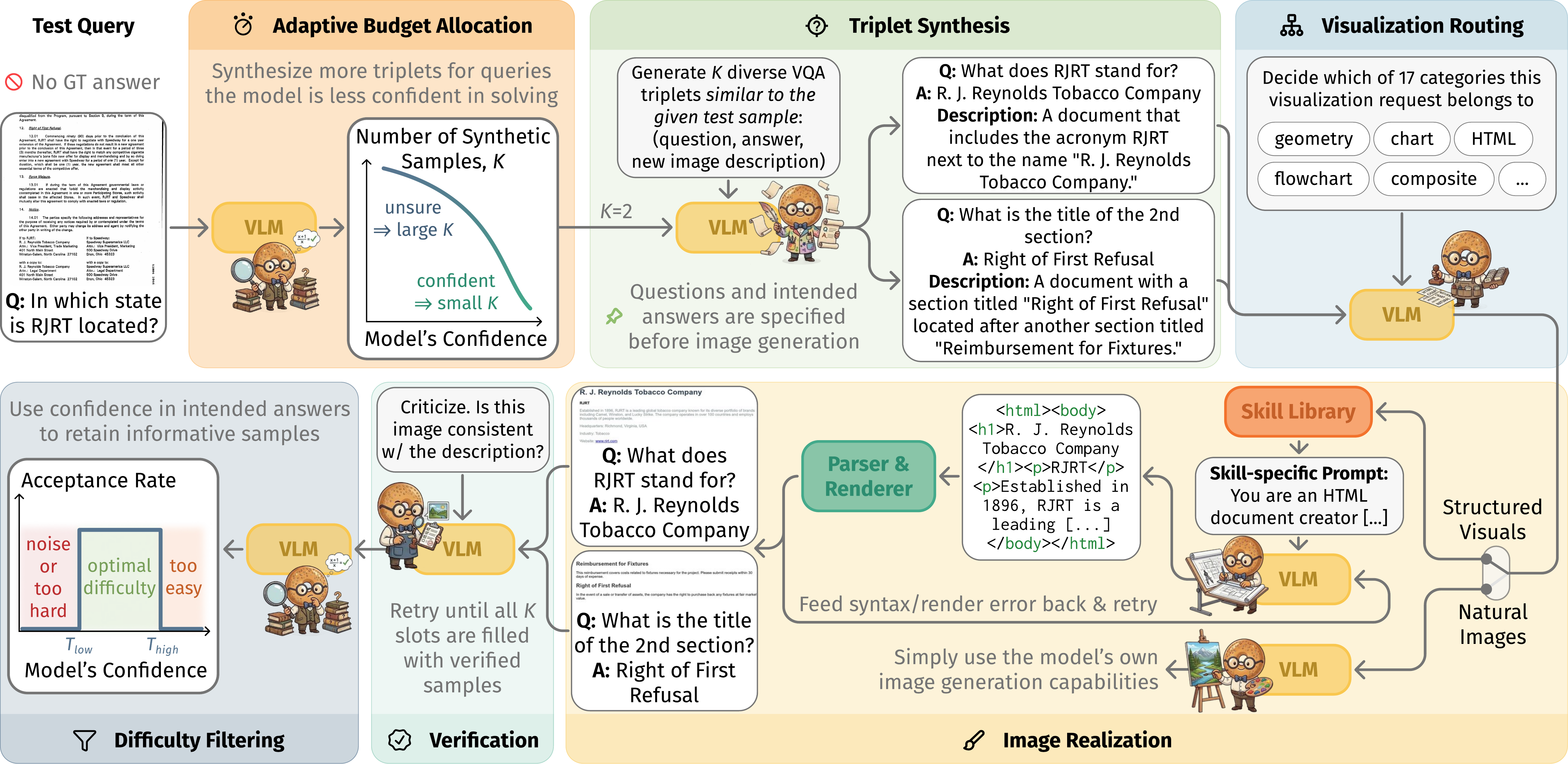}
    \caption{SIMIT's synthetic data generation pipeline.
    Given an unlabeled test query, the model allocates a synthesis budget
    and proposes diverse question-answer pairs with corresponding image
    descriptions. Answers are fixed before images are realized through
    native generation or rendering skills. Verification and difficulty
    filtering retain accurate and informative examples. 
    \textbf{The same model serves as the synthesizer, router, critic, and solver.}
    This figure shows only the data synthesis stage. 
    But the model then self-improves with the final samples 
    via in-context learning or finetuning. 
    }
    \label{fig:detailed_pipeline}
    \vspace{-0.7em}
\end{figure}

\subsection{Self-Supervision by Imagining Similar Problems}

SIMIT self-improves on an unlabeled test query 
by constructing similar
multimodal samples and using them for adaptation. 
Its central idea is 
\emph{question-answer-guided image generation}: 
the model first proposes new questions and their intended answers, then generates images to support those answers.
This avoids directly reusing incorrect predictions on the test query as supervision.
The remaining challenge is to make 
the resulting samples 
valid and useful for the model's own adaptation.

Our synthesis pipeline (Fig.~\ref{fig:detailed_pipeline}) is designed to produce \emph{accurate}, \emph{diverse}, and \emph{informative} examples (see App.~\ref{sec:appendix_synthesis_principles} for a formal argument).
The process begins with \emph{Adaptive Budget Allocation}, which selects a synthesis budget $K$ from the model's initial confidence on the test query.
\emph{Triplet Synthesis} then generates \(K\) diverse triplets \(\{(\tilde q_k,\tilde a_k,\tilde d_k)\}_{k=1}^K\), a question, intended answer, and image description, conditioned on the test query $(I_i,q_i)$.
\emph{Visualization Routing} selects the model's own native image generation or a suitable structured rendering skill for each description $\tilde d_k$. \emph{Image Realization} 
generates the corresponding image $\tilde I_k$.
\emph{Verification} checks whether each image supports its question-answer pair, triggering repair or regeneration when needed.
Finally, \emph{Difficulty Filtering} uses the model's confidence in the intended answers $\tilde a_k$ to discard samples that are too easy or difficult.
\onur{I rewrote this paragraph, Dim}

\mypar{Adaptive Budget Allocation.}
Different test queries require different amounts of practice.
We select a query-specific synthesis budget $K$ 
based on the model’s confidence in its zero-shot prediction.
Less confident queries consequently receive more samples.
A log-odds evidence criterion selects the smallest budget predicted
to reach a desired correctness probability, subject to a given maximum budget. 
Confidence is the mean token probability of the model's zero-shot answer, and $K$ decreases monotonically with confidence. 
Appendix~\ref{sec:appendix_adaptive_K} derives our allocation rule and optimal functional form.


\mypar{Triplet Synthesis.}
Conditioned on the test image and question $(I_i,q_i)$, the model proposes
$K$ 
synthetic triplets \((\tilde q_k,\tilde a_k,\tilde d_k)\): a question, its intended answer, and a textual description of an image
that would support that answer.
Diversity-encouraging prompting promotes variation 
while preserving the target task and its answer format.
Each answer is committed before its image is generated, turning
label acquisition into conditional construction.
Generating both questions and images expands 
practice beyond a fixed test image. 
We use a UMM to generate images without an external model.

\mypar{Visualization Routing.}
The next step is to turn each textual description $\tilde d_k$ into an image $\tilde I_k$. 
But we cannot simply use the model's native image generation capabilities,
as existing UMMs are incapable of generating structured images~\citep{wei2026ggbench}.
They struggle with the precise text,
quantities, and spatial relations required by structured visuals (Fig.~\ref{fig:diagram_library_qualitative}).
Thus, we prompt the same model to identify each description's visualization category. Based on this classification, we route natural-image requests to native generation and structured-image requests to category-specific rendering skills.



\begin{figure}[t]
    \vspace{-1.5ex}
    \centering
    \includegraphics[width=1.0\textwidth]{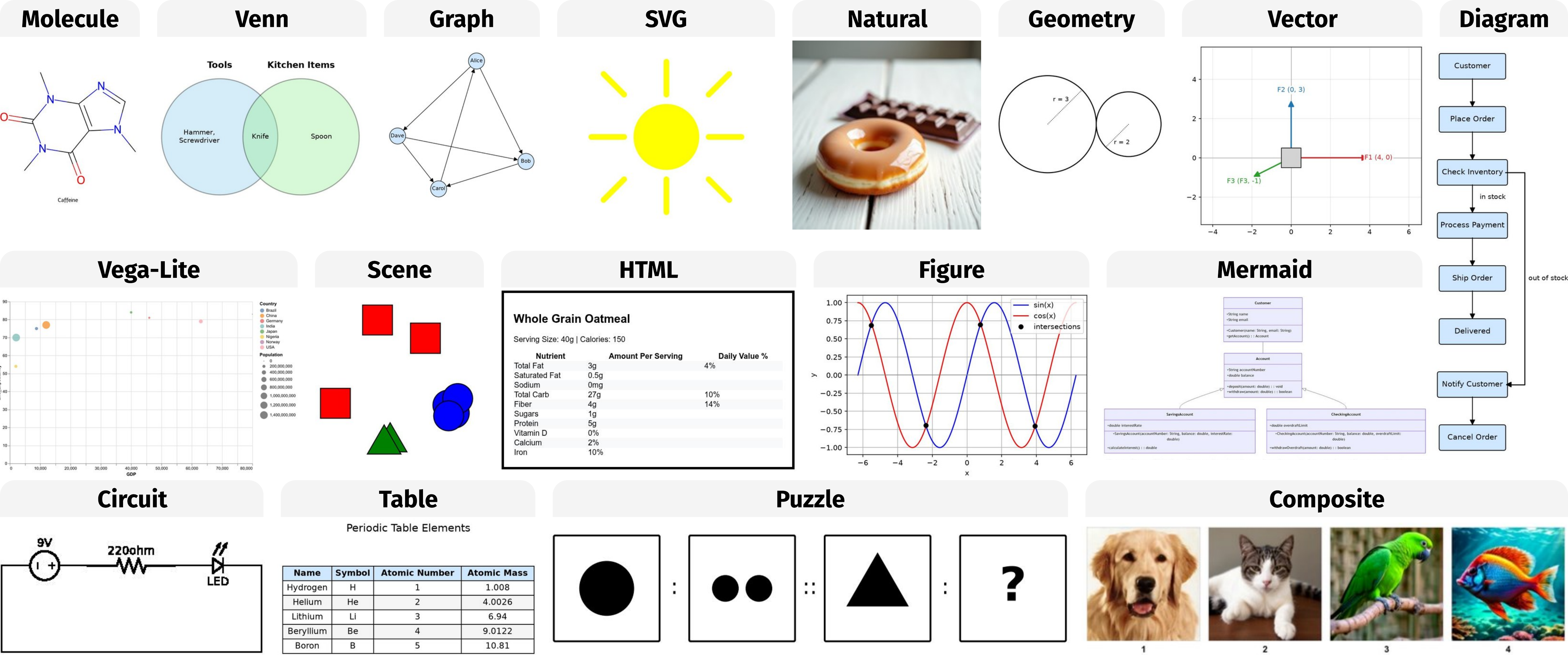}
    \vspace{-1.13em}
    \caption{Representative visuals synthesized by SIMIT.
    Combining native image generation with rendering skills enables
    BAGEL-7B to produce natural images, structured visuals, and
    composite layouts within a single pipeline.
    Corresponding generation prompts are in
    Appendix~\ref{sec:appendix_prompts_diagram_categories}.}
    \label{fig:diagram_categories}
    \vspace{-.45em}
\end{figure}

\mypar{Image Realization.}
For structured visuals, directly generating 
Python/\LaTeX{}/HTML 
code demands programming, layout, and geometric reasoning capabilities 
that current UMMs may not reliably possess.
Whereas prior work develops structured generation capabilities through finetuning or reinforcement learning \citep{chen2026figure, li2025zebra}, our \emph{skill library} uses modular instructions, easy-to-use interfaces, and reusable rendering resources without training \citep{xu2026agentskills}.
Each skill guides the model to produce a compact specification, such as geometric primitives or graph nodes and edges. A deterministic backend parses it and computes supported quantities, geometry, and layout, enabling syntax checks and explicit error feedback.
For example, to draw a regular pentagon inscribed in a circle, the model only specifies the center, radius, and number of sides, and the renderer computes vertices and requested angle labels consistently. Skills can also compose, e.g., the HTML skill can embed natively generated images. The skill library covers common benchmark tasks. Fig.~\ref{fig:diagram_categories} shows representative outputs. It is easily extensible with additional specifications and backends. Appendix~\ref{sec:appendix_implementation_details} provides implementation details, such as prompts, interfaces, rendering backends, etc.

\mypar{Verification.} 
The model acts as a critic, checking consistency between the description $\tilde d_k$ and image $\tilde I_k$. 
Our synthesis-based self-supervision avoids deriving labels from the target prediction, but imperfect image realization can still leave the intended answer unsupported. Thus, validation spans the synthesis pipeline: parsers check syntax, renderers enforce supported constraints, the critic checks consistency. 
Failures trigger repair or regeneration with explicit error feedback and bounded retries.




\textbf{Difficulty Filtering.}
Synthetic examples should be valid, but also  challenging for the model.
We estimate their difficulty from the model's confidence in the intended answer $\tilde a_k$ and retain examples with confidence within $[T_{\mathrm{low}},T_{\mathrm{high}}]$. 
$T_{\mathrm{low}}$ rejects examples that may contain residual
label noise or exceed the model's capabilities. 
$T_{\mathrm{high}}$ removes examples with little learning value that can be harmful rather than merely redundant, e.g., they consume context, displace useful demonstrations, and introduce interference. App.~\ref{sec:appendix_confidence_filter} derives our band-pass rule from an expected utility formulation.


\subsection{Test-Time Update: Use Cases of the Self-Generated Synthetic Data}


The self-generated data can be used for in-context learning (SIMIT-ICL),
fine-tuning (SIMIT-FT), or both. We study the following use cases to reflect practical scenarios
with different constraints.

\mypar{SIMIT-ICL} uses synthesized samples directly as few-shot demonstrations \citep{brown2020fewshot}. Self-generated demonstrations have been explored for text-only LLMs through SG-ICL, SELF-ICL, and analogical prompting \citep{kim2022sgicl, chen2023selficl, yasunaga2024large}. SIMIT additionally uses synthesized visual inputs for the generated question–answer pairs and verifies their consistency.
SIMIT-ICL is a prompting technique requiring only inference access.
It is suitable for
API-accessible models and settings where
fine-tuning is infeasible.

\mypar{SIMIT-FT (Single-Query, SQ)} treats each test query as a separate
$N=1$ adaptation problem.
We finetune the base model on query-specific synthetic examples,
answer the query, and reset before the next query.
This offers a complementary axis of test-time scaling:
spending computation on query-specific learning
instead of conventional parallel or sequential scaling.
It suits requests that prioritize accuracy over latency
and applications that restrict information retention:
synthetic data can be discarded after answering,
and parameter updates do not persist across queries.

\mypar{SIMIT-FT (Test-Set, TS)}
combines synthetic examples from all test queries 
to finetune the model before evaluating it on the target test set.
This allows the supervision generated for one query to benefit
others. It has many applications, such as automatic labeling
and LLM personalization.

\mypar{SIMIT-FT$\to$ICL} 
first finetunes on the self-generated 
data
and then supplies them as 
in-context demos 
to the finetuned model. 
It combines 
weight adaptation with query-specific contextual guidance.
  \section{Experiments}
\label{sec:experiments}

Our main experiments use BAGEL-7B
\citep{deng2025bagel}, which has unified understanding and generation capabilities that 
allow the same model 
to learn from its own imagined examples. 
We evaluate whether self-generated 
samples improve performance across 17 diverse multimodal benchmarks from LMMs-Eval-Lite~\citep{zhang2025lmmseval}, 
spanning five domains: Image Understanding \& Captioning (Flickr30k, NoCaps, TextCaps, RefCOCO, COCO), Visual
Question Answering (GQA, OK-VQA, VizWiz, VQA-v2, TextVQA), Math \& Science (AI2D), Document \& Infographic Understanding (ChartQA, DocVQA, InfoVQA), 
and Multi-disciplinary Reasoning (MM-Bench (CN+EN), SEED-Bench). 
See Appendix~\ref{sec:appendix_dataset_stats} for more details on evaluation metrics.
We report arithmetic mean scores within each domain, with domain-level relative improvements computed from these averages. To summarize performance across heterogeneous metrics in a single scalar, we also report ``Av. Gain'', the mean benchmark-level relative improvement over the zero-shot baseline across all 17 benchmarks.
In Tab.~\ref{tab:ablation_components} and Sec.~\ref{sec:label_noise}, we also report supervision quality as scored by an external evaluator.

\FloatBarrier
\subsection{Comparison with State-of-the-Art Test-Time Methods}

\begin{table}[t!]
    \centering
    \small
    \caption{Comparison with existing test-time methods across five benchmark domains. Colored values are relative gains over the zero-shot baseline. To compare single-query and test-set methods consistently, we report average end-to-end runtime relative to the baseline. ``Image'' domain uses CIDEr in native units. All other domain accuracies are on a 0-100 scale. SIMIT improves performance across all domains, with SIMIT-FT$\to$ICL combination achieving the best results.
    \dimpp{selfR3 and SelfC16 can be removed}
    \dimpp{adding reference to the table? not sure} \onur{tried this but it doesn't fit}
    } 
    \vspace{-0.35em}
    \setlength{\tabcolsep}{2pt}
    \begin{tabularx}{\linewidth}{
        >{\centering\arraybackslash}m{0.9em}
        >{\centering\arraybackslash}m{0.9em}
        l
        *{6}{>{\centering\arraybackslash}X}c
    }
        \toprule
        & & \textbf{Method} & \textbf{Image} & \textbf{VQA} & \textbf{Math Sci} & \textbf{Doc Info} & \textbf{Multi} & \textbf{Av. Gain} & \textbf{Runtime} \\
        
        \midrule
        & & BAGEL-7B, Baseline & 1.08 & 64.6 & 86.4 & 78.4 & 82.2 & \normaldeltanum{0.00\%} & 1$\times$ \\
        
        \midrule
        \multirow{10}{*}{\rotatebox[origin=c]{90}{\textit{Single-Query Access}}}
        & \multirow{8}{*}{\rotatebox[origin=c]{90}{\textit{Test-Time Scaling}}}
        & Thinking ($1{,}000$ tokens) & 0.49 \deltanum{-54.0\%} & 59.4 \deltanum{-8.1\%} & 64.2 \deltanum{-25.7\%} & 72.9 \deltanum{-7.1\%} & 78.7 \deltanum{-4.3\%} & \normaldeltanum{-21.59\%} & 19$\times$ \\
        & & Thinking ($10{,}000$ tokens) & 0.50 \deltanum{-53.9\%} & 59.5 \deltanum{-8.0\%} & 64.6 \deltanum{-25.2\%} & 73.3 \deltanum{-6.5\%} & 79.1 \deltanum{-3.7\%} & \normaldeltanum{-21.28\%} & 44$\times$ \\
        
        & & Self-Refine (iters $=1$) & 1.00 \deltanum{-7.1\%} & 62.5 \deltanum{-3.2\%} & 86.4 \deltanum{+0.0\%} & 77.0 \deltanum{-1.9\%} & 83.3 \deltanum{+1.4\%} & \normaldeltanum{-2.90\%} & 5$\times$ \\
        & & Self-Refine (iters $=3$) & 1.00 \deltanum{-7.1\%} & 62.6 \deltanum{-3.2\%} & 86.4 \deltanum{+0.0\%} & 77.0 \deltanum{-1.9\%} & 83.3 \deltanum{+1.4\%} & \normaldeltanum{-2.87\%} & 16$\times$ \\
        & & Self-Refine (iters $=5$) & 1.00 \deltanum{-7.1\%} & 62.6 \deltanum{-3.2\%} & 86.4 \deltanum{+0.0\%} & 77.0 \deltanum{-1.8\%} & 83.4 \deltanum{+1.5\%} & \normaldeltanum{-2.84\%} & 25$\times$ \\
        
        & & Self-Consistency ($K=8$) & 1.07 \deltanum{-0.3\%} & 64.5 \deltanum{-0.3\%} & 86.2 \deltanum{-0.2\%} & 78.1 \deltanum{-0.4\%} & 82.6 \deltanum{+0.6\%} & \normaldeltanum{-0.20\%} & 5$\times$ \\
        & & Self-Consistency ($K=16$) & 1.08 \deltanum{+0.0\%} & 64.8 \deltanum{+0.2\%} & 86.4 \deltanum{+0.0\%} & 78.1 \deltanum{-0.4\%} & 82.4 \deltanum{+0.3\%} & \normaldeltanum{+0.17\%} & 9$\times$ \\
        & & Self-Consistency ($K=32$) & 1.09 \deltanum{+1.1\%} & 64.7 \deltanum{+0.2\%} & 86.8 \deltanum{+0.5\%} & 78.4 \deltanum{+0.0\%} & 82.0 \deltanum{-0.2\%} & \normaldeltanum{+0.41\%} & 19$\times$ \\
        

        \cmidrule(lr){2-10}
        & \multirow{2}{*}{\rotatebox[origin=c]{90}{\textit{Ours}}}
        & \textbf{SIMIT-ICL} & \textbf{1.17} \deltanum{+8.3\%} & \textbf{68.8} \deltanum{+6.5\%} & \textbf{87.2} \deltanum{+0.9\%} & \textbf{79.3} \deltanum{+1.1\%} & \textbf{84.2} \deltanum{+2.5\%} & \textbf{\normaldeltanum{+6.98\%}} & 25$\times$ \\
        & & \textbf{SIMIT-FT, Single-Query} & \textbf{1.09} \deltanum{+1.4\%} & \textbf{65.8} \deltanum{+1.9\%} & \textbf{86.6} \deltanum{+0.2\%} & \textbf{78.9} \deltanum{+0.6\%} & \textbf{83.3} \deltanum{+1.3\%} & \normaldeltanum{+1.69\%} & 47$\times$ \\

        \midrule
        \multirow{6}{*}{\rotatebox[origin=c]{90}{\textit{Test-Set Access}}}
        & \multirow{4}{*}{\rotatebox[origin=c]{90}{\textit{LFRL}}}
        & TTRL ($K=32$) & 1.08 \deltanum{+0.6\%} & 64.9 \deltanum{+0.3\%} & 86.6 \deltanum{+0.2\%} & 79.6 \deltanum{+1.5\%} & 81.9 \deltanum{-0.3\%} & \normaldeltanum{+0.59\%} & 75$\times$ \\
        & & EMPO ($K=32$) & 1.10 \deltanum{+1.9\%} & 66.1 \deltanum{+2.3\%} & 86.4 \deltanum{+0.0\%} & 79.3 \deltanum{+1.1\%} & 82.1 \deltanum{-0.1\%} & \normaldeltanum{+1.82\%} & 58$\times$ \\
        & & TTRV ($K=32$) & 1.09 \deltanum{+1.2\%} & 66.4 \deltanum{+2.8\%} & 86.8 \deltanum{+0.5\%} & 79.6 \deltanum{+1.5\%} & 82.8 \deltanum{+0.7\%} & \normaldeltanum{+2.07\%} & 41$\times$ \\
        & & SRLM ($K=4$) & 1.07 \deltanum{-1.0\%} & 66.3 \deltanum{+2.5\%} & 86.2 \deltanum{-0.2\%} & 78.6 \deltanum{+0.2\%} & 82.3 \deltanum{+0.2\%} & \normaldeltanum{+0.87\%} & 251$\times$ \\
        
        \cmidrule(lr){2-10}
        & \multirow{2}{*}{\rotatebox[origin=c]{90}{\textit{Ours}}}
        & \textbf{SIMIT-FT, Test-Set} & \textbf{1.14} \deltanum{+5.8\%} & \textbf{69.4} \deltanum{+7.5\%} & \textbf{87.6} \deltanum{+1.4\%} & \textbf{79.4} \deltanum{+1.3\%} & \textbf{83.9} \deltanum{+2.2\%} & \normaldeltanum{+6.39\%} & 55$\times$ \\
        & & \textbf{SIMIT-FT, Test-Set$\to$ICL} & \textbf{1.15} \deltanum{+6.9\%} & \textbf{69.6} \deltanum{+7.7\%} & \textbf{87.8} \deltanum{+1.6\%} & \textbf{79.7} \deltanum{+1.6\%} & \textbf{84.8} \deltanum{+3.2\%} & \textbf{\normaldeltanum{+7.20\%}} & 55$\times$ \\
        \bottomrule
    \end{tabularx}
    \label{tab:main_comparison}
    \vspace{-1em}
\end{table}

\FloatBarrier





Tab.~\ref{tab:main_comparison} compares SIMIT with 
inference-only test-time scaling (TTS) and label-free reinforcement learning (LFRL) methods.
TTS baselines include 
thinking \citep{wei2022chainofthought}, 
Self-Consistency \citep{wang2022selfconsistency}, and
Self-Refine \citep{madaan2023selfrefine}. For LFRL, we evaluate the
self-judging SRLM \citep{yuan2024selfrewarding} and the consensus-based
methods: TTRL \citep{zuo2025ttrl}, EMPO \citep{zhang2026empo}, TTRV
\citep{singh2026ttrv}. 
All methods use 
BAGEL-7B model without any
external models.

For a fair comparison, we tune SIMIT and all LFRL baselines on the same validation data, and report TTS baselines at several compute budgets. Each method is reported at its best configuration within a common, practical runtime budget. This budget
does not favor SIMIT since most LFRL baselines use more runtime than any SIMIT variant. We do not enforce exactly equal runtimes, because more computation does not always improve accuracy, and matching runtimes would replace a method's best
configuration with a weaker one. %
Appendix~\ref{sec:appendix_implementation_details} details the tuning protocol and runtime budget.

\mypar{Main Results.}
SIMIT improves performance across all five domains.
Appendix~\ref{sec:appendix_detailed_results} also reports the per-benchmark results.
SIMIT-ICL improves
16 benchmarks and ties one, achieving a mean
relative gain of $6.98\%$ without parameter updates. Combining test-set
finetuning with query-specific demonstrations raises the mean gain to
$7.20\%$ and outperforms SOTA 
methods in every domain. 
The gains can be substantial on individual benchmarks:
SIMIT-ICL increases NoCaps CIDEr from $0.755$ to $1.162$ ($+53.9\%$),
while test-set SIMIT-FT raises OK-VQA accuracy from $37.6$ to $54.2$
($+44.0\%$).  
Extra computation does not explain SIMIT's gains.
At $25\times$ the zero-shot runtime, SIMIT-ICL achieves a $+6.98\%$ mean gain, compared with at most $+2.07\%$ for LFRL baselines that run longer. 

\mypar{Test-Time Scaling.}
Self-Consistency provides modest gains, while Self-Refine often reduces
performance. BAGEL's thinking mode degrades scores, even with a $10{,}000$-token budget, 
consistent with recent evidence that
excess test-time generation can hurt perception-focused tasks through
overthinking and hallucination
\citep{sammani2026tts}.
Inspection reveals thinking mode yields 
overly 
verbose captions and instruction-following failures 
regarding the expected answer format.

\mypar{Single-Query Adaptation.} 
Our theoretical analysis shows that consensus-based LFRL (TTRL, EMPO, TTRV)
cannot reliably improve on the initial majority-vote prediction
under isolated single-query adaptation (Appendix~\ref{sec:appendix_lfrl_limitation_single_sample_adaptation}). 
Thus, we evaluate LFRL only in the test-set setting.
Their performance gains in the test-set adaptation setting are largely due to cross-sample capability transfer.
SIMIT-FT (SQ) and SIMIT-ICL improve performance even when each query is adapted in isolation.

\mypar{Consensus Reinforces Systematic Errors.} 
The base model often outputs ``unanswerable'' even for answerable queries, incorrectly claiming insufficient visual information (e.g., Fig.~\ref{fig:qualitative_methods}).
TTRL reinforces such biases: ``unanswerable'' is the majority vote in $99.2\%$ of training steps for VizWiz.
Only a few steps yield nonzero group-relative
advantages, and $79.3\%$ of those still reinforce ``unanswerable''.
Thus, consensus rewards can perpetuate the model's biases rather than provide corrective supervision.

\mypar{Generalization.} 
Separately, we also test SIMIT's generalization across models
(Appendix~\ref{sec:appendix_cross_model}) and benchmarks
(Appendix~\ref{sec:appendix_cross_dataset}).
SIMIT-ICL improves over the zero-shot baseline by $+2.21\%$ on SenseNova-U1~\citep{diao2026sensenova} and $+3.97\%$ on Lance~\citep{fu2026lance}, showing that the gains extend beyond BAGEL. 
In the cross-benchmark
experiments, transfer yields mean relative gains of 4.3\% within domains
and 0.5\% across domains, although target-specific adaptation 
remains stronger. This demonstrates the importance of targeted self-improvement and query-conditioned synthesis.

\dimpp{the flow of the last 4 paragraphs can be impoved. It seems like 4 different conclusions, and it's hard to know where to look at all the time. your main reference is Appendices instead of Tab 1. }
\dimpp{it's very hard to understand that there is another completely different experiment on coss-benchamark evaluation with different settings. }

\begin{takeaway}
In-context and in-weight adaptation offer complementary gains: 
ICL leads on 8 benchmarks and test-set FT on 9. Their combination performs best overall. 
ICL does not require weight updates or access to the whole test set. 
Test-set FT 
outperforms 
isolated FT, 
due to cross-sample transfer.
\end{takeaway}


\subsection{Comparison of Self-Generated Synthetic Data vs. Real Data}
\label{sec:comparison_real_synthetic}

\begin{figure}[t]
    \vspace{-1.5ex}
    \centering
    \includegraphics[width=1.0\textwidth]{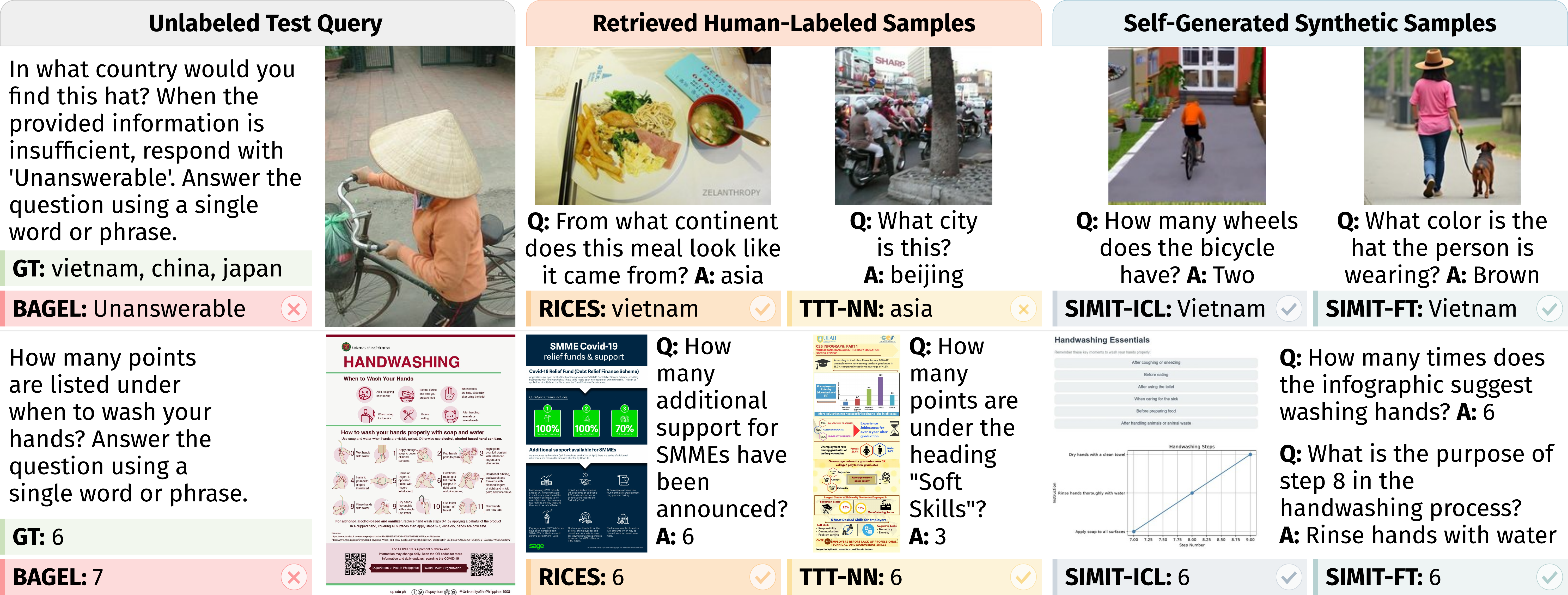}
    \caption{Each method's retrieved/generated
  demonstrations and final answers for two test queries. 
  Nearest real neighbors can be off-topic, and TTT-NN 
  sometimes inherits that mismatch. SIMIT's demonstrations match the query's subject, and both its variants answer correctly. 
  See Appendix \ref{sec:appendix_qualitative_results} for more qualitative samples. Retrieval-based methods are bottlenecked by
  dataset coverage. Self-generated demos, targeted to the query,
  recover from it without requiring annotated retrieval sets.
    }
    \label{fig:qualitative_methods}
    \vspace{-1em}
\end{figure}


In Fig.~\ref{fig:qualitative_methods}, for each adaptation mode, we compare SIMIT's self-generated 
data with a real human-annotated counterpart. 
RICES \citep{alayrac2022flamingo} retrieves the nearest real neighbors with a
multimodal embedding model (Qwen3-VL-Embedding-2B; \cite{li2026qwen3}) and uses
them as in-context demonstrations, matching SIMIT-ICL.
TTT-NN \citep{hardt2024tttnn} finetunes on
those same retrieved neighbors for a single test sample, matching SIMIT-FT
(Single-Query).

\begin{takeaway}
For certain adaptation modes and domains, SIMIT's synthetic data outperform real human-annotated data. 
Appendix~\ref{sec:appendix_scaling_real_synthetic} reports synthetic-to-real exchange rates and scaling behaviors. 
\end{takeaway}

\begin{table}[t]
    \centering
    \small
    \caption{
    We ablate adaptive budget allocation (\textbf{ABA}), diversity-encouraging prompting in triplet synthesis (\textbf{DP}), skill library (\textbf{SL}), verification (\textbf{V}) and difficulty filter (\textbf{DF}). 
    Each component matters.
    }
    \vspace{-0.45em}
    \setlength{\tabcolsep}{1pt}
    \begin{tabularx}{\linewidth}{
        *{5}{>{\hsize=0.55\hsize\centering\arraybackslash}X}
        *{8}{>{\hsize=1.15\hsize\centering\arraybackslash}X}
    }
    \toprule
    \multicolumn{5}{c}{\textbf{Variant}} & \multicolumn{6}{c}{\textbf{Understanding Metrics}} & \multirowcell{2}[-2pt][c]{\textbf{Data}\\\textbf{Quality}} & \multirowcell{2}[-2pt][c]{\textbf{Runtime}\\\textbf{(min)}}\\
    \cmidrule(lr){1-5} \cmidrule(lr){6-11}
    \textbf{ABA} & \textbf{DP} & \textbf{SL} & \textbf{V} & \textbf{DF} & \textbf{Image} & \textbf{VQA} & \textbf{Math Sci} & \textbf{Doc Info} & \textbf{Multi} & \textbf{Av. Gain} & & \\
    \midrule
    \multicolumn{5}{c}{Baseline, BAGEL-7B}
    & 1.08 & 64.6 & 86.4 & 78.4 & 82.2 & \normaldeltanum{0.00\%} & -- & 5 \\
    
    \ding{51} & \ding{51} & \ding{51} & \ding{51} & \ding{51}
    & \textbf{1.14} \deltanum{+5.8\%}
    & \textbf{69.4} \deltanum{+7.5\%}
    & \textbf{87.6} \deltanum{+1.4\%}
    & \textbf{79.4} \deltanum{+1.3\%}
    & \textbf{83.9} \deltanum{+2.2\%}
    & \textbf{\normaldeltanum{+6.39\%}}
    & \textbf{86.0} & 253 \\
    
    \ding{55} & \ding{51} & \ding{51} & \ding{51} & \ding{51}
    & 1.07 \deltanum{-0.6\%}
    & 68.5 \deltanum{+6.0\%}
    & 87.0 \deltanum{+0.7\%}
    & 78.8 \deltanum{+0.5\%}
    & 83.4 \deltanum{+1.5\%}
    & \normaldeltanum{+3.27\%}
    & 83.5 & 376 \\
    
    \ding{55} & \ding{51} & \ding{51} & \ding{51} & \ding{55}
    & 1.11 \deltanum{+2.8\%}
    & 68.3 \deltanum{+5.7\%}
    & 87.4 \deltanum{+1.2\%}
    & 79.0 \deltanum{+0.8\%}
    & 83.3 \deltanum{+1.3\%}
    & \normaldeltanum{+4.35\%}
    & 81.8 & 364 \\
    
    \ding{55} & \ding{55} & \ding{51} & \ding{51} & \ding{55}
    & 1.10 \deltanum{+2.2\%}
    & 68.3 \deltanum{+5.7\%}
    & 86.4 \deltanum{+0.0\%}
    & 77.6 \deltanum{-1.1\%}
    & 81.8 \deltanum{-0.5\%}
    & \normaldeltanum{+3.47\%}
    & 85.6 & 452 \\
    
    \ding{55} & \ding{51} & \ding{55} & \ding{51} & \ding{55}
    & 1.08 \deltanum{+0.1\%}
    & 67.2 \deltanum{+4.0\%}
    & 86.6 \deltanum{+0.2\%}
    & 76.7 \deltanum{-2.2\%}
    & 81.8 \deltanum{-0.5\%}
    & \normaldeltanum{+1.69\%}
    & 82.9 & 516 \\
    
    \ding{55} & \ding{51} & \ding{51} & \ding{55} & \ding{55}
    & 1.11 \deltanum{+3.3\%}
    & 67.3 \deltanum{+4.2\%}
    & 86.6 \deltanum{+0.2\%}
    & 77.8 \deltanum{-0.8\%}
    & 81.3 \deltanum{-1.0\%}
    & \normaldeltanum{+3.08\%}
    & 75.4 & 214 \\
    \bottomrule
    \end{tabularx}
    \label{tab:ablation_components}
    \vspace{-1.1em}
\end{table}

\FloatBarrier
\subsection{Ablation Studies}

We conduct ablations with SIMIT-FT (Test-Set), which outperforms
SIMIT-ICL on more benchmarks. 
We examine
the contributions of the synthesis components and whether generating both
questions and images produces more effective supervision.
For evaluation, we use benchmark scores and data quality scores from agentic evaluation via Claude Sonnet 5 (Appendix~\ref{sec:appendix_prompts_diagram_categories}). 
This external evaluator only audits
the generated data and provides no feedback to the synthesis or
adaptation pipeline. 


\mypar{Component Contributions.}
Tab.~\ref{tab:ablation_components} first removes ABA and then DF. 
DP, SL, and V are then removed individually from the shared
reference with ABA \& DF disabled. This fixes the synthesis budget 
and avoids repeating ABA \& DF
hyperparameter searches for each variant. 
The full pipeline achieves the best scores, 
showing the necessity of each component. 
ABA and DF improve both efficiency and accuracy.
If the model is already confident about its answer, using more samples can hurt. 
Because DF thresholds are tuned jointly with ABA, DF alone without re-tuning can underperform the variant with both disabled. Its higher data quality suggests that DF would recover its gains with dedicated tuning.
Removing V or DP also reduces performance. 
Thus, both validity and diversity matter.

\begin{wrapfigure}{r}{0.49\textwidth}
    \centering
\vspace{-1.1em}
\includegraphics[width=1.0\linewidth]{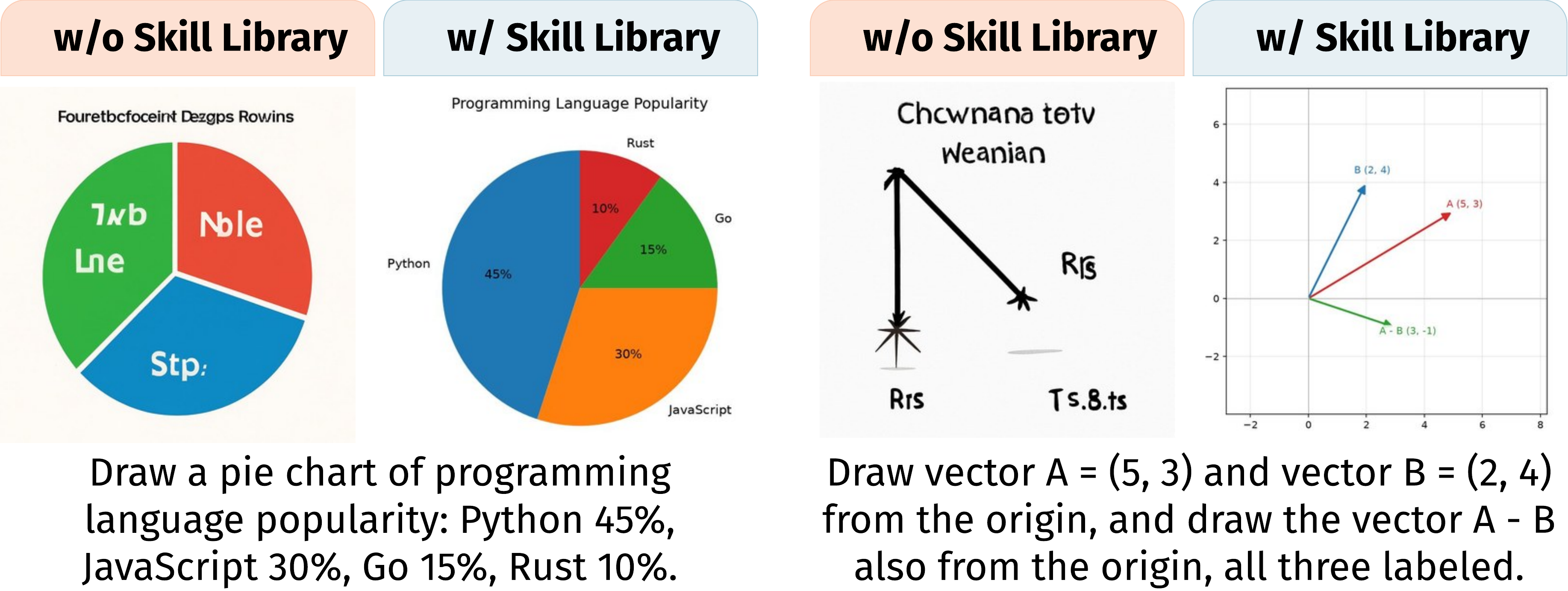}
\vspace{-1.6em}
    \caption{
    Generations with and without our skill library. BAGEL cannot draw structured visuals, and often fails to generate legible text. 
    }
    \label{fig:diagram_library_qualitative}
\vspace{-0.9em}
\end{wrapfigure}

Among the synthesis components, 
removing the skill library particularly reduces
performance. 
Fig.~\ref{fig:diagram_library_qualitative} illustrates why:
native BAGEL generation can fail to produce readable text or 
requested 
structure, while rendering skills produce 
faithful
images. 
Appendix~\ref{sec:appendix_skill_lib_ablation} provides
more examples.


\begin{wraptable}{r}{0.49\textwidth}
\vspace{-1em}
    \centering
    \small
    \caption{Question and image synthesis ablation at matched sample counts (ABA \& DF disabled). \textbf{\Circled{1}} uses original questions from the test queries but generates new images. \textbf{\Circled{2}} uses original images but generates new questions. \textbf{\Circled{3}} generates both.} 
    \vspace{-0.5em}
    \setlength{\tabcolsep}{1pt}
    \begin{tabularx}{\linewidth}{
    >{\hsize=0.4\hsize\centering\arraybackslash}X
    *{5}{>{\hsize=0.6\hsize\centering\arraybackslash}X}
    >{\hsize=0.95\hsize\centering\arraybackslash}X
    >{\hsize=0.56\hsize\centering\arraybackslash}X
    }
        \toprule
         & \textbf{Image} & \textbf{VQA} & \textbf{Math} & \textbf{Doc.} & \textbf{Multi} & \textbf{Av. Gain} & \textbf{Time} \\ 
        \midrule
        \textbf{\Circled{1}} & 1.02 & 67.5 & 85.6 & 77.0 & 83.9 & \normaldeltanum{+1.57\%} & 487 \\
        \textbf{\Circled{2}} & 1.01 & 66.4 & 85.4 & 69.6 & 84.0 & \normaldeltanum{-2.14\%} & 211 \\
        \textbf{\Circled{3}} & 1.11 & 68.3 & 87.4 & 79.0 & 83.3 & \normaldeltanum{+4.35\%} & 364 \\
        \bottomrule
    \end{tabularx}
    \label{tab:ablation_fix_qi}
\vspace{-1.0em}
\end{wraptable}

\mypar{Joint Question \& Image Synthesis.}
Tab.~\ref{tab:ablation_fix_qi} compares joint generation against  retaining either  original question or image. 
%
Joint generation shows the benefit of diversity by expanding non-redundant local support around the test query. 

Fixing the image restricts the available visual evidence.
Fixing the question restricts the instructions under which that evidence is used. 
Generating both components can broaden relevant visual content and question-answer relations around the target query.
The weaker fixed-image variant highlights the benefit of generating new images over approaches that synthesize questions for a fixed image pool \citep{wei2026mmupt, he2026visplay}.
Also, a straightforward
multimodal extension of text-based TT-SI~\citep{acikgoz2026ttsi}, generating
question-answer pairs for a fixed test image, 
misses a
source of useful variation.
But some domains can benefit from retaining
the question. 
Unnecessary 
variation introduces a mismatch between synthetic training examples
and test queries.

\FloatBarrier
\begin{takeaway}
Both supervision quality and diversity matter. 
Useful diversity broadens training coverage while preserving alignment with the target task.
Joint question and image generation generally helps. 
\end{takeaway}



\FloatBarrier
\subsection{Why Does SIMIT Outperform LFRL?}
\label{sec:label_noise}

\begin{wrapfigure}{r}{0.34\textwidth}
\vspace{-3.2em}
    \centering
    \includegraphics[width=\linewidth]{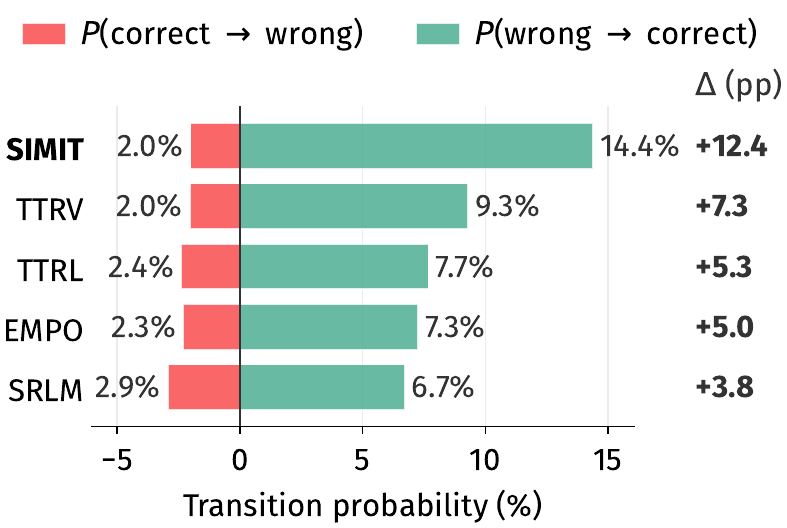}
    \caption{Test queries are split by whether the Self-Consistency baseline already answers correctly. Bars show the probability of 
    \textcolor[HTML]{43AA8B}{correcting initially wrong predictions} vs. \textcolor[HTML]{F94144}{corrupting initially correct ones}; $\Delta$ is the difference.
    Consensus-based supervision mainly stabilizes majority-correct predictions, while SIMIT corrects more errors.
    }
    \label{fig:stratified_gain}
\vspace{-1.2em}
\end{wrapfigure}

We distinguish supervision quality from its utility. Models may fail to benefit from correct supervision, while incorrect supervision may be less harmful in queries they already answer confidently and correctly. We therefore first measure how often adaptation corrects initial errors or disrupts correct predictions, and then assess the correctness of supervision and whether its errors concentrate where the model already fails.

\textbf{Stratified Gain Analysis.}
We split test queries by whether Self-Consistency ($K=32$) answers correctly, then measure the correction rate among initially wrong queries and the disruption rate among initially correct ones, reporting their difference as 
$\Delta = P(\text{wrong}\to\text{correct}) - P(\text{correct}\to\text{wrong}).$ 
As shown in Fig.~\ref{fig:stratified_gain},
SIMIT corrects 14.4\% of initially wrong queries while disrupting only 2.0\% of initially correct ones ($\Delta$=+12.4pp).
The strongest LFRL baseline, TTRV, corrects only 9.3\% at the same disruption rate, and others correct even fewer ($\Delta\leq$+5.3pp).
Their gains over the zero-shot baseline thus combine a smaller net correction with the majority-vote gain, which sharpening preserves by stabilizing low-margin correct predictions. 
SIMIT's advantage lies in correcting more errors, not in preserving more correct predictions.
This gap follows from how consensus-based methods operate: they reward agreement and also predict by majority vote, so training mainly sharpens predictions the majority already answers correctly.
Such prediction-derived supervision can reinforce existing errors~\citep{wang2026beyond}.
In contrast, SIMIT's imagination-based, independently realized supervision offers a route to correct them.

\begin{wrapfigure}{r}{0.34\textwidth}
\centering
\vspace{-1.3em}
\includegraphics[width=\linewidth]{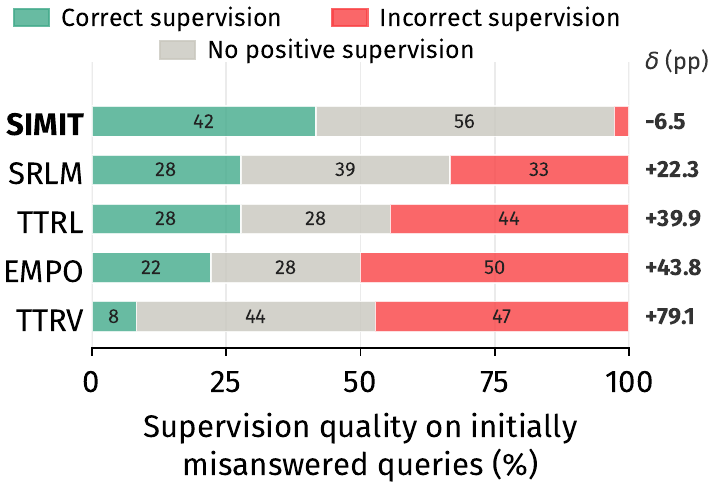}
\caption{
Share of each method's positively-encouraged supervision on queries the base model answers incorrectly that is
\textcolor[HTML]{43AA8B}{correct}, \textcolor[HTML]{C9C7BE}{absent}, or \textcolor[HTML]{F94144}{wrong}. 
$\delta$ is the same error rate's gap
between model-wrong and model-right queries. Here, SIMIT is the only method with net-positive supervision quality.
}
\label{fig:label_noise_stacked}
\vspace{-0.8em}
\end{wrapfigure}

\textbf{Label-Noise, Model-Noise Alignment.} Despite many verification
steps, SIMIT's synthesized triplets are not guaranteed to be clean since the generated image
may fail to align with the intended description, leaving an incorrect
label. Both the overall rate of such errors and whether they concentrate on
queries the base model already fails matter. 
To measure this, for each method, we sample one positively-encouraged pseudo-label per query following
its own reward rule. Then, an external blind evaluator (Claude Sonnet 5)
judges their correctness. Writing
$L{=}1$ for incorrect supervision and $M{=}1$ for a base-model error on the
same query, $\delta = P(L{=}1\mid M{=}1) - P(L{=}1\mid M{=}0)$ measures
whether supervision errors concentrate where the model already fails. On
queries the base model answers incorrectly,
SIMIT's supervision is correct far more often than it is incorrect, while
every LFRL baseline shows the reverse (Fig.~\ref{fig:label_noise_stacked}). SIMIT's overall supervision-error
rate is also roughly a third of every baseline's ($12\%$ vs.\
$33$--$38\%$), and its $\delta$ is small and negative ($-6.5$pp) whereas $\delta$ is large and positive for all LFRL baselines 
($+22$ to $+79$pp; Appendix~\ref{sec:appendix_label_model_noise_alignment}).


\FloatBarrier
\begin{takeaway}
SIMIT corrects more initial errors than the existing methods. Its supervision has a lower error rate and weaker estimated alignment with base-model failures. 
SIMIT is the only method with more correct than incorrect supervision on initially wrong test queries.
\end{takeaway}

  \section{Conclusion}
\label{sec:conclusion}




SIMIT creates query-specific multimodal self-supervision by specifying answers before constructing the corresponding images, allowing a VLM to learn without first solving the target test query. 
Combining finetuning with ICL achieves a $+7.20\%$ mean gain over BAGEL-7B baseline without any annotations or extra models. This turns test-time compute into targeted practice, offering a route to more accurate automatic annotation and personalized agents learning from the tasks they encounter.

  \clearpage

  \ificlrfinal
  \section*{Acknowledgments}
  Dim P. Papadopoulos was supported by the DFF Sapere Aude Starting Grant "ACHILLES". Desmond Elliott was supported by the European Union’s Horizon 2020 research and innovation program under grant agreement No. 101135671 (TrustLLM) and a research grant (VIL53122) from Villum Fonden. This work was supported by the Pioneer Centre for AI, DNRF grant number P1. We would like to thank Marco Schouten, Thanos Delatolas and Stella Frank for proofreading and insightful discussions. 
  \fi
  

  \section*{Reproducibility Statement}

  \ificlrfinal
  To ensure the replicability of our findings, we release our code,
  \else
  To ensure replicability, we will release our code upon publication,
  \fi
  allowing the research community to reproduce our results and build upon our
  contributions. Our experimental setup is based on publicly accessible models,
  ensuring that all resources are readily obtainable by other researchers. We provide
  comprehensive details regarding the prompts and hyperparameters utilized across
  our experiments in Appendix~\ref{sec:appendix_implementation_details}. Also, Appendix~\ref{sec:appendix_dataset_stats}
  contains thorough descriptions of the benchmarks 
  employed in our study. All evaluation benchmarks utilized in this work are established
  and widely used standards within the field.
  Our commitment to
  transparency extends beyond code release, as we meticulously detail every aspect
  of our experimental methodology to enable faithful reproduction of our work.

  \section*{Author Statement on the Use of Large Language Models}
  During the preparation of this paper, large language models were used solely for 
  language polishing and coding assistance. They were not used for research ideation,
  experiment design, analysis, or writing of scientific content. All research
  contributions are the sole responsibility of the authors. As part of our method, we also use vision-language models to generate synthetic training datasets. 

  \bibliography{main}
  \bibliographystyle{iclr2027_conference}

  \clearpage
  \appendix
  \section{Dataset Statistics}
\label{sec:appendix_dataset_stats}


We evaluate on 17 benchmarks across five domains using LMMs-Eval-Lite~\citep{zhang2025lmmseval}, a compact evaluation suite derived from LMMs-Eval. It uses coreset selection to retain representative examples, reducing evaluation cost while preserving broad task coverage. The original study reports strong correlations between scores on the selected subsets and the corresponding full benchmarks, supporting their use for efficient evaluation and ablation studies. Tab.~\ref{tab:lmms_eval_lite} summarizes our datasets and sample counts.


\begin{table}[tbh]
    \centering
    \small
    \caption{LMMs-Eval Lite with additional datasets.}
    \label{tab:lmms_eval_lite}
    \setlength{\tabcolsep}{5pt}
    \begin{tabularx}{\linewidth}{Xlccc}
        \toprule
        \textbf{Task Domain} & \textbf{Dataset} & \textbf{Retrieval/Training} & \textbf{Val.} & \textbf{Eval.} \\
          \midrule

          \multirow{5}{*}{\shortstack[l]{Image Understanding\\ \& Captioning}}
          & Flickr30k~\citep{youngetal2014flickr30k}   & 2000  & 50 & 500 \\
          & NoCaps~\citep{Agrawal2019nocaps}      & 2000   & 50 & 500 \\
          & TextCaps~\citep{sidorov2020textcaps}    & 2000  & 50 & 500 \\
          & RefCOCO~\citep{kazemzadeh2014refcoco}     & 2000  & 50 & 500 \\
          & COCO~\citep{lin2014coco}        & 2000   & 50 & 500 \\
          \midrule      
          
          \multirow{5}{*}{Visual Question Answering}
          & GQA~\citep{Hudson2019gqa}         & 2000 & 50 & 500 \\
          & OK-VQA~\citep{Marino2019okvqa}       & 2000   & 50 & 500 \\
          & VizWiz-VQA~\citep{Gurari2018vizwiz}  & 2000   & 50 & 500 \\
          & VQA-v2(-Small)~\citep{Goyal2017vqav2}      & 2000  & 50 & 500 \\
          & TextVQA~\citep{Singh2019textvqa}     & 2000  & 50 & 500 \\
          \midrule

          \multirow{1}{*}{Math \& Science}
          & AI2D~\citep{kembhavi2016ai2d}        & 2000   & 50 & 500 \\
          \midrule      

          \multirow{3}{*}{\shortstack[l]{Document \& Infographic\\ Understanding}}
          & ChartQA~\citep{masry2022chartqa}     & 2000   & 50 & 500 \\
          & DocVQA~\citep{Mathew2021docvqa}      & 2000   & 50 & 500 \\
          & InfoVQA~\citep{Mathew2022infovqa}     & 2000   & 50 & 500 \\
          \midrule
          
          \multirow{3}{*}{Multi-discipline}
          & MM-Bench (CN)~\citep{liu2024mmbench} & 2000 & 50 & 500 \\
          & MM-Bench (EN)~\citep{liu2024mmbench} & 2000 & 50 & 500 \\
          & SEED-Bench~\citep{li2023seed}  & 2000 & 50 & 500 \\
          \midrule
          
          \textbf{Total} & & \textbf{34000} & \textbf{850} & \textbf{8500} \\
        \bottomrule
    \end{tabularx}
\end{table}

For each benchmark, we use 50 validation examples and 500 evaluation examples, totaling 850 and 8,500 examples, respectively. Validation sets support hyperparameter selection for SIMIT and all baseline methods. For SIMIT, validation data determine the adaptive-budget parameters $(\varepsilon, A_0, A_1, B, K_{\max})$ and difficulty-filtering thresholds $(T_{\mathrm{low}}, T_{\mathrm{high}})$, with Bayesian optimization used for hyperparameter search. We fix the selected hyperparameter configurations before reporting results on the evaluation sets. Ground-truth answers from the evaluation set are used only for scoring. All methods operate on unlabeled queries.

Each benchmark also provides a pool of 2,000 annotated retrieval/training examples, totaling 34,000 examples. These pools support the real-versus-synthetic comparisons in Sec.~\ref{sec:comparison_real_synthetic}: they supply retrieved in-context demonstrations, retrieved neighbors for single-query finetuning, and training examples for test-set finetuning. SIMIT uses its own synthetic examples as adaptation supervision.

We use CIDEr for all five ``Image Understanding \& Captioning'' benchmarks. CIDEr measures agreement with reference human-annotated descriptions through TF--IDF-weighted $n$-gram similarity~\citep{vedantam2015cider}.
The five VQA benchmarks and AI2D use exact-match accuracy. ChartQA uses relaxed accuracy, which tolerates small numerical deviations, while DocVQA and InfoVQA use average normalized Levenshtein similarity (ANLS), which allows partial credit for small character-level differences. MMBench (CN/EN) and SEED-Bench use multiple-choice accuracy. 
Higher scores indicate better performance for every metric.

Captioning scores are reported in native CIDEr units, and all other scores are expressed on a 0-100 scale. Domain scores are arithmetic means across constituent benchmarks, with relative improvements computed from the 
domain averages. To aggregate across heterogeneous metrics, overall gains are computed as the mean of benchmark-level relative improvements over the zero-shot baseline across all 17 benchmarks, giving each benchmark equal weight.


\definecolor{simitblue}{HTML}{277DA1}      
\definecolor{simitbluedark}{HTML}{196382}  
\colorlet{simitcell}{simitblue!12}         
\newcommand{\taxfam}[1]{\textbf{#1}:}
\newcommand{\taxours}[1]{\textcolor{simitbluedark}{\textbf{#1}}}

\begin{table}[t]
\centering
\footnotesize
\setlength{\tabcolsep}{3pt}
\renewcommand{\arraystretch}{1.33}
\caption{Taxonomy of related work. Each row is a problem setup, defined by axes A and B. Each column is a design choice, axis C. Bold text describes a method family, followed by representative methods where applicable.
\textbf{(A) Labels before deployment:} whether labeled examples of the target task are available beyond pretraining (A1) or not (A2).
\textbf{(B) Test-time access to the target:} whether the method sees each test query in isolation (B1), the unlabeled test set as a batch or a stream (B2), or no test data or no estimation of the test distribution, so all adaptation happens before deployment (B3).
\textbf{(C) Adaptation space:} whether the adaptation is carried in the prompt or not stored at all (C1), or written into the model weights (C2).
SIMIT (shaded) uses no labels. It adapts in context or in weights for a single query, and in weights across the test set.}
\label{tab:taxonomy}
\begin{tabularx}{\linewidth}{@{}l >{\raggedright\arraybackslash}X >{\raggedright\arraybackslash}X}
\toprule
\textbf{Setup (A, B)} & \textbf{C1: Context / None} & \textbf{C2: Weights} \\
\midrule
\multicolumn{3}{@{}l}{\textit{\textbf{A1}: Labeled task data}} \\
\quad\textbf{B1}: Per-query
  & \taxfam{Retrieved labeled in-context demos} RICES~\citep{alayrac2022flamingo}
  & \taxfam{Training on labeled neighbors} Local learning~\citep{bottou1992local}, TTT-NN~\citep{hardt2024tttnn} \\
\quad\textbf{B2}: Test set
  & \textbf{Prompt optimization on labeled data, then self-evolving memory at test time}
  & \taxfam{Transductive learning} 
  TSVM~\citep{joachims1999transductive} \\
\quad\textbf{B3}: None
  & \taxfam{Fixed demos or prompts} Few-shot ICL~\citep{alayrac2022flamingo}
  & \taxfam{Supervised or RL post-training} SFT, RLVR, RLHF~\citep{dai2023instructblip,sun2024aligning} \\
\midrule
\multicolumn{3}{@{}l}{\textit{\textbf{A2}: No labeled data}} \\
\quad\textbf{B1}: Per-query
  & \cellcolor{simitcell}\taxfam{Test-time scaling} CoT~\citep{wei2022chainofthought}, Self-Consistency~\citep{wang2022selfconsistency}, Self-Refine~\citep{madaan2023selfrefine}, TTAug~\citep{kaya2026efficient};\newline
    \taxfam{Self-generated demos} Self-ICL~\citep{chen2023selficl}, Analogical Prompting~\citep{yasunaga2024large}, \taxours{SIMIT-ICL}
  & \cellcolor{simitcell}\taxfam{Per-query finetuning, reset after each query} TTT~\citep{sun2020testtimetraining}, TTAdapt~\citep{kaya2026efficient}, TT-SI~\citep{acikgoz2026ttsi}, \taxours{SIMIT-FT (Single-Query)} \\
\quad\textbf{B2}: Test set
  & \textbf{Self-evolving memory accumulated across test queries}
  & \cellcolor{simitcell}\taxfam{Label-free RL} TTRL~\citep{zuo2025ttrl}, TTRV~\citep{singh2026ttrv}, EMPO~\citep{zhang2026empo}, MM-UPT~\citep{wei2026mmupt};\newline
    \taxfam{Unsupervised objectives} TLM~\citep{hu2025tlm}; \taxours{SIMIT-FT (Test-Set)} \\
\quad\textbf{B3}: None
  & \textbf{Self-evolving memory built by solving self-proposed tasks} 
  & \taxfam{Zero-data self-play on self-proposed tasks} Absolute Zero~\citep{zhao2026absolute}, R-Zero~\citep{huang2026rzero}, VisPlay~\citep{he2026visplay} \\
\bottomrule
\end{tabularx}
\end{table}

\section{Extended Related Work}
\label{sec:appendix_extendedrelatedwork}

\mypar{Test-Time Scaling (TTS)} methods improve predictions by allocating
additional inference compute
\citep{zhang2025ttssurvey,yang2025test}. This compute can be deployed along two
main axes. \emph{Sequential} methods extend or revise an evolving solution
trajectory through scratchpad or chain-of-thought reasoning, budget forcing,
or iterative self-feedback
\citep{nye2022showyourwork,wei2022chainofthought,muennighoff2025s1,madaan2023selfrefine}.
\emph{Parallel} methods sample multiple candidate responses and either
aggregate them through self-consistency or majority voting, or select one using
a verifier or reward model, as in best-of-$K$
\citep{wang2022selfconsistency,cobbe2021training,brown2024llmonkeys,snell2025scaling,jeddi2026avis}.
However, not all TTS methods are compatible with our problem setting:
approaches requiring externally trained verifiers or reward models introduce
supervision excluded by test-time self-improvement, which assumes neither
labels nor external models. Self-contained TTS methods can 
elicit latent competence, particularly on verifiable tasks. Nevertheless, their gains
depend on the base model's generative support and the reliability of the
aggregation or selection signal. For instance, majority voting improves over a single prediction when each candidate response is correct with probability $p>0.5$. But when $p<0.5$, it amplifies the prevailing error \citep{wei2026mmupt}.
This limitation is even more pronounced for open-ended tasks such as image captioning, where sampled responses are typically all distinct; consequently, no majority exists,
and voting degenerates to arbitrary candidate selection. More fundamentally,
conventional TTS leaves the model unchanged: computation spent on one query
does not directly correct underlying misconceptions or benefit subsequent
queries. SIMIT instead converts test-time compute into verified,
query-conditioned practice data, enabling adaptation in context or in weight space.

\mypar{Label-Free Reinforcement Learning (LFRL)} closes the loop left open by inference-only TTS: instead of discarding
sampled responses after aggregation, it converts model-intrinsic signals into
pseudo-rewards and updates the model on unlabeled test data. Earlier
self-training methods fit high-confidence, self-consistent solutions using SFT
\citep{huang2023selfimprove}, whereas TTRL uses the majority-voted answer as a
pseudo-label and rewards agreement with it through GRPO
\citep{zuo2025ttrl}. TTRV extends this paradigm to VLMs using
frequency-proportional rewards and entropy regularization
\citep{singh2026ttrv}.
The failures of majority voting are even more consequential under LFRL: when open-ended responses such as captions are all distinct, an arbitrary pseudo-label is selected and then
reinforced through training.
Semantic clustering-based approaches, such as EMPO, relax
exact matching \citep{zhang2026empo}, but free-form
clustering may require an auxiliary model, violating our setting.
Most closely related to SIMIT, MM-UPT additionally
synthesizes questions around existing images; however, it neither generates
new visual inputs nor constructs their answers, and still derives supervision
through majority voting \citep{wei2026mmupt}. 
Because rewards are recomputed online, these methods can surpass the initial
model's majority-vote accuracy: parameter sharing enables cross-query transfer,
and the updated policy can produce better pseudo-labels in later iterations
\citep{zuo2025ttrl}. Nevertheless, their supervision remains highly coupled to the
model's own beliefs. Recent analysis shows that certainty- and
consistency-based intrinsic rewards primarily sharpen the model's initial
output distribution \citep{he2026howfar}. Despite their different reward formulations, all these methods remain fundamentally consensus-based and therefore inherit the limitations of the model's sampled answer distribution. 
Such reinforcement is useful only when confidence aligns with correctness,
but can also amplify systematic biases. For example, a generic
deflection such as \textit{``unanswerable''} may be reinforced whenever it
dominates the sampled responses, even for answerable queries.
MM-UPT similarly demonstrates that intrinsic rewards can amplify
errors on deliberately difficult data when the model lacks relevant
prior knowledge \citep{wei2026mmupt}.
This dependence on the model's existing beliefs is particularly
restrictive under query-isolated single-query adaptation:
cross-query transfer is unavailable, and reinforcing consensus
provides no independent signal for correcting an initially wrong
majority.
%
An orthogonal approach 
is to use self-judging rewards, where the model evaluates the correctness of its own responses, as in SRLM \citep{yuan2024selfrewarding}.
However, such rewards remain bounded by the model's own evaluation capabilities and are
vulnerable to reward hacking \citep{gao2023scaling}. 
SIMIT instead
specifies each answer before visual realization and subsequently verifies the
resulting triplet, decoupling supervision from the solver's current answer
distribution.

\mypar{Test-Time Training (TTT)} updates model parameters during inference to specialize on an observed input or target distribution \citep{sun2020testtimetraining}, motivated by the earlier works on local and transductive learning \citep{bottou1992local, joachims1999transductive}. The central challenge is obtaining reliable supervision aligned with the query without access to its answer. TTT-NN \citep{hardt2024tttnn} finetunes on retrieved neighbors.
However, its performance depends on the coverage and quality of an external corpus.
Other approaches exploit the test context itself.
On ARC, \citet{akyurek2025tttarcagi} augment supplied input–output demonstrations using task-preserving transformations, relying on labeled exemplars and the structure of the grid-based domain.
TTAdapt \citep{kaya2026efficient} iteratively finetunes a VLM on pseudolabels obtained by aggregating token probabilities across augmented versions of the test input.
TLM \citep{hu2025tlm} instead minimizes perplexity on unlabeled test inputs, supporting domain adaptation without explicitly supervising the reasoning required to answer a particular query.
More closely related, TT-SI \citep{acikgoz2026ttsi}
finetunes an LLM on self-generated examples for tool-use queries, validating tool calls and using majority agreement to select function labels. A straightforward extension to VLMs would adapt on generated text-only question–answer pairs. SIMIT instead constructs complete multimodal examples by specifying answers before generating and verifying the corresponding images. This introduces a visual realization problem across natural and structured domains, with supervision grounded in constructed visual content.
  \section{From Target-Risk Minimization to Synthesis Principles}
\label{sec:appendix_synthesis_principles}

Our data-synthesis pipeline design 
follows three main principles: \emph{accuracy}, \emph{learner-relative complexity},
and \emph{diversity}. 
Answer-first construction, structured rendering, and verification support accuracy. 
Difficulty filtering selects examples that challenge the model.
Diversity-encouraging prompting in triplet synthesis expands coverage. 
However, since our goal is targeted improvement, this diversity must remain
relevant to the test query. Thus, our triplet synthesis stage is query-aware.
Together, these mechanisms aim to produce verified, informative,
and non-redundant supervision within the target domain.
Here, we provide a formal motivation for these three principles.

Let $\mathcal T=\{(x_i,y_i)\}_{i=1}^N$ denote the test set, where
$x_i=(I_i,q_i)$ is observed but $y_i$ is unknown. For a synthetic dataset
$S$ and adaptation operator $\mathcal A$, the ideal data-design objective is
\begin{equation}
\label{eq:ideal_data_design}
S^\star
=
\arg\min_{|S|\le B}
R_{\mathcal T}\!\left(\mathcal A(\theta,S)\right),
\qquad
R_{\mathcal T}(\theta)
=
\frac{1}{N}\sum_{i=1}^N
\ell(f_\theta(x_i),y_i).
\end{equation}
Thus, synthetic data is useful only insofar as the resulting adaptation
reduces risk on the given test set.

\mypar{Update Approximation.}
Consider one-step fine-tuning,
$\theta^+=\theta-\eta\widehat g_S$, where $\widehat g_S$ is the gradient
estimated from $S$, and let
$g_{\mathcal T}=\nabla R_{\mathcal T}(\theta)$ be the unavailable oracle
test gradient. If $R_{\mathcal T}$ is $\beta$-smooth and
$\eta=1/\beta$, then
\begin{equation}
\label{eq:gradient_approximation}
R_{\mathcal T}(\theta^+)
\le
R_{\mathcal T}(\theta)
-
\frac{1}{2\beta}\|g_{\mathcal T}\|^2
+
\frac{1}{2\beta}
\|\widehat g_S-g_{\mathcal T}\|^2.
\end{equation}
Hence, good synthetic data should induce an update close to the oracle
test-set update.

Let $\widehat g_S^{\,\mathrm{clean}}$ denote the gradient on the same
synthetic inputs with valid labels, and let
$g_Q=\mathbb E_{z\sim Q}[g^{\mathrm{clean}}(z)]$ be the population
gradient of the synthetic distribution $Q$. The update error decomposes as
\begin{equation}
\label{eq:update_error_decomposition}
\widehat g_S-g_{\mathcal T}
=
\underbrace{
\widehat g_S-\widehat g_S^{\,\mathrm{clean}}
}_{\text{supervision error}}
+
\underbrace{
\widehat g_S^{\,\mathrm{clean}}-g_Q
}_{\text{finite-sample error}}
+
\underbrace{
g_Q-g_{\mathcal T}
}_{\text{distribution mismatch}}.
\end{equation}
Accuracy reduces supervision error; diverse, non-redundant sampling reduces
finite-sample error; and similarity to the test domain reduces distribution
mismatch. In particular, if clean gradients are $L_g$-Lipschitz in a
multimodal task space,
\begin{equation}
\label{eq:wasserstein_gradient}
\|g_Q-g_{\mathcal T}\|
\le
L_g W_1(Q,P_{\mathcal T}),
\end{equation}
thus, minimizing update bias requires both local similarity and representative
coverage of the target distribution. Here, $P_{\mathcal T}$ denotes the empirical distribution over the test set, assigning probability $1/N$ to each example $(x_i, y_i)$, whose true answer $y_i$ is unknown.

\mypar{Learner-relative Complexity.}
Define the utility of a valid example $z$ as its reduction in target risk:
\begin{equation}
U_\theta(z)
=
R_{\mathcal T}(\theta)
-
R_{\mathcal T}
\left(
\theta-\eta g_\theta(z)
\right).
\end{equation}
Smoothness gives
\begin{equation}
\label{eq:sample_utility}
U_\theta(z)
\ge
\eta\langle g_{\mathcal T},g_\theta(z)\rangle
-
\frac{\beta\eta^2}{2}\|g_\theta(z)\|^2.
\end{equation}
Easy examples have negligible gradients, whereas excessively difficult or
misaligned examples can produce unstable or harmful updates. Useful
complexity therefore corresponds to a learner-relative region with positive
expected utility, which SIMIT approximates using confidence-band filtering.

\mypar{Constrained Local Distribution Design.}
Let $\pi_{\mathcal T}$ be a smoothed reference distribution induced by the unlabeled test queries, and let $\mathcal Z_{\mathrm{valid}}$ be the set of valid synthetic examples. Smoothing allows $\pi_{\mathcal T}$ to assign probability to related synthetic examples with new images and questions, beyond the exact test examples represented by $P_{\mathcal T}$.
We seek a useful synthetic distribution that remains
close to the target domain:
\begin{equation}
\label{eq:distribution_design}
\max_Q\;
\mathbb E_{z\sim Q}[U_\theta(z)]
-
\tau D_{\mathrm{KL}}(Q\|\pi_{\mathcal T})
\qquad
\text{s.t.}\quad
Q(\mathcal Z_{\mathrm{valid}})=1.
\end{equation}
Its solution is
\begin{equation}
\label{eq:optimal_synthetic_distribution}
q^\star(z)
\propto
\underbrace{\1_{z\in\mathcal Z_{\mathrm{valid}}}}_{\text{accuracy}}
\underbrace{\pi_{\mathcal T}(z)}_{\text{target similarity}}
\underbrace{\exp\!\left(U_\theta(z)/\tau\right)}_
{\text{learner-relative complexity}}.
\end{equation}
Moreover,
\begin{equation}
-\tau D_{\mathrm{KL}}(Q\|\pi_{\mathcal T})
=
\tau H(Q)
+
\tau\mathbb E_Q[\log\pi_{\mathcal T}(z)],
\end{equation}
revealing an explicit entropy term that discourages mode collapse while
keeping the generated data within the target domain. Thus, the principled
dataset is accurate, diverse but target-local, and informative for the
current learner, consistent with the ACE perspective \citep{zeng2026ACE}.

SIMIT operationalizes these terms through answer-first construction,
structured visual realization and verification for accuracy; diverse,
query-conditioned generation of both questions and images for local
coverage; and difficulty filtering for learner-relative complexity.

  \section{Adaptive Budget Allocation: Derivation}
\label{sec:appendix_adaptive_K}

We derive a criterion for selecting the number of demonstrations $K$ from a single zero-shot confidence estimate. The rule combines a calibrated query-level confidence with an accuracy-budget function, 
avoiding an online search over multiple values of $K$.


Using a single zero-shot confidence estimate keeps budget allocation inexpensive. We do not prefer alternative rollout-based confidence estimation methods since repeated rollouts incur additional decoding cost merely to decide how much computation to allocate \citep{han2024semantic}. Moreover, agreement is an imperfect proxy for correctness. Open-ended tasks admit multiple valid responses, so exact-match disagreement can reflect differences in wording or content; semantic clustering addresses wording variation but requires additional comparisons \citep{kuhn2023semantic}. Conversely, even on verifiable tasks, systematic errors can produce consistently repeated incorrect answers and misleadingly low entropy \citep{farquhar2024detecting}. Such agreement can obscure which queries need further practice. We therefore use calibrated confidence from the initial prediction as a lightweight allocation signal, avoiding repeated sampling and task-dependent answer aggregation.

\mypar{Setup.}
For a test query $q$, let $\hat a_0(q)$ denote the model's zero-shot prediction and let $C_K$ be the event that the prediction obtained with $K$ demonstrations is correct. Let $p_0\in(0,1)$ denote the calibrated confidence associated with $\hat a_0(q)$; calibration means that
\begin{equation}
P(C_0=1\mid p_0=p)=p
\end{equation}
over the query distribution. Our goal is to select the smallest $K\in\{0,\ldots,K_{\max}\}$ whose predicted correctness probability is at least $1-\varepsilon$.

\mypar{Log-odds and Required Evidence.}
For a binary event $C$ and evidence $z$, Bayes' rule can be written in odds form as
\begin{equation}
\frac{P(C\mid z)}{P(\neg C\mid z)}
=
\frac{P(C)}{P(\neg C)}
\frac{P(z\mid C)}{P(z\mid\neg C)}.
\end{equation}
Taking logarithms shows that evidence produces an additive shift in log-odds:
\begin{equation}
\operatorname{logit}P(C\mid z)
=
\operatorname{logit}P(C)
+
\ln\frac{P(z\mid C)}{P(z\mid\neg C)},
\qquad
\operatorname{logit}(p)\equiv\ln\frac{p}{1-p}.
\end{equation}
We therefore measure both the required and supplied improvement on the log-odds scale.

A correctness probability of $1-\varepsilon$ corresponds to the target log-odds
\begin{equation}
b_\varepsilon
\equiv
\operatorname{logit}(1-\varepsilon)
=
\ln\frac{1-\varepsilon}{\varepsilon}.
\end{equation}
Thus, a query with zero-shot confidence $p_0$ requires a log-odds improvement of
\begin{equation}
\label{eq:evidence-needed}
E_{\mathrm{req}}(q)
=
b_\varepsilon-\operatorname{logit}(p_0)
=
\ln\frac{1-\varepsilon}{\varepsilon}
+
\ln\frac{1-p_0}{p_0}.
\end{equation}
If $p_0\geq1-\varepsilon$, this requirement is already satisfied and $K=0$.

\mypar{Evidence Supplied by Demonstrations.}
We estimate the effect of a demonstration budget on a labeled calibration set $\mathcal D$. Let
\begin{equation}
\label{eq:accuracy-curve}
A(K)
=
P_{\mathcal D}(C_K=1)
=
\mathbb E_{(q,y)\sim\mathcal D}
\left[
\1_{\hat a_K(q)=y}
\right]
\end{equation}
denote the population accuracy obtained with $K$ demonstrations, measured using the same demonstration-generation and ordering procedure employed at test time. Define the corresponding population log-odds gain
\begin{equation}
\label{eq:supply-curve}
S(K)
=
\operatorname{logit}A(K)
-
\operatorname{logit}A(0).
\end{equation}

Because $A(K)$ is a population average whereas allocation is performed for an individual query, we adopt the proportional-odds approximation
\begin{equation}
\label{eq:proportional-odds}
\widehat p_K(q)
=
\sigma\left(
\operatorname{logit}(p_0)+S(K)
\right),
\end{equation}
where $\sigma(x)=(1+e^{-x})^{-1}$. This transfers the population log-odds improvement to each query. It is a moment-matching approximation: for a representative query with $p_0=A(0)$, it exactly recovers the measured population accuracy,
\begin{equation}
\widehat p_K(q)
=
\sigma\left(
\operatorname{logit}A(0)
+
\operatorname{logit}A(K)
-
\operatorname{logit}A(0)
\right)
=
A(K).
\end{equation}
For other confidence levels, \eqref{eq:proportional-odds} provides a calibrated population-level prediction rather than a per-query guarantee.

\mypar{Allocation Rule.}
Requiring $\widehat p_K(q)\geq1-\varepsilon$ in \eqref{eq:proportional-odds} gives
\begin{equation}
\operatorname{logit}(p_0)+S(K)
\geq b_\varepsilon.
\end{equation}
Combining this inequality with \eqref{eq:evidence-needed} yields
\begin{equation}
\label{eq:decision-rule}
\boxed{
K^\star(q)
=
\min\left\{
K\in\{0,\ldots,K_{\max}\}:
S(K)\geq
b_\varepsilon-\operatorname{logit}(p_0)
\right\}.
}
\end{equation}
Equivalently, solving for $p_0$ gives
\begin{equation}
\label{eq:threshold-form}
K^\star(q)
=
\min\{K:p_0\geq\tau_K\},
\qquad
\tau_K
=
\sigma\left(b_\varepsilon-S(K)\right).
\end{equation}
When $S(K)$ is non-decreasing, the thresholds satisfy
$\tau_0\geq\tau_1\geq\cdots\geq\tau_{K_{\max}}$. Consequently, more confident queries receive fewer demonstrations. If no threshold is reached, we assign the maximum budget $K_{\max}$.

\mypar{Closed-form Thresholds.}
Empirically, accuracy grows approximately logarithmically with the demonstration budget (Fig. \ref{fig:ablation_kshots}). We therefore assume
\begin{equation}
\label{eq:log-accuracy-curve}
A_K
=
\operatorname{clip}
\left(
A_1+B\ln K,
\eta,
1-\eta
\right),
\qquad
K=1,\ldots,K_{\max},
\end{equation}
where $A_1$ and $B\geq0$ are fitted on the calibration set and $\eta>0$ prevents degenerate probabilities. We use $\eta=10^{-3}$. The zero-shot accuracy $A_0=A(0)$ is a separate case because \eqref{eq:log-accuracy-curve} is undefined at $K=0$.

Define the population odds
\begin{equation}
O_0=\frac{A_0}{1-A_0},
\qquad
O_K=\frac{A_K}{1-A_K}.
\end{equation}
\Eqref{eq:supply-curve} then becomes
\begin{equation}
S(K)=\ln\frac{O_K}{O_0}.
\end{equation}
Substituting this expression into \eqref{eq:threshold-form} gives
\begin{align}
\tau_K
&=
\sigma\left(
\ln\frac{1-\varepsilon}{\varepsilon}
-
\ln\frac{O_K}{O_0}
\right) \notag\
\\&=
\boxed{
\frac{(1-\varepsilon)O_0}
{\varepsilon O_K+(1-\varepsilon)O_0}
}.
\label{eq:closed-form-tau}
\end{align}
For $K=0$, setting $O_K=O_0$ recovers $\tau_0=1-\varepsilon$. All thresholds are therefore computed once from $(\varepsilon,A_0,A_1,B)$ and stored in a lookup table. They are our hyperparameters. We tune them on our validation sets. And, at test time, allocation requires only the 
zero-shot confidence $p_0$ and a comparison against this table.

\begin{algorithm}[ht]
\caption{Closed-form threshold precomputation}
\label{alg:threshold-precomputation}
\begin{algorithmic}[1]
\REQUIRE $\varepsilon,A_0,A_1,B,K_{\max}$, clip bound $\eta$
\STATE $A_0\gets\operatorname{clip}(A_0,\eta,1-\eta)$
\STATE $O_0\gets A_0/(1-A_0)$; \quad $\tau_0\gets1-\varepsilon$
\FOR{$K=1$ \TO $K_{\max}$}
\STATE $A_K\gets
\operatorname{clip}(A_1+B\ln K,\eta,1-\eta)$
\STATE $O_K\gets A_K/(1-A_K)$
\STATE $\tau_K\gets
(1-\varepsilon)O_0/
\big(\varepsilon O_K+(1-\varepsilon)O_0\big)$
\ENDFOR
\RETURN $\tau_0,\ldots,\tau_{K_{\max}}$
\end{algorithmic}
\end{algorithm}

\section{Confidence-based Difficulty Filtering}
\label{sec:appendix_confidence_filter}

The preceding framework retains examples with positive learner-relative
utility. Because this utility is unavailable at test time, SIMIT estimates it
from the solver's confidence in the intended answer. For a verified synthetic
example $z=(\widetilde I,\widetilde q,\widetilde a)$, define
\begin{equation}
    p
    =
    p_\theta(
    \widetilde a
    \mid
    \widetilde I,\widetilde q
    ).
\end{equation}
Let $C\in\{0,1\}$ indicate that the assigned answer is valid for the realized
image, and define the post-verification validity probability
\begin{equation}
    q(p)
    =
    P(C=1\mid p,\text{verified}).
\end{equation}

Let $v(p)\ge0$ denote the potential learning value of an example with
confidence $p$. A valid example provides benefit $\alpha v(p)$, and an invalid
example incurs harm $\beta v(p)$. Retaining an example also incurs a cost $\lambda>0$, accounting for additional adaptation or inference computation and limited context capacity. Even when valid, an uninformative example may displace more useful demonstrations or introduce interference, thereby degrading ICL performance. Thus, the conditional expected utility is
\begin{equation}
\label{eq:confidence_utility}
    U(p)
    =
    v(p)
    \left[
    \alpha q(p)-\beta\bigl(1-q(p)\bigr)
    \right]
    -
    \lambda.
\end{equation}
The Bayes-optimal confidence-only admission rule is
\begin{equation}
\label{eq:optimal_confidence_filter}
    \delta^\star(p)
    =
    \1_{U(p)>0}.
\end{equation}
Indeed, conditioned on $p$, retaining the example contributes $U(p)$ while
rejecting it contributes zero, thus the decision is optimal pointwise.

For a concrete instantiation, assume calibrated confidence,
$q(p)=p$, and use $v(p)=1-p$, reflecting the vanishing learning signal of
high-confidence examples. \Eqref{eq:confidence_utility} becomes
\begin{equation}
\label{eq:quadratic_confidence_utility}
    U(p)
    =
    (1-p)
    \left[
    \alpha p-\beta(1-p)
    \right]
    -
    \lambda,
\end{equation}
which is a concave quadratic. Provided
\begin{equation}
    \lambda
    <
    \frac{\alpha^2}{4(\alpha+\beta)},
\end{equation}
its positive region is the interval
\begin{equation}
\label{eq:confidence_band}
    T_{\mathrm{low}}
    <
    p
    <
    T_{\mathrm{high}},
\end{equation}
where
\begin{equation}
\label{eq:confidence_thresholds}
\boxed{
    T_{\mathrm{low, high}}
    =
    \frac{
    \alpha+2\beta
    \mp
    \sqrt{\alpha^2-4\lambda(\alpha+\beta)}
    }{
    2(\alpha+\beta)
    }
}
\end{equation}
Thus, very low-confidence examples are rejected because their intended
answers are insufficiently reliable or beyond the learner's current
frontier, while very high-confidence examples are rejected because their
learning value is too small to justify their cost. SIMIT implements this
optimal confidence-only policy through the band-pass filter
$p\in[T_{\mathrm{low}},T_{\mathrm{high}}]$.

\section{Consensus-based LFRL under Single-Query Adaptation}
\label{sec:appendix_lfrl_limitation_single_sample_adaptation}


We formalize a limitation of consensus-based rewards: under an exact
KL-regularized policy update, population consensus cannot correct an
incorrect initial majority in the episodic (isolated) single-query
adaptation setting. Following the distribution-space formulation
of \citet{he2026howfar}, we establish this result without assuming that
the majority remains fixed, and extend it to finite-rollout consensus.

\mypar{Episodic Single-query Adaptation.}
Consider a fixed unlabeled query $x$. The model is reset to its initial
parameters before each query, and all rollouts and updates within an
episode concern $x$. Thus, a single-query episode may contain multiple
responses and optimization steps, but no transfer from other queries.
Let $\mathcal{A}$ be a fixed set of $K \geq 2$ answer classes, and let
$p_t(a) := p_t(a \mid x)$ denote the probability that a response belongs
to class $a$ after $t$ updates. Assume $p_0(a)>0$ for every
$a \in \mathcal{A}$ and a unique initial modal answer
$m_0 = \arg\max_a p_0(a)$. Throughout, \emph{majority} denotes the most
probable answer class; its probability need not exceed $0.5$.
Let $a^\star$ denote the correct answer class.

\mypar{Consensus Rewards and Policy Update.}
Writing $m_t \in \arg\max_a p_t(a)$, the population versions of
majority-vote and frequency-based rewards are
\begin{equation}
    r_t^{\mathrm{MV}}(a) = \1_{a=m_t},
    \qquad
    r_t^{\mathrm{F}}(a) = p_t(a).
    \label{eq:lfrl_consensus_rewards}
\end{equation}
These correspond to the core consensus rewards of TTRL and MM-UPT, and of TTRV and EMPO,
respectively; for semantic consensus, $\mathcal{A}$ represents fixed
semantic classes. Both satisfy
\begin{equation}
    r_t(m_t) \geq r_t(a)
    \qquad \forall a \in \mathcal{A}.
    \label{eq:lfrl_reward_order}
\end{equation}
Subtracting an answer-independent baseline or multiplying rewards by a
positive scalar preserves this ordering. TTRV additionally computes an entropy penalty, which measures the diversity of the answers sampled for a query. It subtracts the same
penalty from every response's reward, so the most frequent answer
still receives the highest reward. This penalty also cancels when
GRPO subtracts the group's mean reward.

We analyze the exact distribution-space update with a fixed reference
\begin{equation}
    p_{t+1}
    =
    \arg\max_{p \in \Delta(\mathcal{A})}
    \left\{
        \sum_{a \in \mathcal{A}} p(a) r_t(a)
        -
        \beta_t D_{\mathrm{KL}}(p \,\|\, p_0)
    \right\},
    \qquad \beta_t > 0,
    \label{eq:lfrl_kl_update}
\end{equation}
where $\Delta(\mathcal{A})$ is the probability simplex and the reward
$r_t$ is held fixed during the update. The KL reference is the initial
policy $p_0$, and $\beta_t$ controls regularization rather than the
optimizer's learning rate.
The KL penalty compares the adapted model with the fixed initial
model $p_0$, while new responses are sampled from the current model
$p_t$ at each adaptation round \citep{shao2024deepseekmath}.
\Eqref{eq:lfrl_kl_update} is an idealized policy-level objective,
not the clipped, token-level GRPO surrogate. Since rewards are constant
within each answer class, the same class-level update follows by
optimizing over complete responses and summing their probabilities
within each class.

\mypar{Group-relative Advantage Normalization.}
For a fixed rollout group, GRPO uses advantages
$A_t(a)=(r_t(a)-\mu_t)/s_t$, where $\mu_t$ is the group's mean reward
and $s_t>0$ is its stabilized reward standard deviation. These
statistics are held fixed during the policy update. Since
$A_t(m_t)-A_t(a)=(r_t(m_t)-r_t(a))/s_t$,
this normalization preserves \eqref{eq:lfrl_reward_order}.
More explicitly, replacing $r_t$ by $A_t$ in the exact KL objective
with coefficient $\beta>0$ on the KL term gives
\begin{equation}
    p_{t+1}(a)
    \propto p_0(a)\exp\left(\frac{A_t(a)}{\beta}\right)
    \propto p_0(a)\exp\left(\frac{r_t(a)}{\beta s_t}\right).
    \label{eq:lfrl_advantage_normalization}
\end{equation}
The factor $\exp(-\mu_t/(\beta s_t))$ is the same for every answer,
thus it cancels when the probabilities are rescaled to sum to one.
Using GRPO-normalized advantages therefore gives the same exact
update as \eqref{eq:lfrl_kl_update} with $\beta_t=\beta s_t$,
and the following results apply to both formulations.

\begin{proposition}[Preservation of the Initial Majority]
\label{prop:lfrl_mode_preservation}
Under \eqref{eq:lfrl_reward_order} and \eqref{eq:lfrl_kl_update},
for every $T \geq 0$ and every $a \neq m_0$,
\begin{equation}
    \frac{p_T(m_0)}{p_T(a)}
    \geq
    \frac{p_0(m_0)}{p_0(a)}
    > 1.
    \label{eq:lfrl_mode_preservation}
\end{equation}
Consequently, $\arg\max_a p_T(a)=m_0$ for every $T$.
If $m_0 \neq a^\star$, adaptation cannot correct the modal answer.
Thus, selecting the most probable answer
after consensus-based LFRL adaptation achieves exactly the same expected accuracy as
majority voting with the initial model.
\end{proposition}

\begin{proof}
Define
\begin{equation}
    q_t(a)
    =
    \frac{p_0(a)\exp\bigl(r_t(a)/\beta_t\bigr)}{Z_t},
    \qquad
    Z_t
    =
    \sum_{b \in \mathcal{A}}
    p_0(b)\exp\bigl(r_t(b)/\beta_t\bigr).
    \label{eq:lfrl_exponential_update}
\end{equation}
The objective in \eqref{eq:lfrl_kl_update} equals
$\beta_t\bigl(\log Z_t-D_{\mathrm{KL}}(p\,\|\,q_t)\bigr)$.
Nonnegativity of KL divergence therefore gives the unique optimizer
$p_{t+1}=q_t$.


For the first update and any $a \neq m_0$,
\begin{equation}
    \log \frac{p_1(m_0)}{p_1(a)}
    =
    \underbrace{
        \log \frac{p_0(m_0)}{p_0(a)}
    }_{>0}
    +
    \underbrace{
        \frac{r_0(m_0)-r_0(a)}{\beta_0}
    }_{\geq 0}
    > 0.
    \label{eq:lfrl_log_odds}
\end{equation}
The first term is positive because $m_0$ is the initial majority,
and the second is nonnegative because consensus assigns it a maximal
reward. Thus, $m_0$ remains the most probable answer after the first
update, and its odds against every other answer are at least their
initial values.

Since every update uses the same reference $p_0$, the same argument
applies at each later round whenever the current majority is $m_0$.
Induction therefore proves both preservation of the initial majority
and the odds bound in \eqref{eq:lfrl_mode_preservation}.

Also, if all responses in a group receive the same reward, their advantages
are zero; thus they provide no learning signal that favors one answer
over another. In this case, we allow either skipping the update,
which leaves $p_t$ unchanged, or minimizing only the KL penalty,
whose exact solution is $p_0$. Both cases also preserve the initial majority.
\end{proof}

For the majority-vote reward in \eqref{eq:lfrl_consensus_rewards},
only the majority answer receives a positive reward. If this answer
is incorrect ($a^\star \neq m_0$), the update reduces the probability
of the correct answer below its initial value. Specifically,
\eqref{eq:lfrl_exponential_update} gives
\begin{equation}
    p_{t+1}(a^\star)
    =
    \frac{p_0(a^\star)}
    {1+p_0(m_0)\bigl(\exp(1/\beta_t)-1\bigr)}
    < p_0(a^\star).
    \label{eq:lfrl_wrong_majority}
\end{equation}
Thus, the probability of the correct answer is lower than its initial
value in this case.
With a fixed coefficient and unnormalized majority-vote rewards,
repeated exact optimization returns the same distribution after the
first update. Thus, the fixed-reference model implies suppression
of a correct minority, but not cumulative collapse merely from
increasing the number of updates.

\mypar{Finite-rollout Consensus.} So far, rewards have been defined using the model's exact answer probabilities. In practice, rewards are computed from finitely many
sampled responses, whose most frequent answer may differ from the
model's most probable answer. We now bound the probability that
this sampling variability changes the model's majority during adaptation.

At iteration $t$, draw $N$ conditionally independent answers
from $p_t$, let $\widehat p_t$ be their empirical distribution, and use
either reward in \eqref{eq:lfrl_consensus_rewards} with $p_t$ replaced
by $\widehat p_t$. Define the initial probability gap,
$
    \delta_0
    =
    p_0(m_0)-\max_{a \neq m_0}p_0(a)
    > 0
$.

\begin{corollary}[Finite-rollout Stability]
\label{cor:lfrl_finite_rollouts}
Under the empirical consensus rewards above and the exact update
\eqref{eq:lfrl_kl_update}, the probability that the model's most
probable answer changes during $T$ adaptation rounds satisfies
\begin{equation}
    P \left(
        \exists\, t \in [1, T] : 
        \arg\max_a p_t(a) \neq m_0
    \right)
    \leq
    \min\left\{
        1,\,
        T(K-1)\exp\left(-\frac{N\delta_0^2}{2}\right)
    \right\}.
    \label{eq:lfrl_finite_rollout_bound}
\end{equation}
\end{corollary}

\begin{proof}
If $m_0$ receives more votes than every other answer, both consensus
rewards assign it the highest reward. As long as this happens in
every round, the exact updates preserve $m_0$ as the model's most
probable answer. A change therefore requires a round in which
another answer ties with or receives more votes than $m_0$.
We bound the probability of the first such round.

Let $\mathcal{H}_t$ denote the sampled responses from all rounds
before $t$, and consider any such history in which $m_0$ received
more votes than every other answer in each round.
The preceding updates then favored $m_0$; thus \eqref{eq:lfrl_exponential_update} gives
\[
    \frac{p_t(a)}{p_t(m_0)}
    \leq
    \frac{p_0(a)}{p_0(m_0)}
    \qquad \forall a \neq m_0.
\]
Summing these inequalities over the competing answers and using
$\sum_a p_t(a)=1$ gives $p_t(m_0)\geq p_0(m_0)$.
Thus, the probability gap between $m_0$ and each competitor remains
at least as large as its initial value:
\begin{equation}
    p_t(m_0)-p_t(a)
    =
    p_t(m_0)\left(1-\frac{p_t(a)}{p_t(m_0)}\right)
    \geq p_0(m_0)-p_0(a)
    \geq \delta_0.
\end{equation}

Now fix a competing answer $a\neq m_0$. For each sampled answer
$A_i$, define
\[
    D_i=\1_{A_i=m_0}-\1_{A_i=a}.
\]
This variable equals $1$ for a vote for $m_0$, $-1$ for a vote for
$a$, and $0$ otherwise. Conditional on $\mathcal{H}_t$, the $D_i$
are independent, lie in $[-1,1]$, and have mean
$p_t(m_0)-p_t(a)\geq\delta_0$.
For $a$ to tie with or outvote $m_0$, their sample average must
nevertheless be at most zero. Hoeffding's inequality bounds the
probability of this deviation:
\begin{equation}
    \begin{aligned}
        P\left(
            \widehat p_t(a)\geq\widehat p_t(m_0)
            \,\middle|\,\mathcal{H}_t
        \right)
        &=
        P\left(
            \frac{1}{N}\sum_{i=1}^{N}D_i\leq 0
            \,\middle|\,\mathcal{H}_t
        \right) 
        \leq
        \exp\left(-\frac{N\delta_0^2}{2}\right).
    \end{aligned}
\end{equation}

This bound holds for every history with no earlier sampled tie
or reversal. A union bound over the $K-1$ competing answers and
the $T$ possible rounds therefore bounds the probability of the
first such event by $T(K-1)\exp(-N\delta_0^2/2)$.
Since the model's majority cannot change without such an event,
capping this bound at one proves
\eqref{eq:lfrl_finite_rollout_bound}.
\end{proof}

\begin{corollary}[Stability of a Concentrated Majority]
\label{cor:lfrl_concentrated_majority}
Under the same empirical rewards and exact policy update, suppose
$p_0(m_0) \geq \rho > 0.5$. Let $b_N(\rho)$ denote the probability
that an answer with probability $\rho$ receives at most half the
votes among $N$ independent responses:
\begin{equation}
    b_N(\rho)
    =
    \sum_{j=0}^{\lfloor N/2 \rfloor}
    \binom{N}{j}\rho^j(1-\rho)^{N-j}.
    \label{eq:lfrl_binomial_tail}
\end{equation}
Then, regardless of the number of answer classes, the probability
that $m_0$ remains the model's most probable answer throughout
$T$ adaptation rounds satisfies
\begin{equation}
    P \left(
        \arg\max_a p_t(a)=m_0, \quad
        \forall t \in \{0,\ldots,T\}
    \right)
    \geq
    \bigl(1-b_N(\rho)\bigr)^T.
    \label{eq:lfrl_concentrated_stability}
\end{equation}
\end{corollary}

\begin{proof}
If $m_0$ receives more than half the votes, it necessarily receives
more votes than any competing answer, regardless of how the
remaining votes are divided. Both consensus rewards therefore
assign it the highest reward.

Consider any history $\mathcal{H}_t$ in which this has happened
in every preceding round. As shown in the preceding proof, these
updates preserve $m_0$ as the most probable answer and ensure
$p_t(m_0)\geq p_0(m_0)\geq\rho$.
Conditional on this history, the number of votes $V_t$ for $m_0$
in the next round follows
$\operatorname{Binomial}(N,p_t(m_0))$.
Increasing the probability of $m_0$ makes receiving more than half
the votes more likely. Since $p_t(m_0)\geq\rho$, we obtain
\begin{equation}
    P\left(
        V_t > N/2 \,\middle|\, \mathcal{H}_t
    \right)
    \geq 1-b_N(\rho).
\end{equation}

The first round thus has a strict majority with probability at least
$1-b_N(\rho)$. Every later round has the same lower bound,
conditional on strict majorities in all earlier rounds.
The probability of a strict majority in every round is therefore
at least $\bigl(1-b_N(\rho)\bigr)^T$.
On this event, every update preserves $m_0$, proving the result.
Independence across rounds is not required because the bound
holds conditional on every preceding history of strict majorities.
\end{proof}

\mypar{Interpretation.}
Consensus rewards favor agreement with the model's current majority,
even when that majority is incorrect. Under the stated updates with
population rewards, this ordering prevents correction of the modal
answer; finite sampling preserves the result with the probability
quantified above. For population majority-vote rewards, sharpening can
only improve sampled accuracy over the initial policy when the initial
majority is correct. But with population rewards, the initial
population-majority prediction is preserved, making its accuracy an
invariant ceiling for population-majority decoding under the analyzed
updates.
Thus, the correction is highly improbable for typical values of $N$.
Under the analyzed dynamics, sufficiently concentrated
incorrect majorities are preserved with high probability even with
finite rollout budgets. For example, an initial modal probability of
at least \(0.75\), with 32 rollouts per round, guarantees at least
\(98\%\) probability of retaining that answer throughout ten adaptation
rounds (Corollary~\ref{cor:lfrl_concentrated_majority}).

\mypar{Scope.} 
Note that this result is valid for the episodic isolated single-query adaptation setting. Shared neural parameters across samples and finite optimization steps do not have to obey
\eqref{eq:lfrl_kl_update}, even within a single-query episode. Thus, these methods can outperform the initial majority voting ceiling in the test-set adaptation setting due to such cross-sample capability transfer mechanisms.
Also note that our proofs rely on an idealized, exact KL-regularized policy update (\Eqref{eq:lfrl_kl_update}). In practice, LFRL frameworks use clipped surrogate objectives (e.g., GRPO/PPO) to constrain the policy update step. However, this clipping operation just enforces the preservation of the initial modal answer even more rigidly, by strictly limiting the maximum probability multiplier. Thus, our core claims from the exact theoretical proofs also apply to the empirical clipped implementations.

  \section{Implementation Details}
\label{sec:appendix_implementation_details}

This appendix details the implementation of SIMIT, the baselines, and our evaluation protocols. Throughout synthesis and adaptation, BAGEL-7B is the only learned model: it proposes triplets, selects realization routes, writes rendering specifications, generates natural images, verifies its outputs, estimates confidence, and learns from the retained examples. Parsers, renderers, a sandboxed interpreter, and a headless browser execute model-produced specifications. We report runtimes for a single NVIDIA H100 GPU. All runtimes reported in Tab.~\ref{tab:main_comparison}, \ref{tab:ablation_components} and \ref{tab:ablation_fix_qi} are end-to-end runtimes including data synthesis, training (for SIMIT-FT), and evaluation.   Representative prompts are provided below, with repetitive content omitted. The complete codebase, including all prompts, will be released. 

\subsection{Implementation Details of SIMIT}
\label{sec:appendix_impl_simit}

\mypar{Decoding Configuration.}
For text generation, we use BAGEL's direct-answer mode throughout the data synthesis pipeline, except for the verification critic, which uses thinking mode. 
Most calls use greedy decoding initially and higher-temperature sampling on retries, since greedy decoding with nearly unchanged context often reproduces the same failure.


\mypar{Adaptive Budget Allocation.}
ABA uses one confidence estimate per test query. We greedily decode the zero-shot answer $\hat{a}_0$ using the benchmark's standard prompt and compute the geometric mean of its token probabilities:
\begin{equation}
    p_0 = \exp\left(
        \frac{1}{|\hat{a}_0|}
        \sum_{t=1}^{|\hat{a}_0|}
        \log p_\theta\bigl(\hat{a}_{0,t}\mid\hat{a}_{0,<t},I,q\bigr)
    \right).
\end{equation}
Length normalization reduces dependence on answer length, allowing the same statistic to be used for short answers and captions. Using the thresholds precomputed by Algorithm~\ref{alg:threshold-precomputation}, we select
\begin{equation}
    K^\star(q)=\min\bigl\{K\in\{0,\ldots,K_{\max}\}:p_0\geq\tau_K\bigr\},
\end{equation}
with $K_{\max}$ as the fallback if no threshold is met. Thus, ABA requires one greedy decoding pass per query, without repeated sampling or answer aggregation.

\mypar{Triplet Synthesis.}
Given the test query, BAGEL proposes $K$ triplets in a single call using the prompt below. A regular-expression parser accommodates common formatting deviations, including Markdown emphasis and enumeration. If fewer than $K$ triplets are parsed, we repeat the request with an explicit format reminder, allowing up to three attempts and retaining the largest parsed set. The initial temperature is $0.2$, increasing to $0.7$ on retries. When the slot-filling procedure requires additional candidates, it requests one new triplet at a time using the same prompt. The ablation without diversity-encouraging prompting (DP; Tab.~\ref{tab:ablation_components}) removes the \emph{``Diversity Rules''} block and the word ``diverse'' in the first sentence, leaving the rest unchanged.

\begin{prompt}[Diversity-encouraging Prompting for Triplet Synthesis]
Generate exactly \{K\} diverse visual question-answering training examples.\\\\Each example must have:\\- A visual question (about any visual concept — NOT tied to the given image)\\- A short correct answer\\- A detailed description of a new image where your question and answer are correct, your description should be detailed and specific, one should be able to visualize the image perfectly from the description\\\\DIVERSITY RULES:\\- Cover as many different visual reasoning types as possible across the \{K\} examples: counting, color/attribute, spatial position, object recognition, action/activity, text/OCR, yes/no judgment, comparison, scene type, etc.\\- Use the provided image only as inspiration for difficulty level and visual domain. The generated examples do NOT need to match the provided image.\\- Answers must be short (1-5 words).\\\\Output format (blank line between examples):\\\\Question: [your visual question]\\Answer: [short correct answer]\\Description: [detailed description of a new image where question+answer are correct]\\\\Question: [another diverse visual question]\\Answer: [short correct answer]\\Description: [description for this example]\\\\... (exactly \{K\} Question/Answer/Description triples)
\end{prompt}

\mypar{Visualization Routing.}
Each description is assigned to one of 17 predefined visualization categories. 
BAGEL classifies the textual descriptions using greedy decoding. The routing prompt defines the categories, distinguishes commonly confused cases, and provides few-shot examples.

\begin{prompt}[Visualization Routing Prompt]
You are a precise request classifier. Decide which of seventeen categories 
a visualization request belongs to:
\\~\\
NATURAL: a realistic photo, artistic illustration, or rendered scene of real-world subject 
matter -- people, BODY PARTS (feet, hands, eyes, organs, bones), animals, PLANTS (leaves, 
flowers, trees, roots, seeds), objects, landscapes, food, buildings, vehicles, or any scene 
meant to look like a photograph or painting rather than a technical/schematic figure. NATURAL is correct whenever the content is about real-world things 
rather than abstract geometric shapes.
\\
DIAGRAM: a flowchart, process, or system/architecture diagram -- discrete labeled 
boxes/steps connected by arrows showing a sequence, process, or hierarchical flow.
\\
GRAPH: a graph-theory graph -- nodes and edges with no process/flow meaning, often with 
weights, may be undirected. A list of pairwise relationships between named entities 
(follows, knows, is friends with) is GRAPH -- it describes relationships, not a process.
\\
$\lbrack$... Many Category Descriptions Omitted ...$\rbrack$
\\~\\
Answer with exactly one word: NATURAL, DIAGRAM, GRAPH, MOLECULE, CIRCUIT, VECTOR, VENN, TABLE, 
SCENE, PUZZLE, GEOMETRY, FIGURE, MERMAID, SVG, VEGALITE, HTML, or HTML\_COMPOSITE.
\\~\\
Request: "A golden retriever puppy playing in a field of sunflowers at sunset, photorealistic."\\
Answer: NATURAL
\\
Request: "Draw a flowchart of order processing: receive order, charge card, ship item."\\
Answer: DIAGRAM
\\
Request: "A bar chart showing annual revenue for five companies, ranked from highest to lowest, with dollar values labeled on each bar."\\
Answer: FIGURE
\\
Request: "Create a heatmap of website activity by day of week (rows) and hour of day (columns), colored by number of visits."\\
Answer: VEGALITE
\\
Request: "Draw a weighted graph with nodes A, B, C and edges A-B (5), B-C (3), A-C (2)."\\
Answer: GRAPH
\\
Request: "Show the molecular structure of aspirin."\\
Answer: MOLECULE
\\
Request: "Show a system diagram with a Client, Server, and Database connected by arrows."\\
Answer: DIAGRAM
\\
Request: "Draw a circle with radius 5 and mark the center."\\
Answer: GEOMETRY
\\
Request: "An infographic section with a headline stating '63\% of executives said X', a supporting icon, and a short caption."\\
Answer: HTML
\\
Request: "Three foot type images labeled A, B, C: A shows a high arch, B shows a flat foot, C shows a normal arch."\\
Answer: HTML\_COMPOSITE
\\$\lbrack$... Many Few-Shot Examples Omitted ...$\rbrack$\\
Request: "\{request\}"\\
Answer:
\end{prompt}


\mypar{Skill Library.}
Descriptions routed to \emph{``Natural''} category use BAGEL's native image-generation head at $400\times400$ resolution. Each remaining category uses a \emph{skill} that pairs a specification prompt with a deterministic rendering backend (Tab.~\ref{tab:skill_library}). Eleven categories use declarative representations: a line-based easy-to-use domain-specific language or, for molecules, SMILES. These representations express semantic content, such as nodes and edges, set members, or points and annotations, while delegating layout, geometry, and derived quantities to the backend. For example, the geometry backend computes side lengths, angles, areas, and slopes from coordinates, and the scene backend uses rejection sampling to satisfy overlap constraints. The other five categories support more open-ended content through standard formats executed by established renderers. Each skill prompt specifies the required format 
and provides few-shot examples, as illustrated by the diagram prompt below.

We wrote generic prompts following the best prompting practice: a role description, an explicit output format, and a few format examples. We did not optimize them against benchmark scores. Revisions only fixed format and routing failures observed on validation data, and evaluation splits were not inspected before the design was frozen. The same prompts and skills are used for all benchmarks, adaptation modes, and models.
Following the best prompting practices enables our method to work even for very small models. 
But skills encode rendering formats, not task knowledge or human prior: the model decides what to depict and which answer to assign, and routing examples only choose among 17 renderers. Prompt optimization could further improve results, but our goal is to measure the self-improvement capability of UMMs with a fixed, generic pipeline.

\begin{table}[h]
\centering
\small
\caption{Image realization routes. Natural images use BAGEL's native generation head; structured and composite visuals use the listed specifications and backends.}
\label{tab:skill_library}
\begin{tabular}{lll}
\toprule
\textbf{Category} & \textbf{Intermediate Specification} & \textbf{Rendering Backend} \\
\midrule
Natural & Textual description & BAGEL native image generation \\
Diagram & \texttt{NODE}/\texttt{EDGE} lines, optional edge labels & Top-down auto-layout (Matplotlib) \\
Graph & \texttt{NODE}/\texttt{EDGE} lines, (un)directed, weighted & NetworkX (tree or spring layout) \\
Molecule & SMILES string and name & RDKit \\
Circuit & Series \texttt{COMPONENT} and \texttt{PARALLEL} lines & Schemdraw \\
Vector & Labeled $(x, y)$ components, optional body & Arrow renderer (Matplotlib) \\
Venn & Two set names and region contents & Two-set renderer (Matplotlib) \\
Table & Title, column headers, rows & Table renderer (Matplotlib) \\
Scene & Shape groups: type, color, count, overlap & Rejection-sampling placement \\
Puzzle & Layout type, panels, answer, transformation & Panel-grid renderer \\
Geometry & Points, polygons, circles, annotations & Geometry engine (Matplotlib) \\
HTML\_Composite & Title and labeled panel descriptions & Recursive panel synthesis + Chromium \\
\midrule
Figure & Python script (Matplotlib, NumPy) & Python interpreter \\ 
SVG & SVG markup & CairoSVG \\
Mermaid & Mermaid code & Mermaid CLI \\
Vega-Lite & Vega-Lite v5 JSON specification & vl-convert \\
HTML & HTML with inline CSS & Headless Chromium \\
\bottomrule
\end{tabular}
\end{table}

\begin{prompt}[Diagram Skill Prompt]
You are a precise diagram-structure extractor. Given a description 
of a diagram (a flowchart, process, or architecture), you identify every step (node) and 
every connection (edge) between them -- nothing more. You never draw anything or write any 
code; you only output the structure, exactly in this format and nothing else (no prose):
\\~\\
NODE: <short label>\\
EDGE: <source label> -> <destination label>\\
EDGE: <source label> -> <destination label> : <label, e.g. yes/no>
\\~\\
List every node once, in the order they are first mentioned. Then list every connection 
between them as an EDGE line, adding " : <text>" only when the connection has its own 
label (like a yes/no branch). A node can have more than one incoming or outgoing edge.
\\~\\
Example 1.\\
Request: Draw a 3-step process: Start, then Validate Input -- if invalid, go back to Start, if valid go to Finish.\\
NODE: Start\\
NODE: Validate Input\\
NODE: Finish\\
EDGE: Start -> Validate Input\\
EDGE: Validate Input -> Finish : valid\\
EDGE: Validate Input -> Start : invalid\\
\\~\\
Example 2.\\
Request: Two services, Auth Service and Order Service, both write to a shared Database. Order Service also calls a Payment Gateway.\\
NODE: Auth Service\\
NODE: Order Service\\
NODE: Database\\
NODE: Payment Gateway\\
EDGE: Auth Service -> Database\\
EDGE: Order Service -> Database\\
EDGE: Order Service -> Payment Gateway
\\~\\
Now extract from this request:\\
\{request\}
\end{prompt}




\mypar{Validation and Repair.}
Skill outputs undergo three deterministic checks before visual verification. First, a parser checks the specification's syntax, such as the presence of a \texttt{NODE} line or consistency between table headers and row lengths. Second, static checks detect known failure patterns, including non-ASCII identifiers in code and deprecated plotting arguments. Third, the renderer rejects specifications it cannot execute, including invalid SMILES strings, malformed JSON, Mermaid syntax errors, and references to undefined geometric points.

Failed specifications are returned to BAGEL with the parser or renderer error, allowing up to two repair attempts per realization. Code-writing skills use temperature $0.5$ for repair. Minor output corruption is corrected deterministically; for example, lines containing stray non-English characters are removed when they constitute only a small fraction of the output.


\begin{prompt}[Diagram Skill Repair Prompt after Parser Feedback]
Your previous answer for this request could not be parsed as a 
NODE:/EDGE: list (\{reason\}). Output ONLY lines in the form "NODE: <label>" or 
"EDGE: <source> -> <destination>" (optionally " : <label>"), nothing else.
\\~\\
Request: \{request\}
\end{prompt}

\mypar{Verification.}
BAGEL then acts as a visual critic for every realized image, whether generated natively or rendered. Given the image $\tilde I_k$ and its description $\tilde d_k$, the critic evaluates their consistency and assigns a rubric-based score from $0$ to $100$. Images with $\mathrm{SCORE}\geq50$ pass verification.


When verification fails, BAGEL revises the generation description using the critic's feedback and regenerates the image. The synthetic question, intended answer, and reference description remain fixed: every new image is judged against the \emph{original} description. Each triplet receives up to five generation-verification rounds during repair. Triplets that never pass, or for which no critique is obtained, are discarded.


\begin{prompt}[Verification Prompt]
You are a strict quality-control critic scoring an AI-generated image against this brief: "\{request\}"\\\\
Go through the brief clause by clause. For each specific, checkable detail (counts, colors, species, objects, setting, text, spatial relations), state whether the image satisfies it, with no tolerance for approximation. List every discrepancy you find, however small.\\\\
Then assign a single quality SCORE from 0 to 100 using this rubric:
\begin{quote}
  90-100: every checkable detail in the brief is satisfied, no discrepancies\\
  70-89:  matches the brief overall, only minor/cosmetic discrepancies (e.g. slightly off shade, imprecise count in a busy/cluttered scene)\\
  40-69:  matches part of the brief but has at least one clear factual discrepancy (wrong count, wrong object, wrong text, wrong spatial relation)\\
  0-39:   mostly or entirely fails to match the brief
\end{quote}
End your answer with exactly one line: `SCORE: <integer 0-100>'.
\end{prompt}

\begin{prompt}[Description Revision Prompt after a Failed Verification]
This image description was used to generate an image: "\{request\}"\\
A critic reviewed the generated image and found these problems:\\
"\{critique\}"\\\\
Rewrite the description, keeping everything that was already correct, but adding or changing whatever is needed so the next generation avoids these exact problems. Output ONLY the revised description text, nothing else.
\end{prompt}


\mypar{Breadth-First Slot Filling.}
We schedule realization attempts breadth-first, giving candidates an initial opportunity before spending further computation on individual failures. Each query has $K$ slots, filled sequentially by accepted triplets, and a total budget of $4K$ realization attempts. In the first phase, each candidate receives one attempt without repair or description revision. Failed candidates enter a retry pool; once the initial $K$ candidates are exhausted, we request fresh triplets. This phase uses at most $3K$ attempts, reserving the remaining budget for repair. Subsequent phases cycle through the retry pool, giving each candidate one attempt per round with the full repair and revision budget. Processing stops when all slots are filled or the total budget is exhausted; queries with unfilled slots retain their accepted triplets. Across generation logs from four benchmarks, $32\%$-$80\%$ of eventually accepted triplets passed on their first attempt. Breadth-first scheduling reduced model calls per filled slot by $1.9$--$3.5\times$ relative to depth-first scheduling.


\mypar{Difficulty Filtering.}
For each verified triplet $(\tilde{I},\tilde{q},\tilde{a})$, we estimate the model's confidence in the intended answer $\tilde a_k$ by teacher forcing:
\begin{equation}
    c = \exp\left(
        \frac{1}{|\tilde{a}|}
        \sum_{t=1}^{|\tilde{a}|}
        \log p_\theta\bigl(\tilde{a}_t\mid\tilde{a}_{<t},\tilde{I},\tilde{q}\bigr)
    \right).
\end{equation}
This requires one forward pass, using the synthetic question without additional instructions. DF retains examples with $c\in[T_{\text{low}},T_{\text{high}}]$. Candidates retain their generation order: DF removes those outside the confidence band, and ABA selects the first $K^\star(q)$ survivors. For a fixed ordered candidate sequence, the corresponding online rule stops when $K^\star(q)$ examples pass DF or $K_{\max}$ candidates have been produced.

\mypar{SIMIT-ICL.}
The resulting self-generated examples precede the original test query using the prompt template below. 

\begin{prompt}[In-Context Learning Prompt Format]
Here are \{K\} examples to help you answer the question:\\\\
Example 1: <synthetic image 1>\\
Question: \{synthetic question 1\}\\
Answer: \{intended answer 1\}\\
$\lbrack$... Examples 2 to K ...$\rbrack$\\\\
Now, answer this question: <test image> \{test prompt\}
\end{prompt}


\mypar{SIMIT-FT (Single-Query).}
For each test query, we train a LoRA adapter on its self-generated examples, then answer the original test query using the standard zero-shot prompt, and reset the adapter. 
All examples for a query are packed into one batch and optimized with AdamW at a constant learning rate. We tune the rank $r\in\{2,4,\ldots,64\}$, scale $\alpha\in\{4,8,\ldots,128\}$ independently of $r$, learning rate in $[3\times10^{-5},10^{-3}]$ on a log-uniform scale, and number of steps in $[2,40]$.

\mypar{SIMIT-FT (Test-Set).}
We pool self-generated examples across all test queries in a test set and fine-tune one LoRA adapter using BAGEL's official training code. We only finetune the textual backbone: LoRA is applied without dropout to the attention projections ($q,k,v,o$) and MLP projections (gate, up, down) of the language model's understanding expert. The vision encoder, connector, and image-generation parts remain frozen.

Training uses AdamW with $(\beta_1,\beta_2)=(0.9,0.95)$ and no weight decay, a cosine learning-rate schedule with $10\%$ warmup and a final rate of $10^{-7}$, gradient clipping at $1.0$. 
Across benchmarks, $r\in\{8,64\}$, $\alpha=2r$, and the learning rate ranges from $7\times10^{-5}$ to $2\times10^{-4}$. These are tuned hyperparameters on validation sets. We evaluate an exponential moving average of the weights with decay $0.95$. For a retained dataset of size $N$, the number of steps is $\mathrm{clip}(\lfloor N/2\rfloor,40,150)$, limiting repeated training on small, heavily filtered datasets. Training fits on a single H100 using fully sharded data parallelism (FSDP) and CPU offloading.

\mypar{SIMIT-FT$\to$ICL.}
We replace the base model in SIMIT-ICL with the SIMIT-FT (Test-Set) checkpoint, using the same demonstrations without additional tuning.



\mypar{Efficient Implementation.}
Synthesis uses the M$^\ast$ inference engine~\citep{jha2026mstar}, which supports composite multimodal models through paged attention, continuous batching, and CUDA-graph execution for both BAGEL's understanding and generation paths.
Since a single request decodes at
only $\sim$62 tokens/s and leaves the GPU largely idle, a bounded worker pool
processes multiple test queries concurrently so the server can batch their requests,
yielding a $1.76\times$ end-to-end speedup on one GPU, with similar gains for pool
sizes from 6 to 16. Packing LoRA finetuning batches to the engine's token-budget
ceiling, with proportionally fewer steps, gives a further $1.46\times$. 
However, the largest savings
are algorithmic: breadth-first slot filling cuts model calls per accepted example by
$1.9$--$3.5\times$, and ABA synthesizes on average only $23\%$ (SIMIT-ICL) to $41\%$
(SIMIT-FT) of $K_{\max}$ candidates per query.


\mypar{Efficient Hyperparameter Tuning. } ABA determines how many examples to retain for each query, while DF selects examples within a useful confidence range. To tune these rules without repeating synthesis, we generate and cache $K_{\max}$ candidates per query, their confidence scores, and the zero-shot confidence $p_0$. Each trial selects examples from the cache and reruns only adaptation and evaluation. For a fixed ordered candidate pool, this reproduces the selection rule using the cached confidences. The corresponding adaptive procedure stops once the allocated number of examples has been retained or the candidate cap is reached. Caching is used for tuning, and reported runtimes account for the query-specific adaptive budgets.

The search covers $\varepsilon\in[0,0.5]$, $A_0\in[0.05,0.6]$, $A_1\in[A_0,0.9]$, $B\in[0,0.5]$, $K_{\max}\in\{4,6\}$, $T_{\text{low}}\in[0,0.5]$, and $T_{\text{high}}\in[T_{\text{low}},1]$. We use Bayesian optimization with Optuna's tree-structured Parzen estimator, running an independent study for each benchmark and adaptation mode. The
objective is relative improvement over the zero-shot benchmark score on the validation split.


\subsection{TTS Baselines}

All baselines described here use BAGEL-7B with the same benchmark prompts, answer extraction, and metrics as SIMIT. Test-time scaling (TTS) methods process queries independently. 


\mypar{Thinking.} We simply enable BAGEL's thinking mode, which generates a reasoning trace before the final answer, with budgets of $1,000$ and $10,000$ tokens (\texttt{max\_think\_token\_n}).

\mypar{Self-Consistency.} 
We generate $K\in\{8,16,32\}$ responses per query and return the most frequent answer after text normalization. The first response is decoded greedily, the remaining $K-1$ are sampled at temperature $0.7$. Ties are resolved in favor of the earliest response. On open-ended tasks such as captioning, where responses are often all distinct, this rule returns the greedy answer. Including a greedy response stabilizes selection at small $K$, when ties are common. Since the greedy answer is the model's best estimate, we should use that when the candidate responses are all distinct.

\mypar{Self-Refine.} Following \citet{madaan2023selfrefine}, BAGEL answers the query, critiques its answer against the image, and produces a revised answer conditioned on the feedback. All iterations share an expanding context, with no early stopping. We evaluate $1$, $3$, and $5$ iterations. The refinement prompt preserves the original answer's style, length, and format to support benchmarks with constrained response formats.

\begin{prompt}[Self-Refine Feedback Prompt]
Review the answer you just gave above. Check it carefully against the image: 
is it factually correct, precise, and does it fully address the question? 
Point out any specific errors, missing details, or misreadings of the image. 
Give only feedback, not a revised answer.
\end{prompt}

\begin{prompt}[Self-Refine Prompt]
Now, taking your feedback above into account, give your final answer to the 
original question again, fixing any issues the feedback identified (or 
keeping the answer unchanged if the feedback found none). 
Match the exact style, length, and format of your original answer above -- 
reply with only the answer itself, no explanations or extra commentary.
\end{prompt}

\subsection{LFRL Baselines}

All LFRL methods train one LoRA adapter per benchmark on its 500 unlabeled test queries, allowing cross-query transfer.
We reproduce each method's original algorithm as faithfully as our problem setting
allows and state every deviation explicitly. We do not, however, adopt the
hyperparameters reported in the original papers: they were selected for
different models and benchmarks and transfer poorly to other models and benchmarks. 
For satisfactory performance, we therefore tune each baseline ourselves, under the same hyperparameter search budget as
SIMIT (Tab.~\ref{tab:lfrl_hparams}), so that every method is compared in a
configuration that performs well in our setting. 
For LFRL, the search includes rollout temperature, number of epochs, learning rate, and LoRA rank (Tab.~\ref{tab:lfrl_hparams}).

All LFRL baselines train LoRA adapters with dropout $0.05$, a cosine learning-rate schedule, $3\%$ warmup, gradient clipping at $1.0$, and completions capped at 128 tokens. Responses and supervision signals are recomputed from the current policy each epoch. For the GRPO baselines, each rollout group receives one on-policy update, so the importance ratio is one at the update and PPO-style clipping is inactive. Following TTRL, these updates use no KL penalty. High-resolution inputs are downscaled to a maximum side length of 512 pixels when needed to fit memory. Following TTRV's evaluation protocol, we decode the final TTRL, EMPO, and TTRV checkpoints with majority voting over 32 responses (maj@32). SRLM uses greedy decoding.

Existing self-improvement methods need hyperparameter calibration to perform well, and published settings transfer poorly across models and benchmarks. We therefore tune every method on the same 50 labeled validation examples per benchmark, with the same search budget. 
This calibration is a one-time cost, after which self-improvement on the target distribution is fully automatic. For instance, it needs only 50 labels per benchmark, compared with the 2,000 used by the real-data references in Sec.~\ref{sec:comparison_real_synthetic}. These labels only score candidate configurations and are never used as demonstrations, training targets, or verification signals. SIMIT is less sensitive to this tuning. Disabling ABA and DF removes all seven synthesis hyperparameters, and SIMIT still achieves +4.35\%, above every tuned LFRL baseline (Tab.~\ref{tab:ablation_components}). Fully hyperparameter-free self-improvement remains an open problem.

\begin{table}[h]
\centering
\small
\caption{Hyperparameter search spaces of the LFRL baselines. Defaults are in \textbf{bold}.}
\label{tab:lfrl_hparams}
\begin{tabular}{ll}
\toprule
\textbf{Hyperparameter} & \textbf{Values} \\
\midrule
Rollouts per query $K$ & 32 (TTRL, EMPO, TTRV); 4 (SRLM) \\
Rollout temperature & $\{0.6, \mathbf{1.0}, 1.3, 1.5\}$ \\
Epochs & $\{3, 4, 5, 6, \mathbf{10}\}$ \\
Learning rate & $\{5\times10^{-6}, \mathbf{1\times10^{-5}}, 2\times10^{-5}\}$ \\
LoRA rank $r$ ($\alpha = 2r$) & $\{\mathbf{16}, 32, 64\}$ \\
EMPO clustering threshold (word Jaccard) & $\{0.4, \mathbf{0.6}, 0.8\}$ \\
TTRV entropy weight $\alpha_H$ & $\{0, 0.25, 0.5, \mathbf{0.75}, 1.5\}$ \\
SRLM DPO temperature $\beta$ & $0.1$ \\
\bottomrule
\end{tabular}
\end{table}

\mypar{TTRL.} It uses the majority-voted answer as a pseudo-label. Matching responses receive reward one and all others receive zero. GRPO then updates the policy using group-normalized advantages. 


\mypar{EMPO.} It groups candidate responses by semantic equivalence and uses GRPO to increase semantic concentration. They provide two different variants in their paper. The first variant uses an external model. The second variant uses regular-expression
matching to check the semantic equivalence of two responses.
In our preliminary experiments, we tried both variants.
For the first variant, we used BAGEL itself to check the semantic equivalences using EMPO's verifier prompt, 
since we cannot use an external model due to our self-contained problem formulation. 
For the second variant, we tried naive matching and more sophisticated similarity-based character or word-level matching strategies.  
Clustering normalized responses using word-set Jaccard similarity outperforms others. 
Thus, we report the evaluation results for the best EMPO variant with lexical similarity based soft matching.

\mypar{TTRV.} It uses a frequency reward equal to the empirical probability of each response's answer class under exact matching. It also includes the negative entropy of the empirical answer distribution, weighted by $\alpha_H$; we use $0.75$ as the default.


\mypar{SRLM.} 
BAGEL samples $K=4$ responses per query at temperature $1.0$ and scores each using the prompt below. The highest- and lowest-scoring responses form the chosen and rejected pair; ties and unparseable judgments are discarded. Training uses DPO with $\beta=0.1$ and the frozen base model as reference. SRLM's runtime (Tab.~\ref{tab:main_comparison}) primarily reflects its sequential self-judging calls despite using eightfold fewer responses than the consensus-based methods. 

\begin{prompt}[SRLM Self-Judge Prompt]
Review the user's question and the corresponding response using the additive 5-point scoring system described below. Points are accumulated based on the satisfaction of each criterion:
\\~\\
- Add 1 point if the response is relevant and provides some information related to the user's inquiry, even if it is incomplete or contains some irrelevant content.\\
- Add another point if the response addresses a substantial portion of the user's question, but does not completely resolve the query or provide a direct answer.\\
- Award a third point if the response answers the basic elements of the user's question in a useful way, regardless of whether it seems to have been written by an AI Assistant or if it has elements typically found in blogs or search results.\\
- Grant a fourth point if the response is clearly written from an AI Assistant's perspective, addressing the user's question directly and comprehensively, and is well-organized and helpful, even if there is slight room for improvement in clarity, conciseness or focus.\\
- Bestow a fifth point for a response that is impeccably tailored to the user's question by an AI Assistant, without extraneous information, reflecting expert knowledge, and demonstrating a high-quality, engaging, and insightful answer.
\\~\\
User: \{instruction\}
\\~\\
<response>\{response\}</response>
\\~\\
After examining the user's instruction and the response:
\\~\\
- Briefly justify your total score, up to 100 words.\\
- Conclude with the score using the format: "Score: <total points>"
\\~\\
Remember to assess from the AI Assistant perspective, utilizing web search knowledge as necessary. To evaluate the response in alignment with this additive scoring model, we'll systematically attribute points based on the outlined criteria.
\end{prompt}



\subsection{Experimental Details}
\label{sec:appendix_prompts_diagram_categories}

\mypar{Skill Library Examples.} Fig.~\ref{fig:generation_prompts} shows the textual descriptions that we use to generate the visual examples in Fig.~\ref{fig:diagram_categories}.

\begin{figure}[h]
    \centering

    \noindent
    \promptcell{0.101\linewidth}{5.1cm}{Molecule}{
        Draw the molecular structure of caffeine.
    }\hfill
    \promptcell{0.121\linewidth}{5.1cm}{Venn}{
        Draw a Venn diagram of `Tools' (Hammer, Screwdriver) and `Kitchen Items' (Spoon, Knife), with Knife in the overlap since it can be both a tool and a kitchen item in this puzzle.
    }\hfill
    \promptcell{0.141\linewidth}{5.1cm}{Graph}{
        Draw a directed graph of a small social network's `follows' relationships: Alice follows Bob, Bob follows Carol, Carol follows Alice, Alice follows Dave, Dave follows Carol, and Dave follows Bob.
    }\hfill
    \promptcell{0.136\linewidth}{5.1cm}{SVG}{
        Draw a sun icon with a central yellow circle and 8 yellow rays radiating outward at equal angles. Use line elements for each ray: top, upper-right, right, lower-right, bottom, lower-left, left, upper-left.
    }\hfill
    \promptcell{0.096\linewidth}{5.1cm}{Natural}{
        A glazed donut and a chocolate candy bar placed on a white wooden table, photographed from above.
    }\hfill
    \promptcell{0.111\linewidth}{5.1cm}{Geometry}{
        Draw two circles: one centered at (0,0) with radius 3, another centered at (5,0) with radius 2.
    }\hfill
    \promptcell{0.101\linewidth}{5.1cm}{Vector}{
        Draw three vectors from the origin representing forces: F1 = (4, 0), F2 = (0, 3), F3 = (-2, -1), each labeled with its name.
    }\hfill
    \promptcell{0.161\linewidth}{5.1cm}{Diagram}{
        Draw a flowchart for an online order system: Customer Places Order, then Check Inventory. If In Stock, go to Process Payment, then Ship Order, then Delivered. If Out of Stock, go to Notify Customer, then Cancel Order.
    }

    \vspace{-5.5mm}
    
    \noindent
    \promptcell{0.244\linewidth}{4.7cm}{Vega-Lite}{
        Draw a bubble chart comparing 8 countries by GDP per capita (x-axis, in USD), life expectancy (y-axis, in years), and population (bubble size). Include: USA (63k, 79y, 330M), Germany (46k, 81y, 83M), Japan (40k, 84y, 125M), China (12k, 77y, 1400M), Brazil (8.8k, 75y, 213M), India (2.1k, 70y, 1380M), Nigeria (2k, 54y, 210M), Norway (80k, 83y, 5M). Color by country name.
    }\hfill
    \promptcell{0.124\linewidth}{4.7cm}{Scene}{
        Draw a scene scattered randomly with 4 overlapping blue circles, 3 red squares that do not overlap each other or anything else, and 2 green triangles that overlap each other.
    }\hfill
    \promptcell{0.214\linewidth}{4.7cm}{HTML}{
        Create an HTML nutrition facts label similar to a US food product label. Product: `Whole Grain Oatmeal'. Serving size 40g, Calories 150, Total Fat 3g (4\% DV), Saturated Fat 0.5g, Sodium 0mg, Total Carb 27g (10\% DV), Fiber 4g (14\% DV), Sugars 1g, Protein 5g, Vitamin D 0\%, Calcium 2\%, Iron 10\%. Black border, thick dividers.
    }\hfill
    \promptcell{0.164\linewidth}{4.7cm}{Figure}{
        Plot y = sin(x) and y = cos(x) on the same axes for x ranging from -2*pi to 2*pi, using two different colored lines with a legend distinguishing them, and mark every point in that range where the two curves intersect with a visible dot.
    }\hfill
    \promptcell{0.224\linewidth}{4.7cm}{Mermaid}{
        Draw a UML class diagram for a banking system: Account abstract class with accountNumber and balance fields, and deposit() and withdraw() methods; SavingsAccount subclass with interestRate field; CheckingAccount subclass with overdraftLimit field; Customer class with name and email, associated with Account.
    }

    \vspace{-5.0mm}

    \noindent
    \promptcell{0.114\linewidth}{2.5cm}{Circuit}{
        Draw a circuit with a 9V battery, a 220 ohm resistor, and an LED in series.
    }\hfill
    \promptcell{0.334\linewidth}{2.5cm}{Table}{
        Make a table of the first five elements of the periodic table with columns Name, Symbol, Atomic Number, and Atomic Mass: Hydrogen (H, 1, 1.008), Helium (He, 2, 4.0026), Lithium (Li, 3, 6.94), Beryllium (Be, 4, 9.0122), Boron (B, 5, 10.81).
    }\hfill
    \promptcell{0.244\linewidth}{2.5cm}{Puzzle}{
        Draw an analogy puzzle: a small circle next to a large circle, then a small square next to a large square, then a small triangle next to a question mark representing what comes next.
    }\hfill
    \promptcell{0.284\linewidth}{2.5cm}{Composite}{
        Four panels showing animals labeled 1, 2, 3, 4: panel 1 shows a golden retriever dog, panel 2 shows a tabby cat, panel 3 shows a green parrot perched on a branch, panel 4 shows a colorful tropical fish swimming.
    }
    
    \vspace{-5.0mm}

    \caption{Text prompts used to generate the visual examples in Fig.~\ref{fig:diagram_categories}.}
    \label{fig:generation_prompts}
\end{figure}

\mypar{Ablation Studies.}
Tabs.~\ref{tab:ablation_components} and~\ref{tab:ablation_fix_qi} use SIMIT-FT (Test-Set). To isolate synthesis components without repeating the per-benchmark ABA and DF searches, we first remove ABA and then DF. We then ablate DP, SL, and V individually from this fixed-budget reference. 
They use the same per-benchmark LoRA settings and 150 fine-tuning steps, so differences arise from data construction and selection.

The w/o ABA variant applies the tuned DF confidence band to all $K$ synthetic candidates per query. The w/o DP variant removes the diversity instructions from the synthesis prompt. The w/o SL variant bypasses routing and realizes every description through BAGEL's native image generation. The w/o V variant retains images regardless of the critic's verdict while preserving parser and renderer checks.

Tab.~\ref{tab:ablation_fix_qi} disables ABA and DF to compare synthesis strategies at a fixed requested sample count. The fixed-question variant supplies the test question and generates $K$ answer--description pairs for it. The fixed-image variant generates $K$ question--answer pairs about the test image and skips image realization; its critic checks whether that image supports each pair and discards pairs that fail. Fixed-budget variants share the requested sample count, although the number retained can differ after validation. Runtimes are reported separately. The full adaptive pipeline achieves the highest mean gain and a lower runtime than every ablated variant except w/o V.


\mypar{Data Quality.}
We audit synthetic data at scale using an agentic evaluation protocol with Claude Sonnet 5. This external evaluator receives anonymized (image, question, answer) triplets individually, without the producing variant's identity, and provides no feedback to synthesis or adaptation.  The Data Quality column in Tab.~\ref{tab:ablation_components} reports the percentage of triplets judged correct, excluding those flagged as ambiguous. This audit therefore assesses quality on queries for which the full pipeline supplies supervision. Sec.~\ref{sec:label_noise} also uses the same protocol to measure supervision correctness.

\begin{prompt}[Agentic Evaluation Protocol]
\textbf{Task}

You will judge a list of (image, question, answer) triplets, one at a time.
For each, decide: \textbf{is this triplet correct?} --- does the image actually
support answering the question, and does the given answer correctly answer
that question about that image? You are NOT told which method or system
produced any answer, and you should not try to guess --- judge every triplet
on its own merits, using the same standard for all of them.
\\~\\
\textbf{How to judge correctness (apply per question type)}

- Multiple choice (question lists lettered options): the answer is
correct only if it selects the option that is actually true of the image.

- Short-answer / single-word-or-phrase VQA: the answer is correct if it
is a reasonable, factually accurate response to the question given the
image. Minor differences in phrasing, capitalization, singular/plural, or
a synonym that means the same thing do NOT make an answer incorrect.

- Captioning (``provide a caption'', ``describe the image''): correct if the
caption accurately describes salient, verifiable content of the image. It
does not need to match any specific reference wording, and does not need
to describe every detail --- it needs to not be wrong or fabricated.

- Document / chart / infographic QA: the answer must match what is
actually shown in or computable from the image (a number, label, or
short text span). Treat this like short-answer VQA otherwise.

- ``Unanswerable'' / ``cannot be determined'' as the given answer: correct
ONLY if the image genuinely does not contain enough information to answer
the question. If the image clearly does support an answer, ``Unanswerable''
is INCORRECT.
\\~\\
\textbf{What to ignore}

Cosmetic image imperfections (slightly off rendering, minor visual
artifacts, mild blur/upscaling, imperfect but legible text) are irrelevant
UNLESS they actually obscure or change the information needed to answer the
question. Judge the content, not the production quality.
\\~\\
\textbf{When to flag ``ambiguous'' instead of deciding}

Use flagged\_ambiguous: true sparingly --- only when a careful, reasonable
person genuinely could not decide correct-vs-incorrect (e.g., the image
failed to load / is corrupted, the question itself is nonsensical or
untranslatable, or the answer is a genuine judgment call with real
ambiguity). Do NOT use it just because a question is hard, or because you
are not 100\% certain --- make your best judgment call in those cases instead
and explain why in the rationale. Ambiguous items are excluded from the
statistics and reported separately, so overusing this flag would throw away
real signal.
\\~\\
\textbf{Output format}

Exactly one JSON object per line (JSONL), one line per input triplet, in
any order, written to the exact output path you're given:

\{"audit\_id": "<copied from input>", "verdict": true, "flagged\_ambiguous": false, "rationale": "<one short sentence>"\}

verdict: true = correct triplet, false = incorrect triplet.
rationale: one short sentence explaining the call (or the ambiguity, if
flagged). Every input triplet must get exactly one output line.
\end{prompt}

\mypar{Stratified Gain Analysis.} We compare the best configuration of each method: SIMIT-FT$\to$ICL for SIMIT, and the best hyperparameter-tuned run per benchmark for each LFRL method (TTRL, EMPO, TTRV, SRLM). 
We define per-sample correctness from each benchmark's official metric. For normalized metrics, 
a sample
is correct if its score exceeds 0.5. For the five CIDEr-scored captioning/grounding benchmarks, CIDEr is corpus-relative and unbounded, so a single threshold has no natural interpretation. We instead apply a two-threshold rule to per-image CIDEr: scores $\geq 0.8$ are labeled correct, scores $< 0.2$ are labeled wrong (negligible n-gram overlap with any reference), and scores in between are ambiguous and excluded from our analysis. A transition counts only if a sample crosses fully from one confident bucket to the other. Landing in the ambiguous band does not count as a flip, which if anything biases reported gains conservatively downward.
Per-benchmark probabilities are 
averaged across all 17 
benchmarks to produce Fig. \ref{fig:stratified_gain}, giving each benchmark equal weight regardless of its wrong/correct split size.
  \section{Detailed Per-Benchmark Results}
\label{sec:appendix_detailed_results}

Tab.~\ref{tab:detailed_results} provides the per-benchmark breakdown of the SIMIT results summarized in Table~1, using the metrics described in Appendix~\ref{sec:appendix_dataset_stats}. These results complement the domain-level averages by showing how gains vary across individual benchmarks.

SIMIT-ICL improves on 16 of the 17 benchmarks and ties the baseline on one. Comparing SIMIT-ICL with test-set SIMIT-FT, the former leads on eight benchmarks and the latter on nine, highlighting their complementary strengths. Their combination achieves the highest mean relative improvement over the zero-shot baseline (+7.20\%), although it does not match the stronger individual variant on every benchmark.

\begin{table}[tbh]
  \centering
  \small
  \caption{Per-benchmark results for SIMIT variants. Small numbers next to each score show the relative percentage improvements over the BAGEL-7B baseline.}

  \label{tab:detailed_results}
  \setlength{\tabcolsep}{3pt}
  \begin{tabularx}{\linewidth}{
    @{}
    >{\raggedright\arraybackslash}p{0.16\linewidth}
    >{\raggedright\arraybackslash}p{0.155\linewidth}
    *{5}{>{\centering\arraybackslash}X}
    @{}
  }
        \toprule
        \textbf{Task Domain} & \textbf{Dataset} & \textbf{BAGEL-7B} & \textbf{SIMIT-ICL} & \textbf{SIMIT-FT (SQ)} & \textbf{SIMIT-FT (TS)} & \textbf{SIMIT-FT (TS) $\to$ ICL} \\
        \midrule
      
        \multirowcell{5}[0pt][l]{Image \&\\Captioning}
        & Flickr30k   & 0.877 & 0.886 \deltanum{+1.1} & 0.899 \deltanum{+2.5} & 0.896 \deltanum{+2.2} & 0.896 \deltanum{+2.2} \\
        & NoCaps      & 0.755 & 1.162 \deltanum{+53.9} & 0.762 \deltanum{+0.9} & 0.895 \deltanum{+18.5} & 0.953 \deltanum{+26.2} \\
        & TextCaps    & 1.586 & 1.586 \deltanum{+0.0} & 1.591 \deltanum{+0.3} & 1.591 \deltanum{+0.3} & 1.591 \deltanum{+0.3} \\
        & RefCOCO     & 0.779 & 0.803 \deltanum{+3.1} & 0.814 \deltanum{+4.4} & 0.901 \deltanum{+15.6} & 0.901 \deltanum{+15.6} \\
        & COCO        & 1.380 & 1.385 \deltanum{+0.4} & 1.384 \deltanum{+0.3} & 1.405 \deltanum{+1.8} & 1.406 \deltanum{+1.9} \\
        \midrule      

        \multirowcell{5}[0pt][l]{Visual\\Question\\Answering}
        & GQA         & 76.0 & 76.8 \deltanum{+1.1} & 76.4 \deltanum{+0.5} & 77.0 \deltanum{+1.3} & 77.0 \deltanum{+1.3} \\
        & OK-VQA       & 37.6 & 48.5 \deltanum{+28.9} & 41.6 \deltanum{+10.7} & 54.2 \deltanum{+44.0} & 54.2 \deltanum{+44.0} \\
        & VizWiz-VQA  & 47.0 & 54.6 \deltanum{+16.1} & 47.7 \deltanum{+1.4} & 53.0 \deltanum{+12.7} & 53.0 \deltanum{+12.7} \\
        & VQA-v2(-Small) & 78.1 & 78.6 \deltanum{+0.7} & 78.6 \deltanum{+0.7} & 78.5 \deltanum{+0.5} & 78.7 \deltanum{+0.7} \\
        & TextVQA     & 84.5 & 85.6 \deltanum{+1.3} & 84.9 \deltanum{+0.5} & 84.6 \deltanum{+0.1} & 85.1 \deltanum{+0.7} \\ 
        \midrule

        \multirow{1}{*}{Math \& Science}
        & AI2D        & 86.4 & 87.2 \deltanum{+0.9} & 86.6 \deltanum{+0.2} & 87.6 \deltanum{+1.4} & 87.8 \deltanum{+1.6} \\
        \midrule      

        \multirowcell{3}[0pt][l]{Doc. \&\\Infographic}
        & ChartQA     & 80.6 & 81.6 \deltanum{+1.2} & 81.2 \deltanum{+0.7} & 83.0 \deltanum{+3.0} & 83.0 \deltanum{+3.0} \\
        & DocVQA      & 93.2 & 93.6 \deltanum{+0.4} & 93.2 \deltanum{+0.0} & 93.2 \deltanum{+0.0} & 93.2 \deltanum{+0.0} \\
        & InfoVQA     & 61.5 & 62.8 \deltanum{+2.1} & 62.3 \deltanum{+1.4} & 62.0 \deltanum{+0.9} & 62.8 \deltanum{+2.1} \\
        \midrule

        \multirow{3}{*}{Multi-discipline}
        & MM-Bench (CN) & 84.8 & 87.2 \deltanum{+2.8} & 85.6 \deltanum{+0.9} & 86.4 \deltanum{+1.9} & 88.0 \deltanum{+3.8} \\
        & MM-Bench (EN) & 84.1 & 86.4 \deltanum{+2.7} & 85.6 \deltanum{+1.8} & 85.6 \deltanum{+1.8} & 85.6 \deltanum{+1.8} \\
        & SEED-Bench    & 77.6 & 79.0 \deltanum{+1.8} & 78.6 \deltanum{+1.3} & 79.8 \deltanum{+2.8} & 80.8 \deltanum{+4.1} \\

        \midrule
        \multicolumn{2}{@{}l}{\textbf{Mean Relative Improvement}} & \normaldeltanum{0.00\%} & \normaldeltanum{+6.98\%} & \normaldeltanum{+1.69\%} & \normaldeltanum{+6.39\%} & \textbf{\normaldeltanum{+7.20\%}} \\
        
        \bottomrule

  \end{tabularx}
\end{table}

  \section{Self-Generated Synthetic vs. Real Data: Additional Analyses}
\label{sec:appendix_scaling_real_synthetic}

How much can additional synthetic data improve performance, and when
does it match human-annotated data? We compare the two supervision
sources across three adaptation settings, then examine whether they
benefit the same benchmarks.

\mypar{Experimental Setup.}
We pair each synthetic-data method with its real-data counterpart:
SIMIT-ICL with retrieved in-context demonstrations (RICES),
SIMIT-FT (Single-Query) with finetuning on retrieved neighbors
(TTT-NN), and SIMIT-FT (Test-Set) with supervised finetuning on the
real-data pool. These comparisons use the same adaptation procedure
but different supervision sources. The real-data methods have access
to human annotations and, for retrieval, an external embedding model.
They therefore serve as privileged reference methods outside our
self-contained setting, rather than label-free baselines.

To compare the sources at identical sample counts, we disable dynamic $K$ mechanisms in our pipeline: adaptive
budget allocation (ABA) and difficulty filtering (DF). ABA varies the number
of generated candidates across queries, while DF changes how many are
retained according to their difficulty. Keeping either enabled would
mix the effect of adding samples with that of selecting or allocating
them. We instead use a fixed count for every query, while retaining
the other synthesis components, including verification. Throughout
this section, \emph{unfiltered} means that DF is disabled; it does not
mean that samples are unverified. 

Because this comparison disables two important components of our pipeline,
these results should not be interpreted as the performance of the full SIMIT pipeline.
With ABA and DF, SIMIT achieves approximately $6.4\%$ mean relative
gain over zero-shot, compared with $4.35\%$ when both are disabled
(Tab.~\ref{tab:ablation_components}). These components are disabled
here to fix sample counts, so the scaling curves do not represent
SIMIT with its full pipeline. Selection may be particularly important
for ICL: redundant, irrelevant, or incorrect demonstrations can
disrupt the context, whereas finetuning generally continues to
improve with additional samples. Thus, ICL is less tolerant
to imperfect supervision than finetuning, although the curves
do not isolate label noise from other sources of interference.


For in-context learning and single-query finetuning, we vary the
number of demonstrations or training examples per query over
$K\in\{0,\ldots,4\}$. Here, \emph{Single-Query} refers to adapting
independently for one target query, with model weights reset between
queries; each episode may use multiple training examples.
For test-set finetuning, the corresponding total training-set sizes
are $N_{\text{train}}\in\{0,500,1000,1500,2000\}$, with one shared model evaluated
across the target test set. These controlled curves characterize
fixed-count supervision, not SIMIT with its full adaptive pipeline.

\begin{figure}[t]
  \centering
  \includegraphics[width=\linewidth]{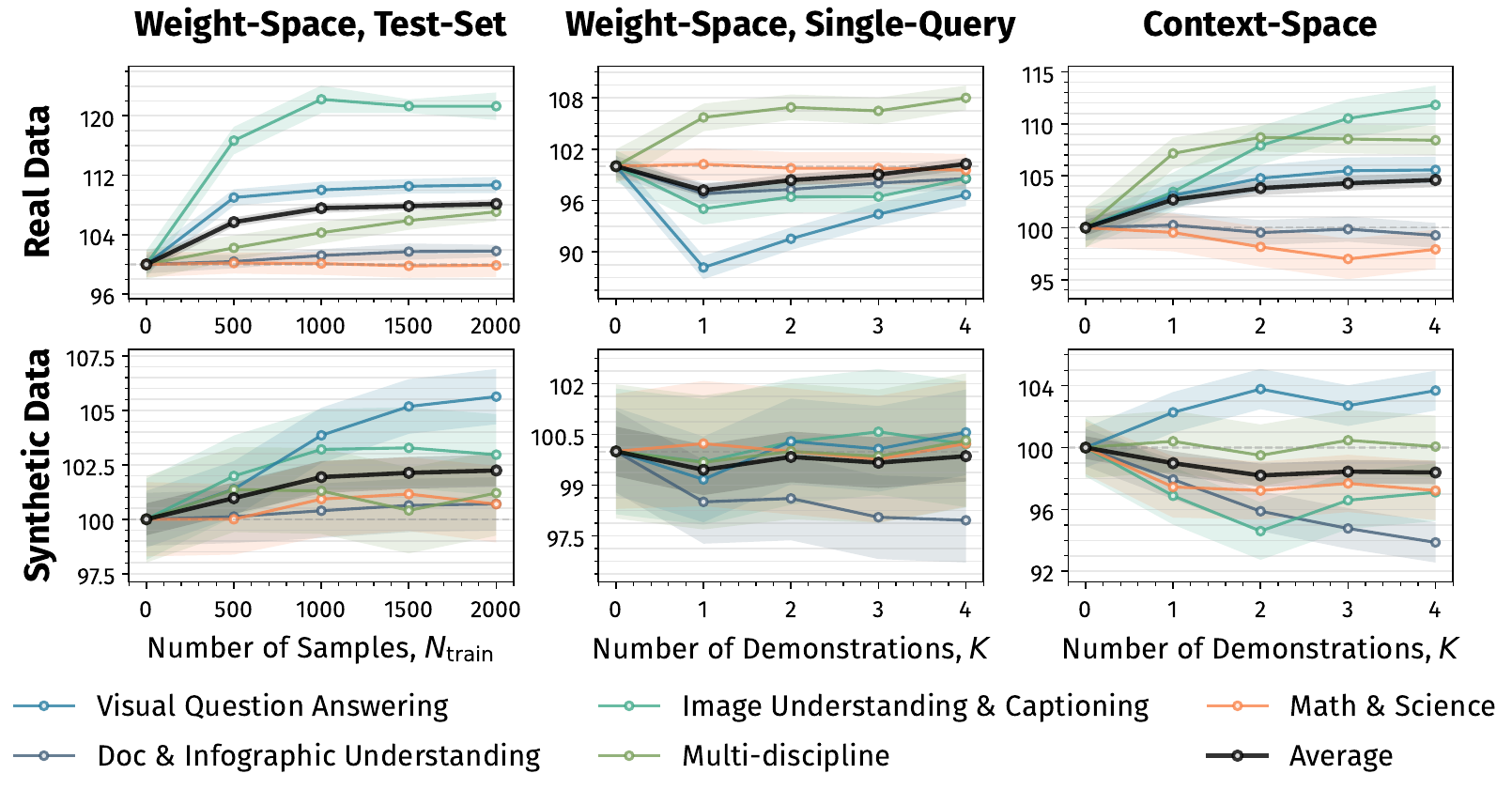}
  \caption{Relative performance as a percentage of the zero-shot baseline. 
  Context-Space and Weight-Space
  (Single-Query) use $K\in\{0,\ldots,4\}$ examples per target query;
  Weight-Space (Test-Set) uses $N_{\text{train}}\in\{0,500,1000,1500,2000\}$ training
  examples in total. ABA and DF are disabled to match sample counts. Thus,
  these plots do not reflect SIMIT's scaling behavior with all components.
  Additional real/synthetic samples do not benefit all adaptation settings equally.}
  \label{fig:ablation_kshots}
  \vspace{-1.8em}
\end{figure}

\mypar{Does Adding More Samples Consistently Help?}
Unfiltered synthetic ICL typically peaks at $K=2$-$3$, then
plateaus or declines, whereas finetuning on the same underlying
samples generally continues to improve
(Fig.~\ref{fig:ablation_kshots}). One possible explanation is that
additional visual demonstrations consume context and introduce
irrelevant or misleading information. Such demonstrations directly
condition the prediction, whereas finetuning combines their learning
signals through parameter updates. The curves do not isolate this
mechanism, but they suggest that selecting informative demonstrations
is particularly important for ICL.

\begin{wrapfigure}{r}{0.39\textwidth}
  \vspace{-1.33em}
  \centering
  \includegraphics[width=\linewidth]{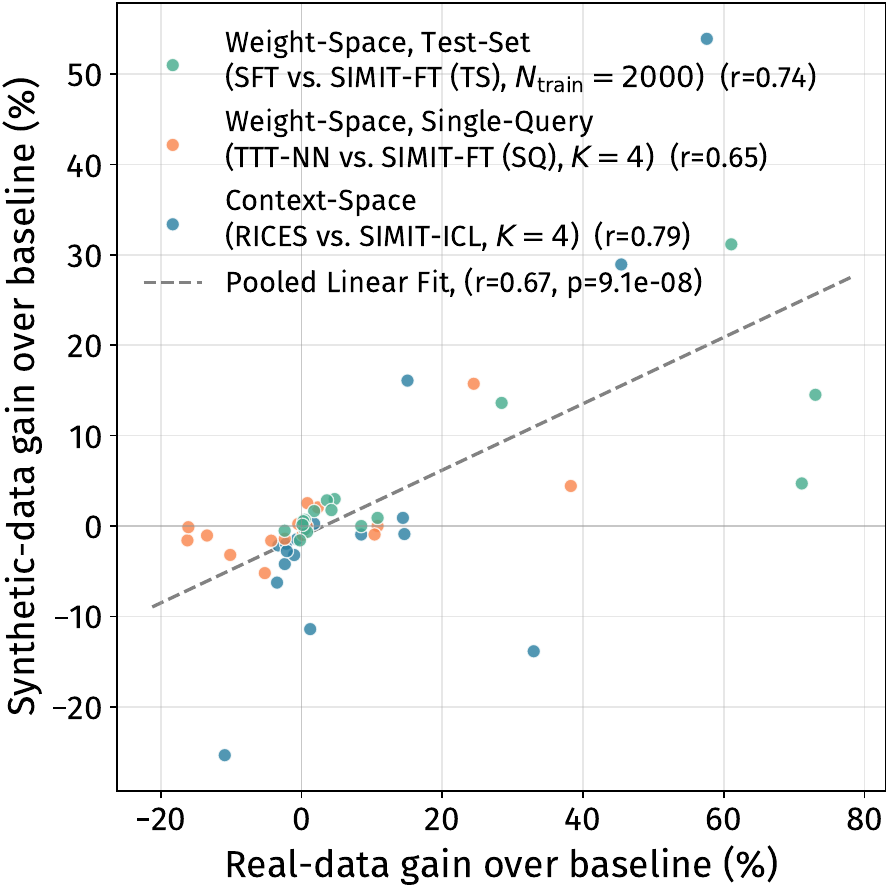}
  \caption{Per-benchmark relative gains from real and synthetic data
  at matched sample counts: $K=4$ for ICL and single-query
  finetuning, and $N_{\text{train}}=2000$ for test-set finetuning.
  Benchmarks that benefit more from real data also tend to benefit
  more from synthetic data.}
  \label{fig:real_vs_synthetic_scatter}
  \vspace{-2.4em}
\end{wrapfigure}

For test-set finetuning, increasing the synthetic-data budget does
not recover the largest real-data gains within the range studied.
Six benchmarks show an unfiltered synthetic gain above $2\%$:
OK-VQA, VizWiz, ChartQA, NoCaps, RefCOCO, and SeedBench. On each,
synthetic data falls short of the real-data result at $N_{\text{train}}=2000$,
recovering between $7\%$ (RefCOCO) and $79\%$ (SeedBench) of its gain.
On the other 11 benchmarks, gains are small, making a
synthetic-to-real sample ratio difficult to estimate reliably.
Thus, these experiments do not establish a stable exchange rate
between the two sources, nor do they determine whether larger
synthetic budgets would eventually close the gap.

\mypar{Do Two Sources Benefit the Same Benchmarks?}
Despite their performance differences, real- and synthetic-data gains
are positively correlated at matched sample counts
(Fig.~\ref{fig:real_vs_synthetic_scatter}).
The correlations are $r=0.79$ for ICL, $r=0.65$ for single-query
finetuning, and $r=0.74$ for test-set finetuning, with a pooled
$r=0.67$ across all $3\times17$ observations ($p<10^{-6}$).
Thus, synthetic supervision tends to help where real supervision
also helps.

Could the correlation arise simply because benchmarks with lower
zero-shot scores have more room to improve? We examine the 12
benchmarks with fixed score ceilings, excluding the five CIDEr-based
captioning benchmarks. For each supervision source, we fit a linear
model predicting the performance gain from the zero-shot score.
We then subtract the predicted gain from the observed gain and
correlate these residuals across the two sources. This tests whether
benchmarks that benefit more than expected from real data also
benefit more than expected from synthetic data.
The correlation remains strong even after this adjustment: pooled $r=0.83$,
compared with $r=0.90$ before adjustment. Within each setting, the
residual correlations are $r=0.79$ for ICL, $r=0.93$ for test-set
finetuning, and $r=0.77$ for single-query finetuning
(all $p<0.01$). Thus, a linear relationship with zero-shot
performance does not fully explain the association between
real- and synthetic-data gains.

\mypar{Where Do the Sources Differ?}
Real data generally provides larger gains, particularly on
Doc.\ \& Infographic tasks. However, synthetic data outperforms
real data on Math \& Science under both finetuning settings,
showing that query-conditioned generation can sometimes provide
more useful supervision than retrieved examples.

Synthetic data also offers more consistent benefits under
single-query adaptation. TTT-NN produces large gains on some
benchmarks but substantial regressions on others, consistent with
retrieved examples sometimes being poorly matched to the target
query. SIMIT-FT (Single-Query) can also regress, but its worst-case
decline is smaller, and it outperforms its real-data counterpart
on more benchmarks than in either of the other adaptation settings.
These results suggest that query-conditioned synthesis provides
more reliable supervision when adapting to an individual query
with only a few examples.


\mypar{Implications.}
These findings are consistent with SIMIT rehearsing and reorganizing
existing capabilities through targeted supervision, while remaining
limited by what the model can reliably generate.
Its advantage on
Math \& Science under finetuning, alongside real data's stronger gains
elsewhere, suggests complementary strengths: query-specific practice
from synthesis and broader visual and factual coverage from real data.


Real and synthetic data offer complementary benefits.
Real examples provide authentic content but depend on the coverage
of the available pool and the quality of retrieval. Synthetic
examples can target each query and vary in difficulty, but may
contain realization errors or redundant content. Diminishing returns
from unfiltered synthesis should consequently not be treated as an
intrinsic ceiling on synthetic supervision: sample count alone does
not capture how much valid, informative support the data provide.
The gains from ABA and DF reinforce this distinction.

  \section{Cross-Dataset Generalization}
\label{sec:appendix_cross_dataset}




\begin{figure}[b!]
\centering
\includegraphics[width=1.0\linewidth]{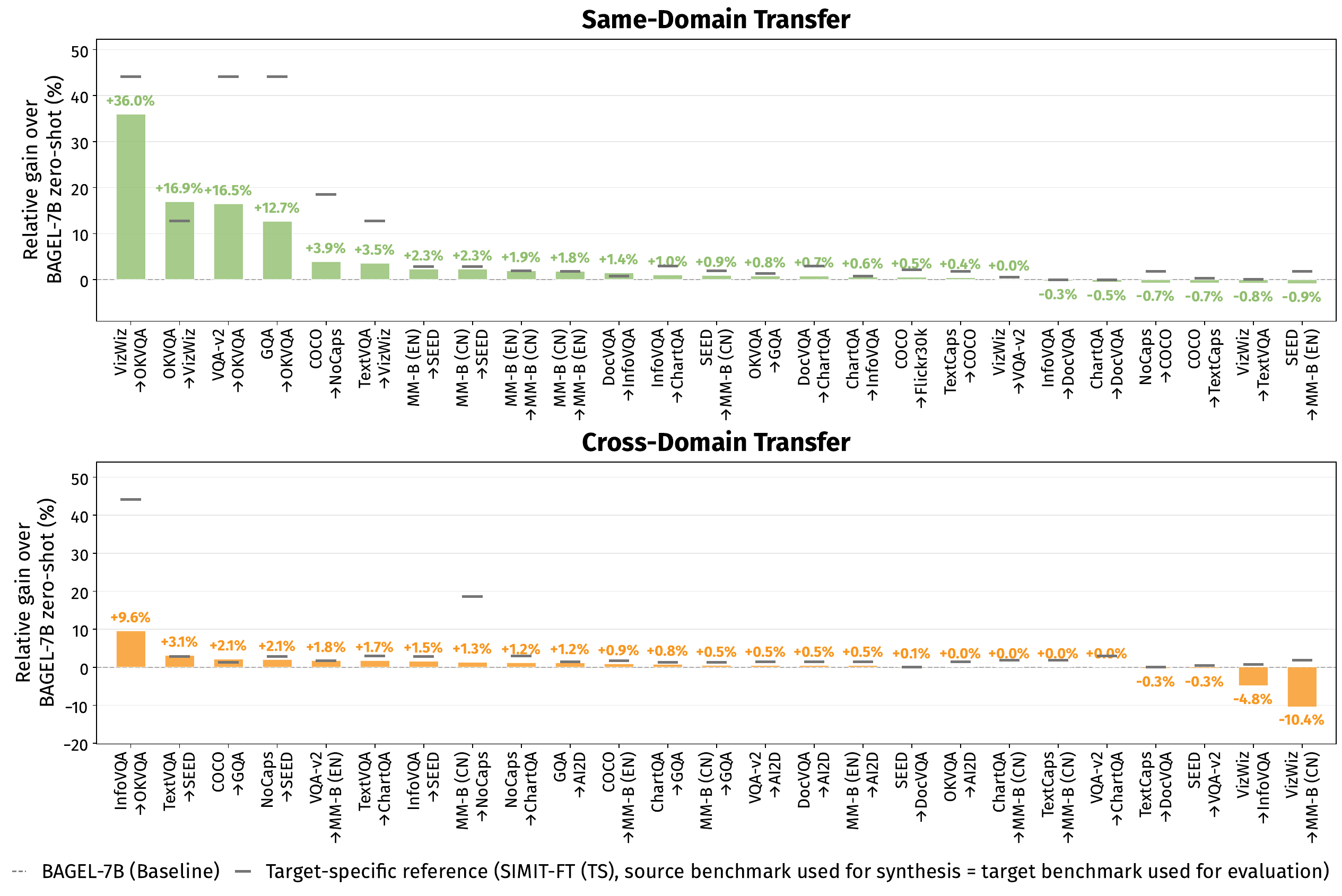}
\caption{Per-pair cross-dataset transfer.
    Each bar shows the relative gain over zero-shot BAGEL-7B
    after finetuning on synthetic data generated for source $X$
    and evaluating on target $Y$.
    Horizontal markers indicate target-specific SIMIT-FT
    (Test-Set) results. 
    }
\label{fig:cross_dataset_full}
\end{figure}

\begin{wrapfigure}{r}{0.5\textwidth}
\vspace{-0.1em}
\centering
\includegraphics[width=\linewidth]{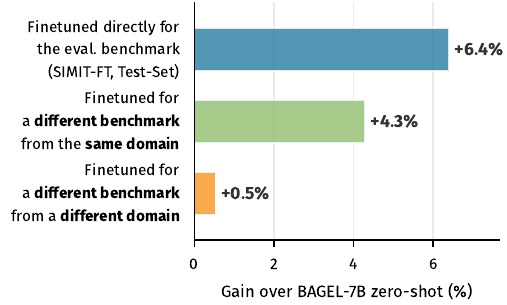}
\caption{Cross-dataset transfer, averaged over a matched set of source 
benchmarks: the same 14 benchmarks are each finetuned and evaluated on a
target from their own domain and on one from a different domain, isolating
domain relation from source data quality.
Synthetic data transfers well to related benchmarks, but
target-specific synthesis yields the largest average gain.}
\label{fig:cross_dataset}
\end{wrapfigure}

We evaluate whether SIMIT's self-generated training data remains useful
beyond the benchmark for which it was synthesized.
Fig.~\ref{fig:cross_dataset} shows substantial transfer to other
benchmarks within the same domain (+4.3\% on average), while
cross-domain transfer remains positive but much smaller (+0.5\%).
Thus, query-conditioned synthesis produces supervision with value
beyond its original target, particularly for related tasks.
Target-specific finetuning achieves a larger average gain (+6.4\%) and outperforms transferred
checkpoints on most evaluated pairs.
These findings support the value of target-local query-conditioned dataset:
synthetic data generated for other tasks can help, but directing
synthesis toward the evaluation queries is generally more effective.

\mypar{Experimental Design.} 
We evaluate 50 source-target pairs: 25 within the same domain and
25 across different domains. For each pair $X \to Y$, we finetune
BAGEL-7B on synthetic examples generated from the unlabeled queries
of source benchmark $X$, using SIMIT-FT (Test-Set), and evaluate
the resulting checkpoint on target benchmark $Y$ without further
adaptation. We compare against zero-shot BAGEL-7B and a
target-specific checkpoint trained on synthetic data generated
for $Y$. 
Synthetic data quality may vary across source benchmarks.
Comparing different source sets could therefore conflate transfer
with the model's ability to generate useful data for each source.
To control this factor, both transfer settings use the same
14 source benchmarks and reuse the same source-trained checkpoints.
We first average gains over the selected targets for each source,
then average equally across sources:
\begin{equation}
    \overline{g}_{r}
    =
    \frac{1}{|\mathcal{S}|}
    \sum_{X \in \mathcal{S}}
    \frac{1}{|\mathcal{T}_{r}(X)|}
    \sum_{Y \in \mathcal{T}_{r}(X)}
    g_{X \to Y},
    \qquad
    r \in \{\mathrm{same},\mathrm{cross}\},
\end{equation}
where $\mathcal{S}$ contains the 14 shared sources,
$\mathcal{T}_{r}(X)$ contains the selected targets in setting $r$,
and $g_{X \to Y}$ is the percentage improvement over the zero-shot
score on $Y$.
This marginalizes over the same empirical source distribution,
preventing sources with more evaluated targets from receiving
greater weight. It controls differences in source composition
and source-level data quality, although target difficulty and
source-target interactions can still affect the comparison.
The target-specific bar in Fig.~\ref{fig:cross_dataset} reports
the full-suite average separately.

\mypar{Transfer Patterns.}
Fig.~\ref{fig:cross_dataset_full} shows improvements on most
evaluated pairs, with particularly strong transfer to OK-VQA:
all tested sources improve its performance, yielding gains
between +9.6\% and +36.0\%.
Conversely, transfer to DocVQA is small or slightly negative,
consistent with the limited improvement observed from
target-specific finetuning on this benchmark.
Positive averages do not imply uniformly beneficial transfer:
VizWiz-VQA transfers negatively to InfoVQA (-4.8\%) and
MM-Bench (CN) (-10.4\%), showing the potential cost of
adapting to a mismatched 
distribution.

\mypar{Target Conditioning Matters.}
Target-specific synthesis outperforms transferred supervision
on most pairs, including many within the same domain.
The exceptions clarify the scope of this advantage.
Transfer between MM-Bench (EN) and MM-Bench (CN) matches
target-specific performance, consistent with their closely
related tasks. OK-VQA $\to$ VizWiz-VQA exceeds the target-specific
reference (+16.9\% versus approximately +13\%), showing that
a related source can occasionally provide more useful supervision.

Overall, these results support conditioning synthesis on the
target workload rather than relying on synthetic data generated
for other tasks. The comparison tests target-conditioned data
against data conditioned on a different benchmark; it therefore
provides evidence for target alignment, rather than a direct
evaluation of fully generic, unconditioned synthesis.

\FloatBarrier
\begin{takeaway}
Our synthetic datasets generalize broadly enough to be useful outside their intended benchmark, especially for related benchmarks.
But target-specific, query-conditioned synthesis is consistently the better choice. 
SIMIT’s gains arise not merely from demonstration format or additional visual context, but from constructing a target-local dataset. 
\end{takeaway}
  \section{Cross-Model Generalization}
\label{sec:appendix_cross_model}

\begin{table}[h]
  \centering
  \small
  \caption{Per-benchmark SIMIT-ICL results for BAGEL-7B and two additional unified multimodal models, SenseNova-U1-8B and Lance-3B. Small numbers next to each score show the relative percentage improvement over that model's own zero-shot baseline. SenseNova-U1-8B's zero-shot RefCOCO score is near-zero CIDEr\textsuperscript{\textdagger}, so no meaningful percentage gain can be reported for this cell. It is excluded from the mean relative improvement.}

  \label{tab:detailed_results_all_models}
  \setlength{\tabcolsep}{2pt}
  \begin{tabularx}{\linewidth}{
    @{}
    >{\raggedright\arraybackslash}p{0.16\linewidth}
    >{\raggedright\arraybackslash}p{0.155\linewidth}
    *{3}{>{\centering\arraybackslash}p{0.100\linewidth}>{\centering\arraybackslash}X}
    @{}
  }
        \toprule
        \multirowcell{2}[-2pt][l]{\textbf{Task Domain}} & \multirowcell{2}[-2pt][l]{\textbf{Dataset}} &
        \multicolumn{2}{c}{\textbf{BAGEL-7B}} &
        \multicolumn{2}{c}{\textbf{SenseNova-U1-8B}} &
        \multicolumn{2}{c}{\textbf{Lance-3B}} \\
        \cmidrule(lr){3-4}\cmidrule(lr){5-6}\cmidrule(lr){7-8}
         & & Zero-Shot & SIMIT-ICL & Zero-Shot & SIMIT-ICL & Zero-Shot & SIMIT-ICL \\
        \midrule
      
        \multirowcell{5}[0pt][l]{Image \&\\Captioning}
        & Flickr30k   & 0.877 & 0.886 \deltanum{+1.1}  & 0.768 & 0.770 \deltanum{+0.3}  & 0.343 & 0.343 \deltanum{+0.0}  \\
        & NoCaps      & 0.755 & 1.162 \deltanum{+53.9} & 0.524 & 0.574 \deltanum{+9.6}  & 0.422 & 0.426 \deltanum{+0.8}  \\
        & TextCaps    & 1.586 & 1.586 \deltanum{+0.0}  & 0.677 & 0.677 \deltanum{+0.0}  & 0.877 & 0.877 \deltanum{+0.0}  \\
        & RefCOCO     & 0.779 & 0.803 \deltanum{+3.1}  & 0.000 & 0.057 \textsuperscript{\textdagger}    & 0.157 & 0.173 \deltanum{+9.8}  \\
        & COCO        & 1.380 & 1.385 \deltanum{+0.4}  & 1.247 & 1.248 \deltanum{+0.1}  & 0.411 & 0.412 \deltanum{+0.3}  \\
        \midrule
        
        \multirowcell{5}[0pt][l]{Visual\\Question\\Answering}
        & GQA         & 76.0 & 76.8 \deltanum{+1.1}  & 69.4 & 73.0 \deltanum{+5.2}  & 67.8 & 68.2 \deltanum{+0.6}  \\
        & OK-VQA      & 37.6 & 48.5 \deltanum{+28.9} & 53.2 & 53.9 \deltanum{+1.4}  & 27.6 & 28.0 \deltanum{+1.7}  \\
        & VizWiz-VQA  & 47.0 & 54.6 \deltanum{+16.1} & 48.2 & 53.1 \deltanum{+10.1} & 13.0 & 14.3 \deltanum{+10.2} \\
        & VQA-v2      & 78.1 & 78.6 \deltanum{+0.7}  & 76.6 & 77.1 \deltanum{+0.7}  & 69.7 & 69.8 \deltanum{+0.2}  \\
        & TextVQA     & 84.5 & 85.6 \deltanum{+1.3}  & 77.5 & 78.9 \deltanum{+1.8}  & 58.1 & 59.5 \deltanum{+2.4}  \\
        \midrule
        
        \multirow{1}{*}{Math \& Science}
        & AI2D        & 86.4 & 87.2 \deltanum{+0.9}  & 87.8 & 88.4 \deltanum{+0.7}  & 40.2 & 41.6 \deltanum{+3.5}  \\
        \midrule
        
        \multirowcell{3}[0pt][l]{Doc. \&\\Infographic}
        & ChartQA     & 80.6 & 81.6 \deltanum{+1.2}  & 79.0 & 80.0 \deltanum{+1.3}  & 25.8 & 26.6 \deltanum{+3.1}  \\
        & DocVQA      & 93.2 & 93.6 \deltanum{+0.4}  & 93.2 & 94.0 \deltanum{+0.8}  & 35.7 & 37.5 \deltanum{+5.1}  \\
        & InfoVQA     & 61.5 & 62.8 \deltanum{+2.1}  & 70.9 & 72.0 \deltanum{+1.6}  & 17.3 & 18.0 \deltanum{+3.8}  \\
        \midrule
        
        \multirow{3}{*}{Multi-discipline}
        & MM-Bench (CN) & 84.8 & 87.2 \deltanum{+2.8} & 88.0 & 88.0 \deltanum{+0.0} & 46.4 & 52.8 \deltanum{+13.8} \\
        & MM-Bench (EN) & 84.1 & 86.4 \deltanum{+2.7} & 88.6 & 89.4 \deltanum{+0.8} & 58.3 & 65.2 \deltanum{+11.7} \\
        & SEED-Bench    & 77.6 & 79.0 \deltanum{+1.8} & 80.0 & 81.0 \deltanum{+1.3} & 69.6 & 70.0 \deltanum{+0.6} \\
        
        \midrule
        \multicolumn{2}{@{}l}{\textbf{Mean Relative Improvement}} & \normaldeltanum{0.00\%} & \normaldeltanum{+6.98\%} & \normaldeltanum{0.00\%} & \normaldeltanum{+2.21\%} & \normaldeltanum{0.00\%} & \normaldeltanum{+3.97\%} \\
        
        \bottomrule
  \end{tabularx}
\end{table}

To test whether SIMIT's gains are specific to BAGEL-7B, we apply the identical,
unmodified pipeline (prompts, skill library, verification, difficulty
filtering, and adaptive budget allocation) to SenseNova-U1-8B and Lance-3B, two
other unified multimodal models. Because neither model ships a public
fine-tuning interface, we evaluate SIMIT-ICL only: a purely in-context,
weight-free adaptation. This is exactly the setting where SIMIT's
model-agnostic design matters most, since it requires no training access and
could equally be deployed against a closed, API-only model.

Across the same
17-benchmark suite used for BAGEL-7B, SIMIT-ICL improves SenseNova-U1-8B 
on most benchmarks 
for a mean relative
gain of +2.21\%. RefCOCO is excluded from this average because SenseNova-U1's
zero-shot score there is degenerate, approximately 0 CIDEr. But SIMIT-ICL still
recovers from it, shown in Tab.~\ref{tab:detailed_results_all_models}. Gains are largest in Visual Question Answering (+3.4\%, driven by
VizWiz-VQA +10.1\% and GQA +5.2\%) and Image Understanding \& Captioning
(+3.4\%), and smallest in Math \& Science and Multi-discipline (+0.7\% each).
VQA and Image \& Captioning are BAGEL's two largest-gaining domains as well. This cross-model
agreement in which domains benefit suggests the pattern is driven by the
target data and task structure that SIMIT's synthetic datasets best support,
rather than by BAGEL-specific idiosyncrasies.

Lance-3B, the weakest zero-shot model of the three, shows a large overall
gain (+3.97\%), concentrated in Multi-discipline reasoning
(MM-Bench CN +13.8\%, MM-Bench EN +11.7\%), while captioning barely moves.
This suggests SIMIT-ICL's benefit scales with how much headroom a model's
zero-shot behavior leaves on a given task, rather than with model size or
architecture, and confirms that SIMIT's benefits generalize to other models beyond the one used to develop the method.

Note that while our evaluated models are all UMMs, actually image generation is not strictly
required. A standard understanding-only VLM can also act as the synthesizer by
either restricting synthesis to its own image-generation-free skill library
(structured visuals only; \citealp{yang2025scaling}) or by retaining the test image and generating only
new questions, as in our fixed-image ablation (Tab.~\ref{tab:ablation_fix_qi}).




\begin{takeaway}
SIMIT-ICL's benefits are not BAGEL-specific: the same pipeline improves SenseNova-U1 and Lance-3B, gaining most where the base model has the most zero-shot headroom.
\end{takeaway}
  \section{Label-Noise, Model-Noise Alignment: Additional Analyses}
\label{sec:appendix_label_model_noise_alignment}

\textbf{Experimental Protocol.} Sec.~\ref{sec:label_noise} reports $\delta$, not the raw corruption rate
$c$, because two supervision sources with identical $c$ can have different
training effects depending on whether their errors concentrate on the
solver's existing failures: label noise independent of the input leaves the
optimum unchanged for substantial noise rates \citep{natarajan2013learning},
while noise concentrated on the model's own errors reinforces them under
continued fitting. Equivalently, $\delta = \Cov(L,M)/\Var(M)$. 
Thus, in order to compare the supervision quality of different methods, for each method, we need to measure the expected label noise and its correlation with model noise.

\textbf{Evaluator.} Correctness of a label is judged by Claude Sonnet 5, a blind external evaluator 
that never
sees the producing method, its reward, or sibling candidates. It sees only one
(image, question, answer) triplet at a time. This blind external evaluator
is used exclusively for this audit: it supplies no signal
to SIMIT's own synthesis, verification, or adaptation pipeline, and to no
baseline's training. It is used only for evaluation purposes.
We use the same prompting with our data quality experiments (See ``Agentic Evaluation Protocol'' in Appendix~\ref{sec:appendix_implementation_details}).

\textbf{Overall Procedure.} For the $i^{\text{th}}$ test sample in the benchmarks, we first measure the associated model noise realization $M^{(i)}$, whether the base model answers it incorrectly. Then, we measure the label noise for different methods using their supervision signals: $L_{\mathrm{SIMIT}}^{(i)}$, $L_{\mathrm{TTRL}}^{(i)}$, $L_{\mathrm{TTRV}}^{(i)}$, $L_{\mathrm{EMPO}}^{(i)}$, and $L_{\mathrm{SRLM}}^{(i)}$. This gives us five different $(L^{(i)}, M^{(i)})$ pairs to compare the supervision quality of different methods. We use these recorded realizations $\{(L^{(i)}, M^{(i)})\}_{i=1}^N$ to get Monte Carlo estimates for the mean, variance, and covariance of the binary random variables $L$ and $M$.
We report four summary statistics: $c = P(L{=}1)$, $c_{\text{wrong}} = P(L{=}1\mid M{=}1)$, $c_{\text{correct}} = P(L{=}1\mid M{=}0)$ and $\delta = c_{\text{wrong}} - c_{\text{correct}}$. 

\begin{figure}[t!]
\centering
\includegraphics[width=1.0\linewidth]{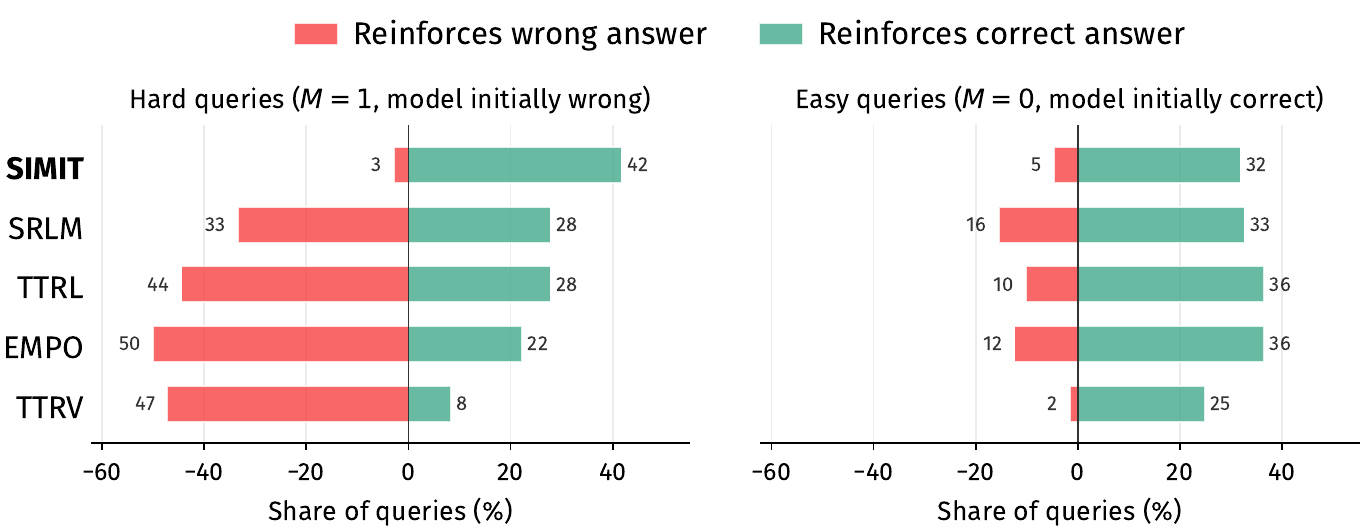}
\caption{Outcome decomposition (incorrect supervision / no positive
supervision / correct supervision) split by base-model correctness. \textbf{(Left)}
Hard, initially-wrong queries ($M{=}1$). \textbf{(Right)} Easy, initially-correct queries ($M{=}0$).}
\label{fig:label_noise_decomposition_full}
\vspace{-1.2em}
\end{figure}

\textbf{Definition of Label Noise.} For SIMIT, defining a ``pseudolabel'' for the given test query is trivial: use the intended answers in the self-generated synthetic samples.
Thus, the label noise definition is also trivial.
However, for the LFRL baselines, defining a ``pseudolabel'' is not trivial. They do not simply 
train on individual samples with associated labels. Therefore, RL training complicates the label noise
definition for LFRL methods.
We use each query's
actual rollout-level rewards and advantages,
apply that method's own GRPO normalization unmodified, and sample one
rollout with probability proportional to its positive advantage,
$P(j\mid i)=\max(A_{ij},0)/\sum_k\max(A_{ik},0)$. This experimental design avoids auditing the
majority-voted answer directly, which would make TTRL and TTRV
indistinguishable whenever they share rollouts despite their different
reward rules. Auditing every positive-advantage rollout and aggregating
judgments by $P(j\mid i)$ would give each query's exact expected
supervision-error rate, but at a multiple of the evaluator cost. Sampling
one rollout from the same distribution is an unbiased Monte Carlo estimator
of that exact quantity for every statistic reported below. $c$,
$c_{\text{wrong}}$, $c_{\text{correct}}$, and $\delta$ are all means or
covariances of the per-query rate, and independent, mean-zero measurement
error in an outcome does not bias a mean or covariance, only its variance.
And, we audit the \emph{positively encouraged} pseudo-label only, the response whose
probability a method's own update actually increases, not the full RL
update, since suppressed (negative-advantage) responses are a separate
question from what training teaches the model to say. 
For SIMIT, one
DF+ABA-retained synthetic example is sampled uniformly among those
surviving verification and filtering for that query.

Whenever every candidate has non-positive
advantage (or SIMIT retains nothing), we record
a \emph{``no positive supervision''} case rather than default to any answer. From such records, we report the ``Availability'' metric, the share of test queries with any positive supervision. 
All other reported statistics are estimated from test queries with a recorded available supervision.


\begin{wraptable}{r}{0.5\textwidth}
\vspace{-1.0em}
\setlength{\tabcolsep}{6pt}
\centering
\caption{Full label-noise, model-noise alignment audit, pooled across
queries. $c$: overall supervision-error rate. $c_{\text{wrong}}$/$c_{\text{correct}}$:
error rate conditioned on the base model being wrong/right on that query.
$\delta=c_{\text{wrong}}-c_{\text{correct}}$. Avail.: share of queries with
any positive supervision at all.}
\label{tab:noise_alignment_full}
\small
\begin{tabularx}{\linewidth}{
    l
    *{5}{>{\centering\arraybackslash}X}
}
\toprule
\textbf{Method} & $c$\textbf{ (\%)} & $c_{\text{wrong}}$\textbf{ (\%)} & $c_{\text{correct}}$\textbf{ (\%)} & $\delta$\textbf{ (pp)} & \textbf{Avail.\ (\%)} \\
\midrule
TTRL  & 33.0 & 61.5 & 21.7 & $+39.9$ & 52.1 \\
EMPO  & 38.5 & 69.2 & 25.4 & $+43.8$ & 53.9 \\
TTRV  & 35.2 & 85.0 &  5.9 & $+79.1$ & 32.7 \\
SRLM  & 38.4 & 54.5 & 32.3 & $+22.3$ & 50.9 \\
\textbf{SIMIT} & \textbf{12.1} & \textbf{6.3} & 12.8 & \textbf{$-6.5$} & 38.2 \\
\bottomrule
\end{tabularx}
\end{wraptable}

$\delta$ conditions on supervision being available at all, so it is silent
about abstention: a method that skips most hard queries and is wrong on the
rest is indistinguishable, by $\delta$ alone, from one that never skips and
is wrong less often overall. Tab.~\ref{tab:noise_alignment_full} and
Fig.~\ref{fig:label_noise_decomposition_full} instead assign every query
to exactly one of three outcomes: correct supervision, incorrect
supervision, or no positive supervision. They are also split by whether the base model
already solves it. On hard queries, SIMIT is the only method with a
positive net rate (correct minus incorrect supervision, $+38.9$pp); TTRV's
is the most negative ($-38.9$pp), and the other three LFRL baselines are
also negative but smaller in magnitude (TTRL $-16.7$pp, EMPO $-27.8$pp,
SRLM $-5.6$pp). On easy queries, every method nets positive, including all
four LFRL baselines. The methods disagree specifically about what
happens on the queries that most need fixing, not about queries the model
already solves.

  \todo{DON'T FORGET TO INCLUDE QUALITATIVE RESULTS AGAIN! I COMMENTED OUT BECAUSE IMAGES ARE TOO BIG, SLOW RENDERING}
  \clearpage
\section{Skill Library Ablation: Additional Samples}
\label{sec:appendix_skill_lib_ablation}

\begin{figure}[tbh]
    \centering
    \setlength{\tabcolsep}{4pt}
    \renewcommand{\arraystretch}{1.2}
    \begin{tabularx}{\textwidth}{
        @{}
        >{\raggedright\arraybackslash}m{0.3\textwidth}
        >{\centering\arraybackslash}m{0.33\textwidth}
        >{\centering\arraybackslash}m{0.33\textwidth}
        @{}
    }
        \toprule
        \textbf{Prompt}
        & \textbf{w/o Skill Library}
        & \textbf{w/ Skill Library}
        \\
        \midrule
        Draw a graph with two separate disconnected components: one with nodes A, B, C forming a triangle of edges, another with nodes X and Y connected by one edge.
        & \sampleimage{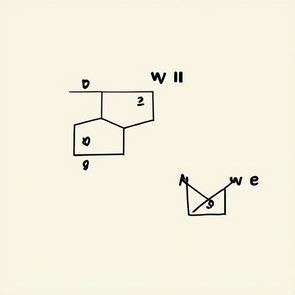}
        & \sampleimage{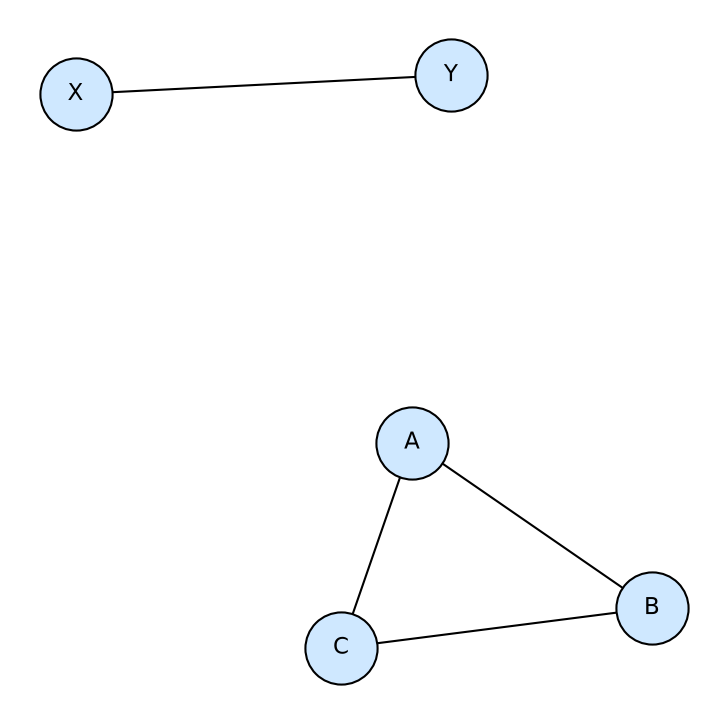}
        \\ \addlinespace[3pt]
        Draw a Venn diagram comparing `Citrus Fruits' (Lemon, Lime) and `Round Fruits' (Apple, Grapefruit), with Orange in the overlap of both sets.
        & \sampleimage{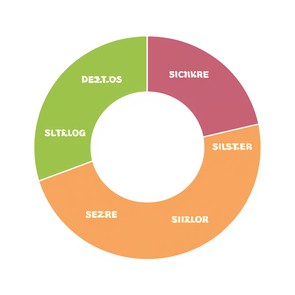}
        & \sampleimage{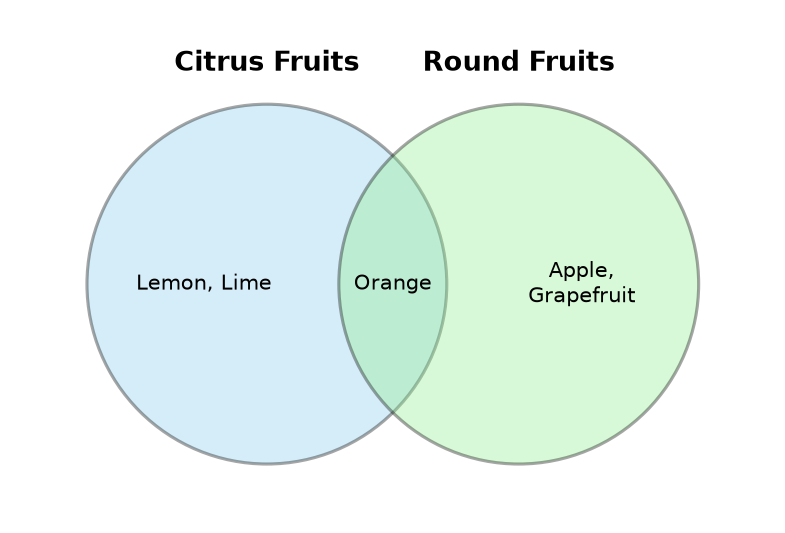}
        \\ \addlinespace[3pt]
        Create an invoice for a freelance web developer: company name `DevCraft Studio', client `TechCorp Inc', invoice date 2024-03-15, line items: Frontend Development 60h @ \$120/h = \$7,200, API Integration 20h @ \$100/h = \$2,000, Testing 10h @ \$80/h = \$800. Subtotal \$10,000, Tax 10\% = \$1,000, Total \$11,000. Professional dark-header styling.
        & \sampleimage{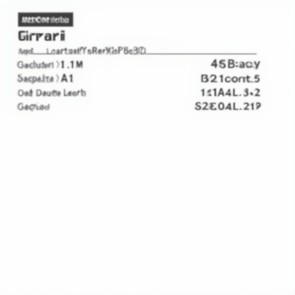}
        & \sampleimage{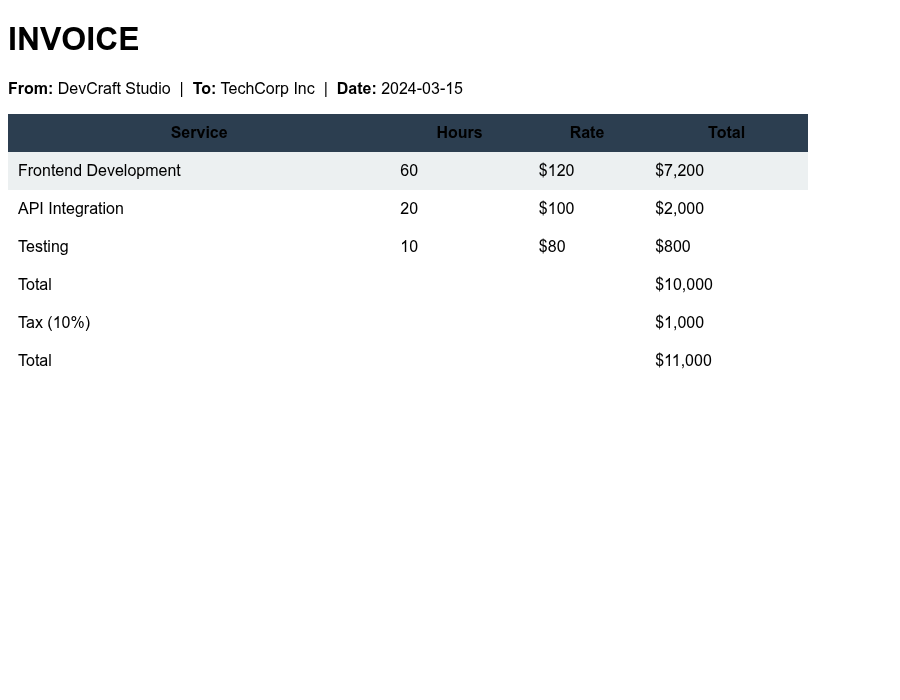}
        \\ \addlinespace[3pt]
        Draw a state machine: Idle goes to Running on `start', Running goes to Paused on `pause', Paused goes to Running on `resume', Running goes to Idle on `stop'.
        & \sampleimage{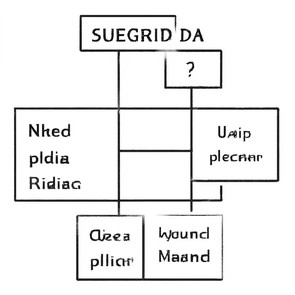}
        & \sampleimage{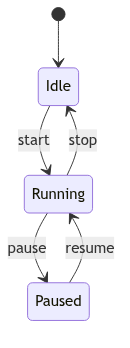}
        \\ \addlinespace[3pt]
        Draw a Gantt chart for a software development project over 8 weeks: Week 1 is Requirements gathering; Weeks 2-3 are System Design; Weeks 3-6 are Backend Development (starting mid-week 3); Weeks 4-6 are Frontend Development; Week 7 is Integration Testing; Week 8 is Deployment.
        & \sampleimage{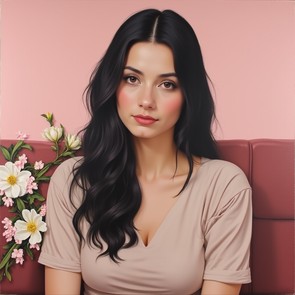}
        & \sampleimage{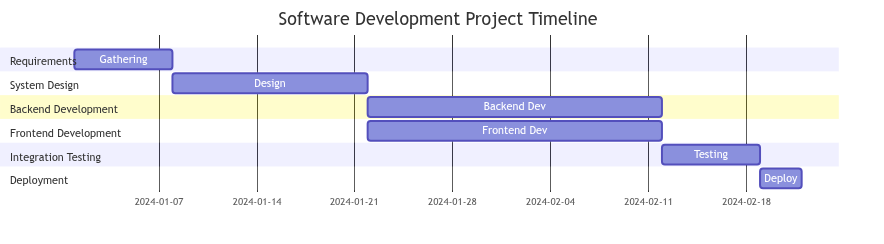}
        \\
        \bottomrule
    \end{tabularx}
    \caption{Skill library ablation. Native generation produces structural errors and, in the bottom example, entirely unrelated content; deterministic rendering improves fidelity to the prompts.}
    \label{fig:diagram_library_qualitative_appendix_extra}
\end{figure}

Fig.~\ref{fig:diagram_library_qualitative_appendix_extra} illustrates failures of native image generation on structured content: outputs may resemble the requested format but contain completely incorrect text or relations. In the bottom example, the model hallucinates a completely unrelated portrait instead of a Gantt chart. Our skill library improves fidelity by rendering the specified content and structure through deterministic backends.

  \section{Qualitative Results}
\label{sec:appendix_qualitative_results}

\begin{figure}[tbh]
    \centering
    \footnotesize
    \setlength{\tabcolsep}{4pt}
    \renewcommand{\arraystretch}{1.2}
    \renewcommand{\tabularxcolumn}[1]{m{#1}}
    \begin{tabularx}{\textwidth}{
        @{}
        >{\hsize=0.9375\hsize\linewidth=\hsize\raggedright\arraybackslash}X
        >{\hsize=1.03125\hsize\linewidth=\hsize\raggedright\arraybackslash}X
        >{\hsize=1.03125\hsize\linewidth=\hsize\raggedright\arraybackslash}X
        @{}
    }
    \textbf{Unlabeled test query} & \textbf{Retrieved real samples} & \textbf{Self-generated synthetic samples} \\
    \midrule
    \includegraphics[width=\linewidth,height=4.2cm,keepaspectratio]{figures/samples_images/nocaps_val_226_query}\par\vspace{4pt}
    \textbf{Prompt:} \textit{Provide a one-sentence caption for the provided image.}\par\vspace{4pt}
    \textbf{GT:} Two people sitting down beside their instruments and playing guitar , Two men playing the guitar with multiple music equipment on the ground., Two men play guitar surrounded by drums and speakers in front of some trailers., Two men sit, one on a chair and the other on the floor, while they each play a guitar, and they are surrounded buy drums and other band equipment partly loaded into trucks., two people sitting and playing instruments together by themselves, Two musicians sit outside their vehicle practicing on their guitars..\par\vspace{4pt}
    \textbf{Pretrained:} A man sits in front of an EVH amplifier.\ \textcolor{red!70!black}{\ding{55}} {\scriptsize(CIDEr=0.007)}
    &
    \includegraphics[width=\linewidth,height=3cm,keepaspectratio]{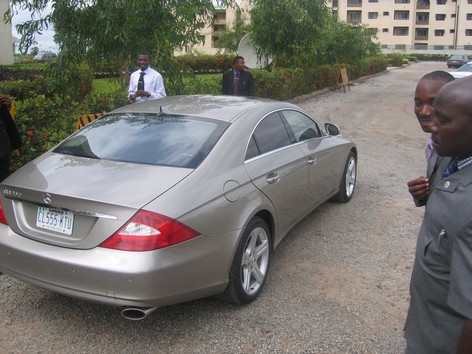}\par\vspace{2pt}
    \textbf{Q:} Describe the image.\newline
    \textbf{A:} A silver car parked on the side of a gravel road.\par\vspace{6pt}
    \includegraphics[width=\linewidth,height=3cm,keepaspectratio]{figures/samples_images/nocaps_val_226_real2}\par\vspace{2pt}
    \textbf{Q:} Describe the image.\newline
    \textbf{A:} A smiling man in a red shirt and smiling woman with long black curly hair with a water bottle next to them.\par\vspace{8pt}
    \textbf{RICES:} Two men playing guitars and drums on a street.\ \textcolor{green!60!black}{\ding{51}} {\scriptsize(CIDEr=0.625)}\par\vspace{3pt}
    \textbf{TTT-NN:} A man with a guitar and a sign that says EVK on it.\ \textcolor{red!70!black}{\ding{55}} {\scriptsize(CIDEr=0.233)}
    &
    \includegraphics[width=\linewidth,height=3cm,keepaspectratio]{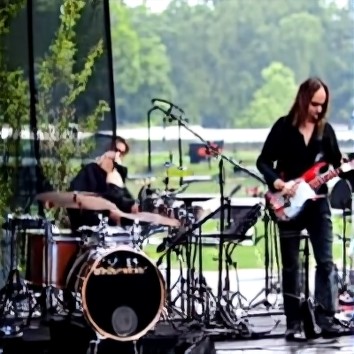}\par\vspace{2pt}
    \textbf{Q:} How many people are on stage?\newline
    \textbf{A:} Two\par\vspace{6pt}
    \includegraphics[width=\linewidth,height=3cm,keepaspectratio]{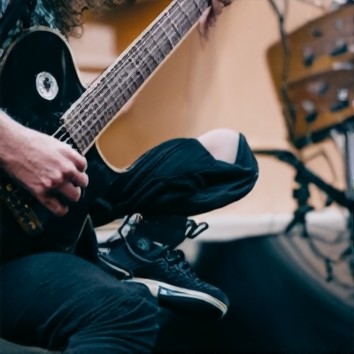}\par\vspace{2pt}
    \textbf{Q:} What type of shoe does the guitarist wear?\newline
    \textbf{A:} Sneakers\par\vspace{8pt}
    \textbf{SIMIT-ICL:} Two men sitting on the ground playing guitars.\ \textcolor{green!60!black}{\ding{51}} {\scriptsize(CIDEr=0.599)}\par\vspace{3pt}
    \textbf{SIMIT-FT (SQ):} Two men playing guitars in front of a white amp that says EVH.\ \textcolor{green!60!black}{\ding{51}} {\scriptsize(CIDEr=0.366)}
    \\

    
    \end{tabularx}
    \caption{Qualitative comparison of adaptation using retrieved real samples and self-generated synthetic samples. Both SIMIT-ICL and SIMIT-FT (SQ) improve on the pretrained model.}
    \label{fig:qual_ok_vqa_val2014_381}
\end{figure}

\begin{figure}[tbh]
    \centering
    \footnotesize
    \setlength{\tabcolsep}{4pt}
    \renewcommand{\arraystretch}{1.2}
    \renewcommand{\tabularxcolumn}[1]{m{#1}}
    \begin{tabularx}{\textwidth}{
        @{}
        >{\hsize=0.9375\hsize\linewidth=\hsize\raggedright\arraybackslash}X
        >{\hsize=1.03125\hsize\linewidth=\hsize\raggedright\arraybackslash}X
        >{\hsize=1.03125\hsize\linewidth=\hsize\raggedright\arraybackslash}X
        @{}
    }
    \textbf{Unlabeled test query} & \textbf{Retrieved real samples} & \textbf{Self-generated synthetic samples} \\
    \midrule
    \includegraphics[width=\linewidth,height=4.2cm,keepaspectratio]{figures/samples_images/coco2017_cap_val_387_query}\par\vspace{4pt}
    \textbf{Prompt:} \textit{Provide a one-sentence caption for the provided image.}\par\vspace{4pt}
    \textbf{GT:} three little kids that have different ties on, A jumping boy wearing three different neck ties, Three boys jumping in the air while wearing ties., Three boys dressed in jeans and ties pose., The three boys are creating their own dance moves.\par\vspace{4pt}
    \textbf{Pretrained:} A young man wearing a shirt and tie.\ \textcolor{red!70!black}{\ding{55}} {\scriptsize(CIDEr=0.086)}
    &
    \includegraphics[width=\linewidth,height=3cm,keepaspectratio]{figures/samples_images/coco2017_cap_val_387_real1}\par\vspace{2pt}
    \textbf{Q:} Please carefully observe the image and come up with a caption for the image. 
    \textbf{A:} A man swings at a tennis ball on a tennis court.\par\vspace{6pt}
    \includegraphics[width=\linewidth,height=3cm,keepaspectratio]{figures/samples_images/coco2017_cap_val_387_real2}\par\vspace{2pt}
    \textbf{Q:} Please carefully observe the image and come up with a caption for the image. 
    \textbf{A:} a close up of a person wearing a bow tie \par\vspace{8pt}
    \textbf{RICES:} A man is shown in three different poses.\ \textcolor{green!60!black}{\ding{51}} {\scriptsize(CIDEr=0.465)}\par\vspace{3pt}
    \textbf{TTT-NN:} A young man is shown in three different poses.\ \textcolor{green!60!black}{\ding{51}} {\scriptsize(CIDEr=0.443)}
    &
    \includegraphics[width=\linewidth,height=3cm,keepaspectratio]{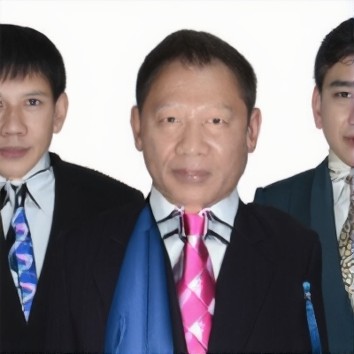}\par\vspace{2pt}
    \textbf{Q:} How many people are wearing ties?\newline
    \textbf{A:} Three\par\vspace{6pt}
    \includegraphics[width=\linewidth,height=3cm,keepaspectratio]{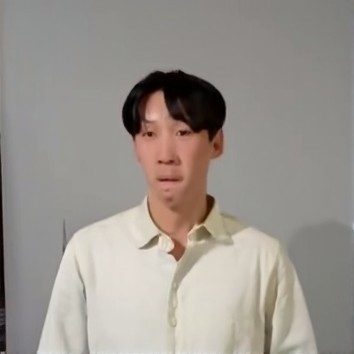}\par\vspace{2pt}
    \textbf{Q:} What color is the shirt worn by the person in the middle?\newline
    \textbf{A:} White\par\vspace{8pt}
    \textbf{SIMIT-ICL:} Three people wearing ties and jeans posing for a picture.\ \textcolor{green!60!black}{\ding{51}} {\scriptsize(CIDEr=0.973)}\par\vspace{3pt}
    \textbf{SIMIT-FT (SQ):} Three young men wearing ties and jeans posing for a picture.\ \textcolor{green!60!black}{\ding{51}} {\scriptsize(CIDEr=0.876)}
    \\

    \midrule
    
    \includegraphics[width=\linewidth,height=4.2cm,keepaspectratio]{figures/samples_images/vizwiz_vqa_val_345_query}\par\vspace{4pt}
    \textbf{Prompt:} \textit{What color is this cat? \newline
     When the provided information is insufficient, respond with `Unanswerable'. \newline
     Answer the question using a single word or phrase.}\par\vspace{4pt}
    \textbf{GT:} brown, tan, grey black, sable\par\vspace{4pt}
    \textbf{Pretrained:} Unanswerable\ \textcolor{red!70!black}{\ding{55}} {\scriptsize(exact\_match=0.00)}
    &
    \includegraphics[width=\linewidth,height=3cm,keepaspectratio]{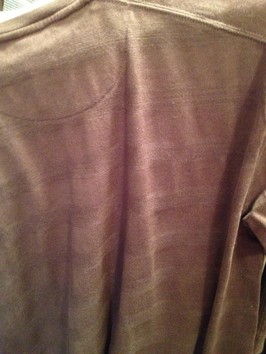}\par\vspace{2pt}
    \textbf{Q:} What color is this shirt?\newline
    \textbf{A:} tan\par\vspace{6pt}
    \includegraphics[width=\linewidth,height=3cm,keepaspectratio]{figures/samples_images/vizwiz_vqa_val_345_real2}\par\vspace{2pt}
    \textbf{Q:} What color are my shoes?\newline
    \textbf{A:} brown\par\vspace{8pt}
    \textbf{RICES:} brown\ \textcolor{green!60!black}{\ding{51}} {\scriptsize(exact\_match=1.00)}\par\vspace{3pt}
    \textbf{TTT-NN:} tan\ \textcolor{red!70!black}{\ding{55}} {\scriptsize(exact\_match=0.30)}
    &
    \includegraphics[width=\linewidth,height=3cm,keepaspectratio]{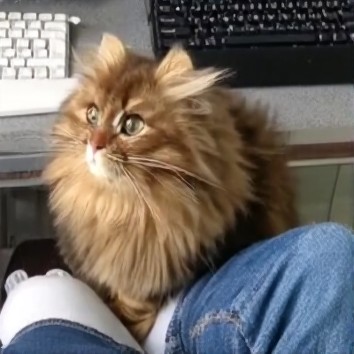}\par\vspace{2pt}
    \textbf{Q:} How many legs does the cat have? 
    \textbf{A:} Four\par\vspace{6pt}
    \includegraphics[width=\linewidth,height=3cm,keepaspectratio]{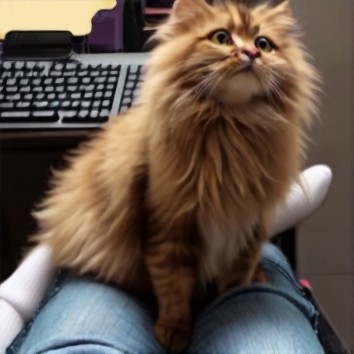}\par\vspace{2pt}
    \textbf{Q:} Is the cat sitting or lying down?\newline
    \textbf{A:} Sitting\par\vspace{8pt}
    \textbf{SIMIT-ICL:} Brown.\ \textcolor{green!60!black}{\ding{51}} {\scriptsize(exact\_match=1.00)}\par\vspace{3pt}
    \textbf{SIMIT-FT (SQ):} Brown\ \textcolor{green!60!black}{\ding{51}} {\scriptsize(exact\_match=1.00)}
    \\
    \end{tabularx}
    \caption{Qualitative comparison of adaptation using retrieved real samples and self-generated synthetic samples. Both SIMIT-ICL and SIMIT-FT (SQ) improve on the pretrained model.}
    \label{fig:qual_vizwiz_vqa_val_87}
\end{figure}

\begin{figure}[tbh]
    \centering
    \footnotesize
    \setlength{\tabcolsep}{4pt}
    \renewcommand{\arraystretch}{1.2}
    \renewcommand{\tabularxcolumn}[1]{m{#1}}
    \begin{tabularx}{\textwidth}{
        @{}
        >{\hsize=0.9375\hsize\linewidth=\hsize\raggedright\arraybackslash}X
        >{\hsize=1.03125\hsize\linewidth=\hsize\raggedright\arraybackslash}X
        >{\hsize=1.03125\hsize\linewidth=\hsize\raggedright\arraybackslash}X
        @{}
    }
    \textbf{Unlabeled test query} & \textbf{Retrieved real samples} & \textbf{Self-generated synthetic samples} \\
    \midrule

    \includegraphics[width=\linewidth,height=4.2cm,keepaspectratio]{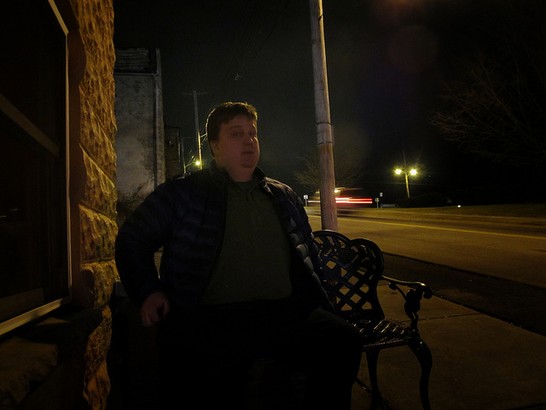}\par\vspace{4pt}
    \textbf{Prompt:} \textit{What is the light source? \newline
     When the provided information is insufficient, respond with `Unanswerable'. \newline
     Answer the question using a single word or phrase.}\par\vspace{4pt}
    \textbf{GT:} street light, streetlight\par\vspace{4pt}
    \textbf{Pretrained:} Unanswerable\ \textcolor{red!70!black}{\ding{55}} {\scriptsize(exact\_match=0.00)}
    &
    \includegraphics[width=\linewidth,height=3cm,keepaspectratio]{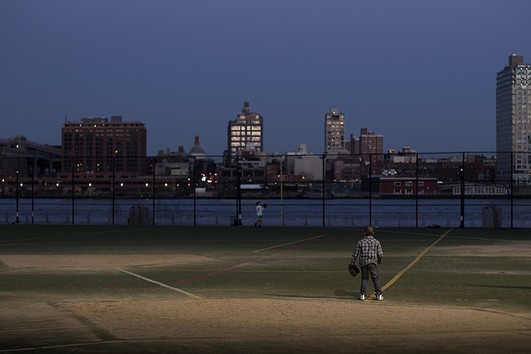}\par\vspace{2pt}
    \textbf{Q:} How is a parking lot illuminated at night?\newline
    \textbf{A:} lights\par\vspace{6pt}
    \includegraphics[width=\linewidth,height=3cm,keepaspectratio]{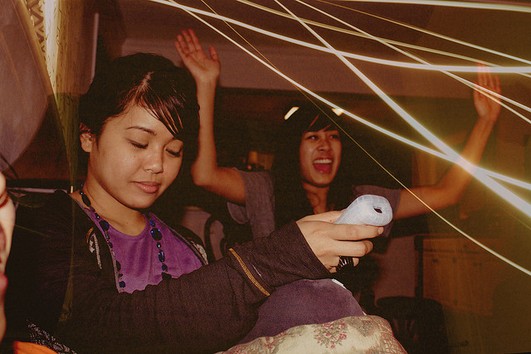}\par\vspace{2pt}
    \textbf{Q:} Where are the strands coming from?\newline
    \textbf{A:} ceiling\par\vspace{8pt}
    \textbf{RICES:} street light\ \textcolor{green!60!black}{\ding{51}} {\scriptsize(exact\_match=1.00)}\par\vspace{3pt}
    \textbf{TTT-NN:} lights\ \textcolor{red!70!black}{\ding{55}} {\scriptsize(exact\_match=0.00)}
    &
    \includegraphics[width=\linewidth,height=3cm,keepaspectratio]{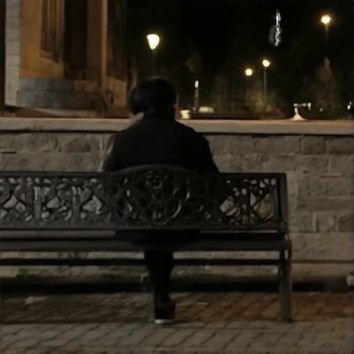}\par\vspace{2pt}
    \textbf{Q:} How many people are sitting on the bench?\newline
    \textbf{A:} One\par\vspace{6pt}
    \includegraphics[width=\linewidth,height=3cm,keepaspectratio]{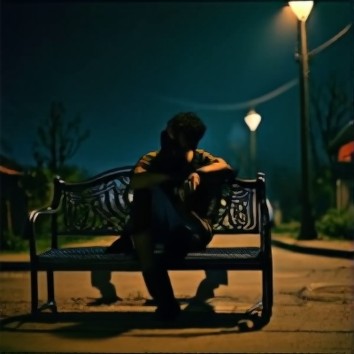}\par\vspace{2pt}
    \textbf{Q:} How many people are sitting on the bench?\newline
    \textbf{A:} One\par\vspace{8pt}
    \textbf{SIMIT-ICL:} Street light\ \textcolor{green!60!black}{\ding{51}} {\scriptsize(exact\_match=1.00)}\par\vspace{3pt}
    \textbf{SIMIT-FT (SQ):} Street light\ \textcolor{green!60!black}{\ding{51}} {\scriptsize(exact\_match=1.00)}
    \\


    
    \midrule
    \includegraphics[width=\linewidth,height=4.2cm,keepaspectratio]{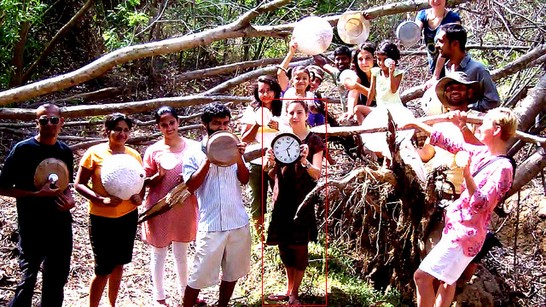}\par\vspace{4pt}
    \textbf{Prompt:} \textit{Provide a short description for this region.}\par\vspace{4pt}
    \textbf{GT:} woman holding clock in middle black dress, middle holding the clock, lady holding up clock\par\vspace{4pt}
    \textbf{Pretrained:} A person is standing up.\ \textcolor{red!70!black}{\ding{55}} {\scriptsize(CIDEr=0.222)}
    &
    \includegraphics[width=\linewidth,height=3cm,keepaspectratio]{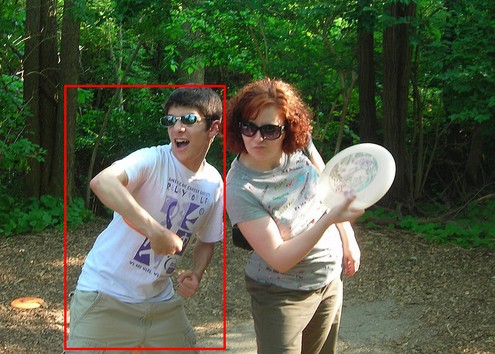}\par\vspace{2pt}
    \textbf{Q:} Please carefully observe the area circled in the image and come up with a caption for the area.\newline
    \textbf{A:} person on left\par\vspace{6pt}
    \includegraphics[width=\linewidth,height=3cm,keepaspectratio]{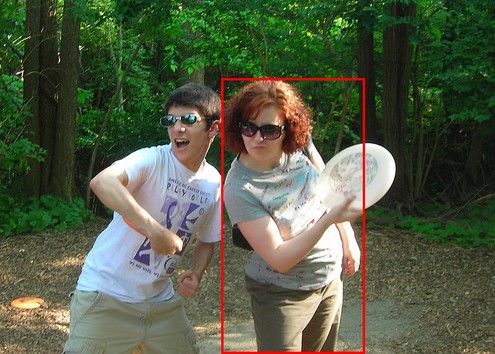}\par\vspace{2pt}
    \textbf{Q:} Please carefully observe the area circled in the image and come up with a caption for the area.\newline
    \textbf{A:} red head\par\vspace{8pt}
    \textbf{RICES:} A person is standing up\ \textcolor{red!70!black}{\ding{55}} {\scriptsize(CIDEr=0.222)}\par\vspace{3pt}
    \textbf{TTT-NN:} person\ \textcolor{red!70!black}{\ding{55}} {\scriptsize(CIDEr=0.000)}
    &
    \includegraphics[width=\linewidth,height=3cm,keepaspectratio]{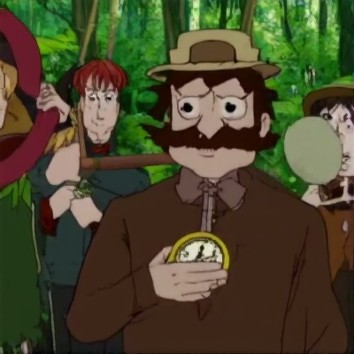}\par\vspace{2pt}
    \textbf{Q:} How many people are holding clocks?\newline
    \textbf{A:} One\par\vspace{6pt}
    \includegraphics[width=\linewidth,height=3cm,keepaspectratio]{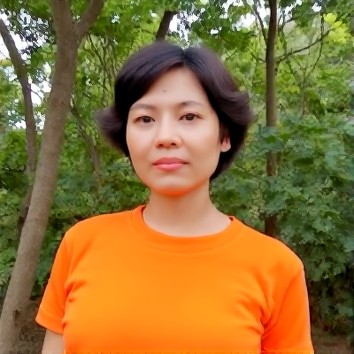}\par\vspace{2pt}
    \textbf{Q:} What is the dominant color of the clothing worn by the person in the center-left of the image?\newline
    \textbf{A:} Orange\par\vspace{8pt}
    \textbf{SIMIT-ICL:} A woman holding a clock.\ \textcolor{green!60!black}{\ding{51}} {\scriptsize(CIDEr=1.361)}\par\vspace{3pt}
    \textbf{SIMIT-FT (SQ):} A woman holding a clock.\ \textcolor{green!60!black}{\ding{51}} {\scriptsize(CIDEr=1.361)}
    \\
    \end{tabularx}
    \caption{Qualitative comparison of adaptation using retrieved real samples and self-generated synthetic samples. Both SIMIT-ICL and SIMIT-FT (SQ) improve on the pretrained model.}
    \label{fig:qual_nocaps_val_226}
\end{figure}

\begin{figure}[tbh]
    \centering
    \footnotesize
    \setlength{\tabcolsep}{4pt}
    \renewcommand{\arraystretch}{1.2}
    \renewcommand{\tabularxcolumn}[1]{m{#1}}
    \begin{tabularx}{\textwidth}{
        @{}
        >{\hsize=0.9375\hsize\linewidth=\hsize\raggedright\arraybackslash}X
        >{\hsize=1.03125\hsize\linewidth=\hsize\raggedright\arraybackslash}X
        >{\hsize=1.03125\hsize\linewidth=\hsize\raggedright\arraybackslash}X
        @{}
    }
    \textbf{Unlabeled test query} & \textbf{Retrieved real samples} & \textbf{Self-generated synthetic samples} \\
    \midrule
    \includegraphics[width=\linewidth,height=4.2cm,keepaspectratio]{figures/samples_images/nocaps_val_45_query}\par\vspace{4pt}
    \textbf{Prompt:} \textit{Provide a one-sentence caption for the provided image.}\par\vspace{4pt}
    \textbf{GT:} A happy man in a sailor outside has a pitcher of milk, A man in a white hat holding a glass of milk., A man in a white uniform and hat holding a glass of milk., A friendly man in a white unifom holds a glass of milk., A white man in a white suit holds up a glass of white liquid., Actor of a milk man, dressed in all whites and a hat, drinking a mug of milk.\par\vspace{4pt}
    \textbf{Pretrained:} A man in a white shirt with the word Oberweis on it.\ \textcolor{red!70!black}{\ding{55}} {\scriptsize(CIDEr=0.301)}
    &
    \includegraphics[width=\linewidth,height=3cm,keepaspectratio]{figures/samples_images/nocaps_val_45_real1}\par\vspace{2pt}
    \textbf{Q:} Describe the image.\newline
    \textbf{A:} The man has a drink and using his human face for his emotions, and the person beside of him is smiling. \par\vspace{6pt}
    \includegraphics[width=\linewidth,height=3cm,keepaspectratio]{figures/samples_images/nocaps_val_45_real2}\par\vspace{2pt}
    \textbf{Q:} Describe the image. 
    \textbf{A:} A man with a mustache wearing a white hat. \par\vspace{8pt}
    \textbf{RICES:} A man with a mustache holding a glass of milk.\ \textcolor{green!60!black}{\ding{51}} {\scriptsize(CIDEr=2.251)}\par\vspace{3pt}
    \textbf{TTT-NN:} A man in a white shirt with "OBERWEIS DAIRY" on it is holding a glass of milk.\ \textcolor{green!60!black}{\ding{51}} {\scriptsize(CIDEr=1.698)}
    &
    \includegraphics[width=\linewidth,height=3cm,keepaspectratio]{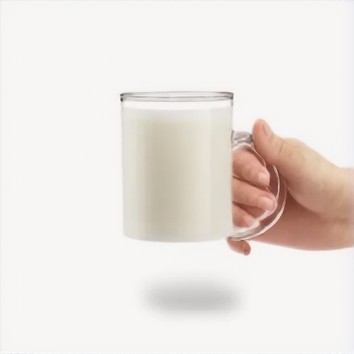}\par\vspace{2pt}
    \textbf{Q:} What is the person holding?\newline
    \textbf{A:} A glass mug filled with milk.\par\vspace{6pt}
    \includegraphics[width=\linewidth,height=3cm,keepaspectratio]{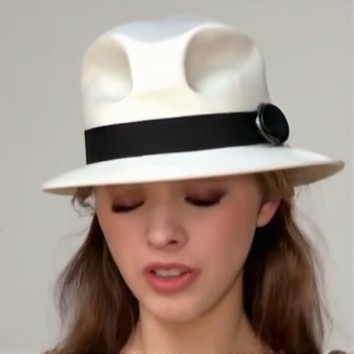}\par\vspace{2pt}
    \textbf{Q:} What is on the person's head?\newline
    \textbf{A:} A white hat with a black band.\par\vspace{8pt}
    \textbf{SIMIT-ICL:} A man wearing a white hat and holding a glass mug filled with milk.\ \textcolor{green!60!black}{\ding{51}} {\scriptsize(CIDEr=1.565)}\par\vspace{3pt}
    \textbf{SIMIT-FT (SQ):} A man in a white shirt with Oberweis Dairy on the pocket is holding a glass mug of milk.\ \textcolor{green!60!black}{\ding{51}} {\scriptsize(CIDEr=0.941)}
    \\



    \midrule
    
    \includegraphics[width=\linewidth,height=4.2cm,keepaspectratio]{figures/samples_images/vizwiz_vqa_val_47_query}\par\vspace{4pt}
    \textbf{Prompt:} \textit{What is this item? \newline
     When the provided information is insufficient, respond with `Unanswerable'. \newline
     Answer the question using a single word or phrase.}\par\vspace{4pt}
    \textbf{GT:} angel, angel statuette\par\vspace{4pt}
    \textbf{Pretrained:} Unanswerable.\ \textcolor{red!70!black}{\ding{55}} {\scriptsize(exact\_match=0.00)}
    &
    \includegraphics[width=\linewidth,height=3cm,keepaspectratio]{figures/samples_images/vizwiz_vqa_val_47_real1}\par\vspace{2pt}
    \textbf{Q:} What is this item?\newline
    \textbf{A:} plastic pig another on his back\par\vspace{6pt}
    \includegraphics[width=\linewidth,height=3cm,keepaspectratio]{figures/samples_images/vizwiz_vqa_val_47_real2}\par\vspace{2pt}
    \textbf{Q:} What is this?\newline
    \textbf{A:} teddy bear rose\par\vspace{8pt}
    \textbf{RICES:} angel\ \textcolor{green!60!black}{\ding{51}} {\scriptsize(exact\_match=1.00)}\par\vspace{3pt}
    \textbf{TTT-NN:} angel cat\ \textcolor{red!70!black}{\ding{55}} {\scriptsize(exact\_match=0.00)}
    &
    \includegraphics[width=\linewidth,height=3cm,keepaspectratio]{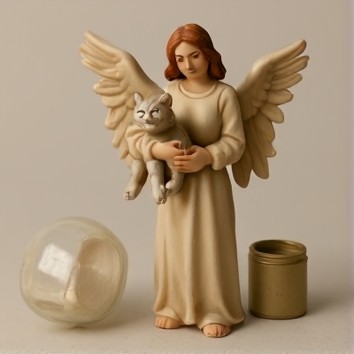}\par\vspace{2pt}
    \textbf{Q:} How many objects are visible in the image? 
    \textbf{A:} Three\par\vspace{6pt}
    \includegraphics[width=\linewidth,height=3cm,keepaspectratio]{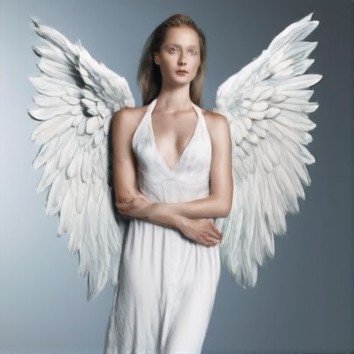}\par\vspace{2pt}
    \textbf{Q:} What is the primary color of the angel's wings? 
    \textbf{A:} White\par\vspace{8pt}
    \textbf{SIMIT-ICL:} Angel\ \textcolor{green!60!black}{\ding{51}} {\scriptsize(exact\_match=1.00)}\par\vspace{3pt}
    \textbf{SIMIT-FT (SQ):} Angel.\ \textcolor{green!60!black}{\ding{51}} {\scriptsize(exact\_match=1.00)}
    \\
    \end{tabularx}
    \caption{Qualitative comparison of adaptation using retrieved real samples and self-generated synthetic samples. Both SIMIT-ICL and SIMIT-FT (SQ) improve on the pretrained model.}
    \label{fig:qual_coco2017_cap_val_387}
\end{figure}

\begin{figure}[tbh]
    \centering
    \footnotesize
    \setlength{\tabcolsep}{4pt}
    \renewcommand{\arraystretch}{1.2}
    \renewcommand{\tabularxcolumn}[1]{m{#1}}
    \begin{tabularx}{\textwidth}{
        @{}
        >{\hsize=0.9375\hsize\linewidth=\hsize\raggedright\arraybackslash}X
        >{\hsize=1.03125\hsize\linewidth=\hsize\raggedright\arraybackslash}X
        >{\hsize=1.03125\hsize\linewidth=\hsize\raggedright\arraybackslash}X
        @{}
    }
    \textbf{Unlabeled test query} & \textbf{Retrieved real samples} & \textbf{Self-generated synthetic samples} \\
    \midrule
    \includegraphics[width=\linewidth,height=4.2cm,keepaspectratio]{figures/samples_images/coco2017_cap_val_329_query}\par\vspace{4pt}
    \textbf{Prompt:} \textit{Provide a one-sentence caption for the provided image.}\par\vspace{4pt}
    \textbf{GT:} A group of foot ball players standing on top of a football field., Players and a referee playing on a football field., a couple of football players in the air, Two football players reach for the ball at the same time., A receiver and a defender jump for the ball as a referee looks on during a football game.\par\vspace{4pt}
    \textbf{Pretrained:} A football player jumping up to catch a football.\ \textcolor{red!70!black}{\ding{55}} {\scriptsize(CIDEr=0.609)}
    &
    \includegraphics[width=\linewidth,height=3cm,keepaspectratio]{figures/samples_images/coco2017_cap_val_329_real1}\par\vspace{2pt}
    \textbf{Q:} Please carefully observe the image and come up with a caption for the image. 
    \textbf{A:} A little league baseball player hits a ball.\par\vspace{6pt}
    \includegraphics[width=\linewidth,height=3cm,keepaspectratio]{figures/samples_images/coco2017_cap_val_329_real2}\par\vspace{2pt}
    \textbf{Q:} Please carefully observe the image and come up with a caption for the image. 
    \textbf{A:} A group of women playing a game of soccer on field.\par\vspace{8pt}
    \textbf{RICES:} A football player jumping in the air to catch a football.\ \textcolor{red!70!black}{\ding{55}} {\scriptsize(CIDEr=0.729)}\par\vspace{3pt}
    \textbf{TTT-NN:} A group of men playing a game of football on field.\ \textcolor{red!70!black}{\ding{55}} {\scriptsize(CIDEr=1.151)}
    &
    \includegraphics[width=\linewidth,height=3cm,keepaspectratio]{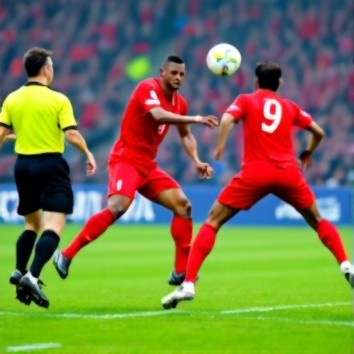}\par\vspace{2pt}
    \textbf{Q:} How many players are in mid-air?\newline
    \textbf{A:} Two\par\vspace{6pt}
    \includegraphics[width=\linewidth,height=3cm,keepaspectratio]{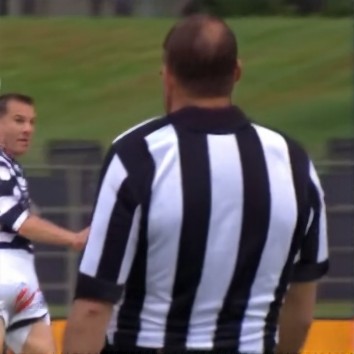}\par\vspace{2pt}
    \textbf{Q:} What is the player wearing number 38 doing?\newline
    \textbf{A:} Refereeing\par\vspace{8pt}
    \textbf{SIMIT-ICL:} A referee watching two players jump for a football.\ \textcolor{green!60!black}{\ding{51}} {\scriptsize(CIDEr=1.356)}\par\vspace{3pt}
    \textbf{SIMIT-FT (SQ):} A group of football players on a field.\ \textcolor{green!60!black}{\ding{51}} {\scriptsize(CIDEr=1.798)}
    \\



    \midrule
    \includegraphics[width=\linewidth,height=4.2cm,keepaspectratio]{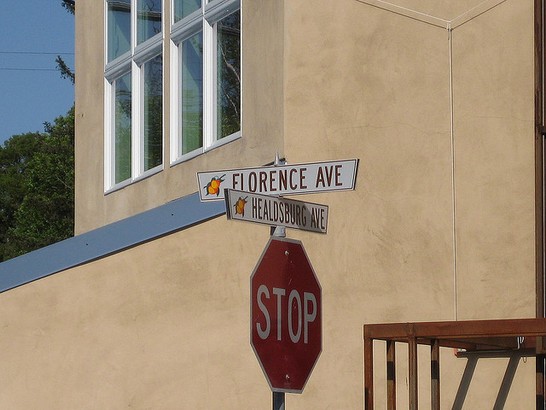}\par\vspace{4pt}
    \textbf{Prompt:} \textit{What kind of house is next to the stop sign? \newline
     When the provided information is insufficient, respond with `Unanswerable'. \newline
     Answer the question using a single word or phrase.}\par\vspace{4pt}
    \textbf{GT:} apartment, condo\par\vspace{4pt}
    \textbf{Pretrained:} Unanswerable\ \textcolor{red!70!black}{\ding{55}} {\scriptsize(exact\_match=0.00)}
    &
    \includegraphics[width=\linewidth,height=3cm,keepaspectratio]{figures/samples_images/ok_vqa_val2014_8_real1}\par\vspace{2pt}
    \textbf{Q:} What type of kitchen is this?\newline
    \textbf{A:} galley\par\vspace{6pt}
    \includegraphics[width=\linewidth,height=3cm,keepaspectratio]{figures/samples_images/ok_vqa_val2014_8_real2}\par\vspace{2pt}
    \textbf{Q:} What style of architecture is this building? 
    \textbf{A:} modern\par\vspace{5pt}
    \textbf{RICES:} apartment\ \textcolor{green!60!black}{\ding{51}} {\scriptsize(exact\_match=1.00)}\par\vspace{3pt}
    \textbf{TTT-NN:} galaxy\ \textcolor{red!70!black}{\ding{55}} {\scriptsize(exact\_match=0.00)}
    &
    \includegraphics[width=\linewidth,height=3cm,keepaspectratio]{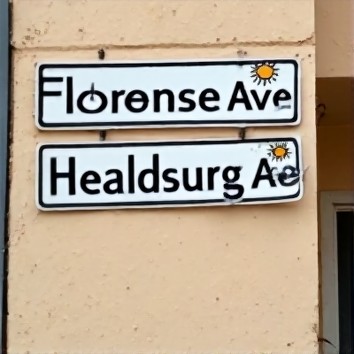}\par\vspace{2pt}
    \textbf{Q:} How many street signs are visible in the image? 
    \textbf{A:} Two\par\vspace{6pt}
    \includegraphics[width=\linewidth,height=3cm,keepaspectratio]{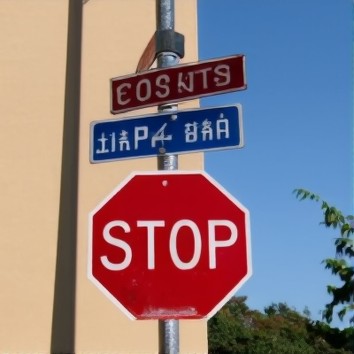}\par\vspace{2pt}
    \textbf{Q:} What is the color of the stop sign in the image? 
    \textbf{A:} Red\par\vspace{5pt}
    \textbf{SIMIT-ICL:} apartment\ \textcolor{green!60!black}{\ding{51}} {\scriptsize(exact\_match=1.00)}\par\vspace{3pt}
    \textbf{SIMIT-FT (SQ):} Apartment\ \textcolor{green!60!black}{\ding{51}} {\scriptsize(exact\_match=1.00)}
    \\
    \end{tabularx}
    \caption{Qualitative comparison of adaptation using retrieved real samples and self-generated synthetic samples. Both SIMIT-ICL and SIMIT-FT (SQ) improve on the pretrained model.}
    \label{fig:qual_ok_vqa_val2014_8}
\end{figure}

\begin{figure}[tbh]
    \centering
    \footnotesize
    \setlength{\tabcolsep}{4pt}
    \renewcommand{\arraystretch}{1.2}
    \renewcommand{\tabularxcolumn}[1]{m{#1}}
    \begin{tabularx}{\textwidth}{
        @{}
        >{\hsize=0.9375\hsize\linewidth=\hsize\raggedright\arraybackslash}X
        >{\hsize=1.03125\hsize\linewidth=\hsize\raggedright\arraybackslash}X
        >{\hsize=1.03125\hsize\linewidth=\hsize\raggedright\arraybackslash}X
        @{}
    }
    \textbf{Unlabeled test query} & \textbf{Retrieved real samples} & \textbf{Self-generated synthetic samples} \\
    
    \midrule
    \includegraphics[width=\linewidth,height=4.2cm,keepaspectratio]{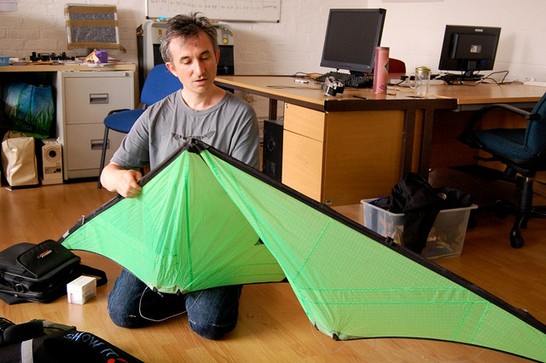}\par\vspace{4pt}
    \textbf{Prompt:} \textit{Where would he use this? \newline
     When the provided information is insufficient, respond with `Unanswerable'. \newline
     Answer the question using a single word or phrase.}\par\vspace{4pt}
    \textbf{GT:} outside\par\vspace{4pt}
    \textbf{Pretrained:} Unanswerable\ \textcolor{red!70!black}{\ding{55}} {\scriptsize(exact\_match=0.00)}
    &
    \includegraphics[width=\linewidth,height=3cm,keepaspectratio]{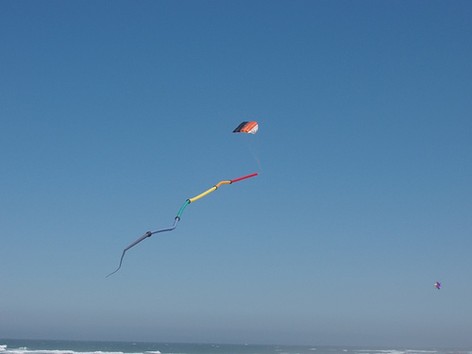}\par\vspace{2pt}
    \textbf{Q:} Where are these kites?\newline
    \textbf{A:} sky\par\vspace{6pt}
    \includegraphics[width=\linewidth,height=3cm,keepaspectratio]{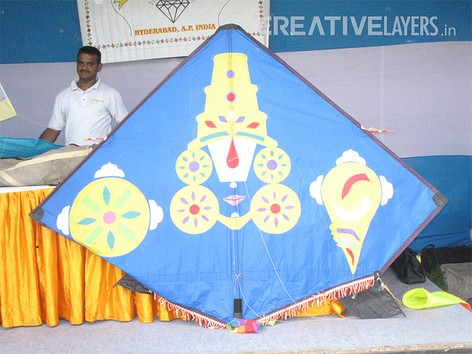}\par\vspace{2pt}
    \textbf{Q:} What does this man fly for fun?\newline
    \textbf{A:} kite\par\vspace{8pt}
    \textbf{RICES:} sky\ \textcolor{red!70!black}{\ding{55}} {\scriptsize(exact\_match=0.00)}\par\vspace{3pt}
    \textbf{TTT-NN:} sky\ \textcolor{red!70!black}{\ding{55}} {\scriptsize(exact\_match=0.00)}
    &
    \includegraphics[width=\linewidth,height=3cm,keepaspectratio]{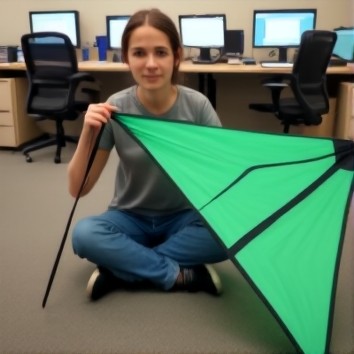}\par\vspace{2pt}
    \textbf{Q:} How many people are in the image?\newline
    \textbf{A:} One person.\par\vspace{6pt}
    \includegraphics[width=\linewidth,height=3cm,keepaspectratio]{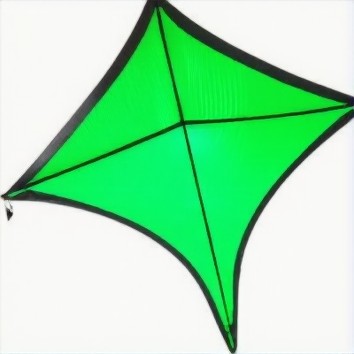}\par\vspace{2pt}
    \textbf{Q:} What color is the kite?\newline
    \textbf{A:} Green.\par\vspace{8pt}
    \textbf{SIMIT-ICL:} Outside.\ \textcolor{green!60!black}{\ding{51}} {\scriptsize(exact\_match=1.00)}\par\vspace{3pt}
    \textbf{SIMIT-FT (SQ):} Outside.\ \textcolor{green!60!black}{\ding{51}} {\scriptsize(exact\_match=1.00)}
    \\

    \midrule
    \includegraphics[width=\linewidth,height=4.2cm,keepaspectratio]{figures/samples_images/vizwiz_vqa_val_87_query}\par\vspace{4pt}
    \textbf{Prompt:} \textit{What is this a picture of? \newline
     When the provided information is insufficient, respond with `Unanswerable'. \newline
     Answer the question using a single word or phrase.}\par\vspace{4pt}
    \textbf{GT:} dog in lace dress, dog\par\vspace{4pt}
    \textbf{Pretrained:} Unanswerable.\ \textcolor{red!70!black}{\ding{55}} {\scriptsize(exact\_match=0.00)}
    &
    \includegraphics[width=\linewidth,height=3cm,keepaspectratio]{figures/samples_images/vizwiz_vqa_val_87_real1}\par\vspace{2pt}
    \textbf{Q:} What kind of dog is this? 
    \textbf{A:} tan\par\vspace{6pt}
    \includegraphics[width=\linewidth,height=3cm,keepaspectratio]{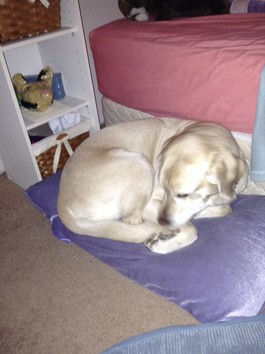}\par\vspace{2pt}
    \textbf{Q:} What is this? 
    \textbf{A:} yellow labrador on purple cushion\par\vspace{8pt}
    \textbf{RICES:} dog\ \textcolor{green!60!black}{\ding{51}} {\scriptsize(exact\_match=1.00)}\par\vspace{3pt}
    \textbf{TTT-NN:} tan and white dog\ \textcolor{red!70!black}{\ding{55}} {\scriptsize(exact\_match=0.00)}
    &
    \includegraphics[width=\linewidth,height=3cm,keepaspectratio]{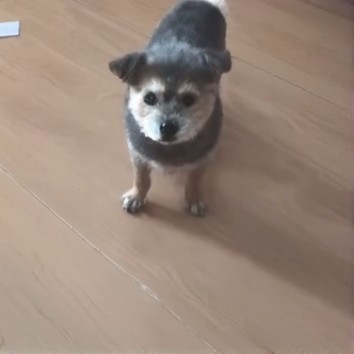}\par\vspace{2pt}
    \textbf{Q:} What color is the belly of the animal in the picture? 
    \textbf{A:} Pink\par\vspace{6pt}
    \includegraphics[width=\linewidth,height=3cm,keepaspectratio]{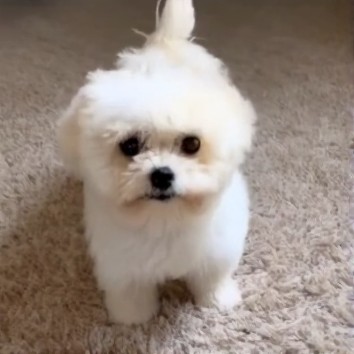}\par\vspace{2pt}
    \textbf{Q:} Is the animal in the picture a cat or a dog? 
    \textbf{A:} Dog\par\vspace{8pt}
    \textbf{SIMIT-ICL:} Dog\ \textcolor{green!60!black}{\ding{51}} {\scriptsize(exact\_match=1.00)}\par\vspace{3pt}
    \textbf{SIMIT-FT (SQ):} Dog.\ \textcolor{green!60!black}{\ding{51}} {\scriptsize(exact\_match=1.00)}
    \\

    
    \end{tabularx}
    \caption{Qualitative comparison of adaptation using retrieved real samples and self-generated synthetic samples. Both SIMIT-ICL and SIMIT-FT (SQ) improve on the pretrained model.}
    \label{fig:qual_ok_vqa_val2014_22}
\end{figure}

\begin{figure}[tbh]
    \centering
    \footnotesize
    \setlength{\tabcolsep}{4pt}
    \renewcommand{\arraystretch}{1.2}
    \renewcommand{\tabularxcolumn}[1]{m{#1}}
    \begin{tabularx}{\textwidth}{
        @{}
        >{\hsize=0.9375\hsize\linewidth=\hsize\raggedright\arraybackslash}X
        >{\hsize=1.03125\hsize\linewidth=\hsize\raggedright\arraybackslash}X
        >{\hsize=1.03125\hsize\linewidth=\hsize\raggedright\arraybackslash}X
        @{}
    }
    \textbf{Unlabeled test query} & \textbf{Retrieved real samples} & \textbf{Self-generated synthetic samples} \\


    \midrule
    \includegraphics[width=\linewidth,height=4.2cm,keepaspectratio]{figures/samples_images/ok_vqa_val2014_477_query}\par\vspace{4pt}
    \textbf{Prompt:} \textit{What art is this? \newline
     When the provided information is insufficient, respond with `Unanswerable'. \newline
     Answer the question using a single word or phrase.}\par\vspace{4pt}
    \textbf{GT:} pottery\par\vspace{4pt}
    \textbf{Pretrained:} Unanswerable\ \textcolor{red!70!black}{\ding{55}} {\scriptsize(exact\_match=0.00)}
    &
    \includegraphics[width=\linewidth,height=3cm,keepaspectratio]{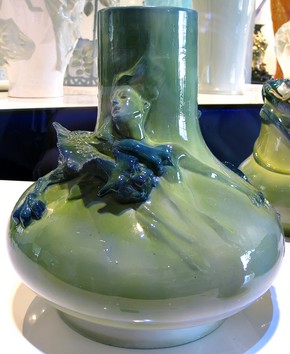}\par\vspace{2pt}
    \textbf{Q:} What is this vase made of?\newline
    \textbf{A:} glass\par\vspace{6pt}
    \includegraphics[width=\linewidth,height=3cm,keepaspectratio]{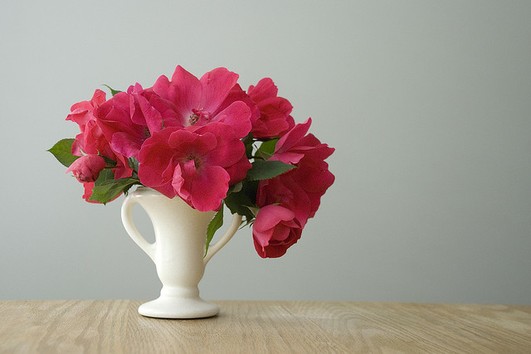}\par\vspace{2pt}
    \textbf{Q:} What material is the receptacle holding the flowers made from?\newline
    \textbf{A:} porcelain\par\vspace{8pt}
    \textbf{RICES:} Unanswerable\ \textcolor{red!70!black}{\ding{55}} {\scriptsize(exact\_match=0.00)}\par\vspace{3pt}
    \textbf{TTT-NN:} porcelain\ \textcolor{red!70!black}{\ding{55}} {\scriptsize(exact\_match=0.00)}
    &
    \includegraphics[width=\linewidth,height=3cm,keepaspectratio]{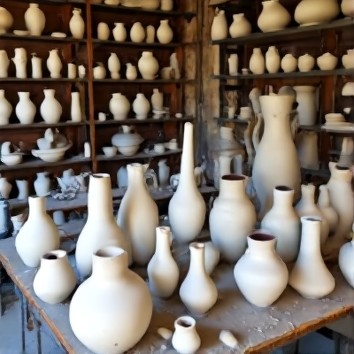}\par\vspace{2pt}
    \textbf{Q:} How many vases are visible in the image?\newline
    \textbf{A:} Over 20\par\vspace{6pt}
    \includegraphics[width=\linewidth,height=3cm,keepaspectratio]{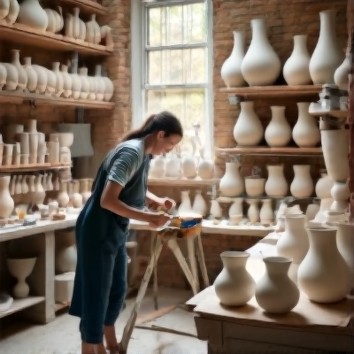}\par\vspace{2pt}
    \textbf{Q:} What is the primary activity taking place in the image?\newline
    \textbf{A:} Pottery making\par\vspace{8pt}
    \textbf{SIMIT-ICL:} Pottery\ \textcolor{green!60!black}{\ding{51}} {\scriptsize(exact\_match=1.00)}\par\vspace{3pt}
    \textbf{SIMIT-FT (SQ):} Pottery\ \textcolor{green!60!black}{\ding{51}} {\scriptsize(exact\_match=1.00)}
    \\

    \midrule
    \includegraphics[width=\linewidth,height=4.2cm,keepaspectratio]{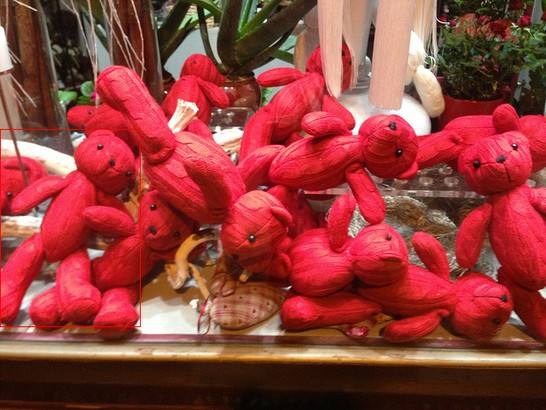}\par\vspace{4pt}
    \textbf{Prompt:} \textit{Provide a short description for this region.}\par\vspace{4pt}
    \textbf{GT:} left most red bear standing, left full bear, left most creature\par\vspace{4pt}
    \textbf{Pretrained:} Leftest one.\ \textcolor{red!70!black}{\ding{55}} {\scriptsize(CIDEr=0.000)}
    &
    \includegraphics[width=\linewidth,height=3cm,keepaspectratio]{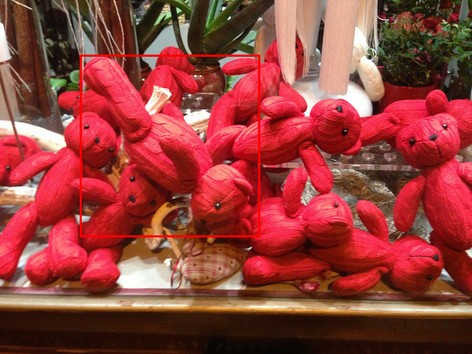}\par\vspace{2pt}
    \textbf{Q:} Please carefully observe the area circled in the image and come up with a caption for the area. 
    \textbf{A:} larger bear, upside down center leftish\par\vspace{6pt}
    \includegraphics[width=\linewidth,height=3cm,keepaspectratio]{figures/samples_images/refcoco_bbox_val_211_real2}\par\vspace{2pt}
    \textbf{Q:} Please carefully observe the area circled in the image and come up with a caption for the area. 
    \textbf{A:} plaid bear on left\par\vspace{8pt}
    \textbf{RICES:} Red bear on the left.\ \textcolor{green!60!black}{\ding{51}} {\scriptsize(CIDEr=1.289)}\par\vspace{3pt}
    \textbf{TTT-NN:} left bear\ \textcolor{green!60!black}{\ding{51}} {\scriptsize(CIDEr=0.991)}
    &
    \includegraphics[width=\linewidth,height=3cm,keepaspectratio]{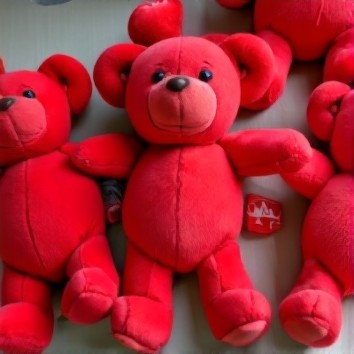}\par\vspace{2pt}
    \textbf{Q:} What is the dominant color of the objects in the image?\newline
    \textbf{A:} Red\par\vspace{6pt}
    \includegraphics[width=\linewidth,height=3cm,keepaspectratio]{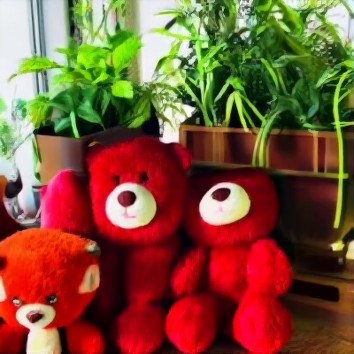}\par\vspace{2pt}
    \textbf{Q:} Are there any plants visible in the background?\newline
    \textbf{A:} Yes\par\vspace{8pt}
    \textbf{SIMIT-ICL:} A red teddy bear.\ \textcolor{green!60!black}{\ding{51}} {\scriptsize(CIDEr=0.625)}\par\vspace{3pt}
    \textbf{SIMIT-FT (SQ):} Red bear on the left.\ \textcolor{green!60!black}{\ding{51}} {\scriptsize(CIDEr=1.289)}
    \\
    \end{tabularx}
    \caption{Qualitative comparison of adaptation using retrieved real samples and self-generated synthetic samples. Both SIMIT-ICL and SIMIT-FT (SQ) improve on the pretrained model.}
    \label{fig:qual_ok_vqa_val2014_477}
\end{figure}

\begin{figure}[tbh]
    \centering
    \footnotesize
    \setlength{\tabcolsep}{4pt}
    \renewcommand{\arraystretch}{1.2}
    \renewcommand{\tabularxcolumn}[1]{m{#1}}
    \begin{tabularx}{\textwidth}{
        @{}
        >{\hsize=0.9375\hsize\linewidth=\hsize\raggedright\arraybackslash}X
        >{\hsize=1.03125\hsize\linewidth=\hsize\raggedright\arraybackslash}X
        >{\hsize=1.03125\hsize\linewidth=\hsize\raggedright\arraybackslash}X
        @{}
    }
    \textbf{Unlabeled test query} & \textbf{Retrieved real samples} & \textbf{Self-generated synthetic samples} \\
    \midrule
    \includegraphics[width=\linewidth,height=4.2cm,keepaspectratio]{figures/samples_images/ok_vqa_val2014_115_query}\par\vspace{4pt}
    \textbf{Prompt:} \textit{The man is opening what? \newline
     When the provided information is insufficient, respond with `Unanswerable'. \newline
     Answer the question using a single word or phrase.}\par\vspace{4pt}
    \textbf{GT:} umbrella\par\vspace{4pt}
    \textbf{Pretrained:} Unanswerable\ \textcolor{red!70!black}{\ding{55}} {\scriptsize(exact\_match=0.00)}
    &
    \includegraphics[width=\linewidth,height=3cm,keepaspectratio]{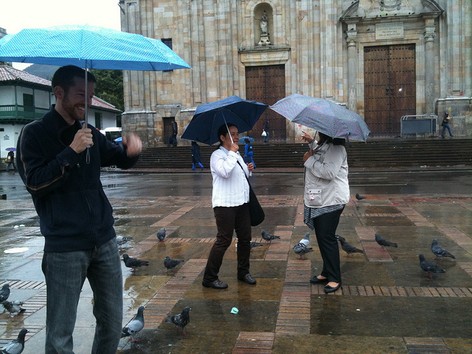}\par\vspace{2pt}
    \textbf{Q:} What is the name of the device that is protecting people from the rain in this picture?\newline
    \textbf{A:} umbrella\par\vspace{6pt}
    \includegraphics[width=\linewidth,height=3cm,keepaspectratio]{figures/samples_images/ok_vqa_val2014_115_real2}\par\vspace{2pt}
    \textbf{Q:} What is this on top of?\newline
    \textbf{A:} coffee shop\par\vspace{8pt}
    \textbf{RICES:} Umbrella\ \textcolor{green!60!black}{\ding{51}} {\scriptsize(exact\_match=1.00)}\par\vspace{3pt}
    \textbf{TTT-NN:} umbrella\ \textcolor{green!60!black}{\ding{51}} {\scriptsize(exact\_match=1.00)}
    &
    \includegraphics[width=\linewidth,height=3cm,keepaspectratio]{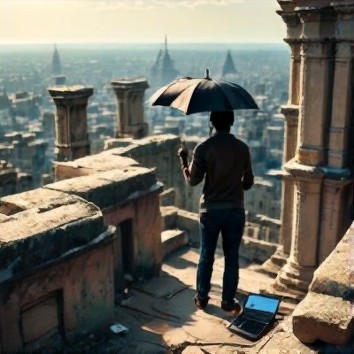}\par\vspace{2pt}
    \textbf{Q:} What is the man holding?\newline
    \textbf{A:} Umbrella\par\vspace{6pt}
    \includegraphics[width=\linewidth,height=3cm,keepaspectratio]{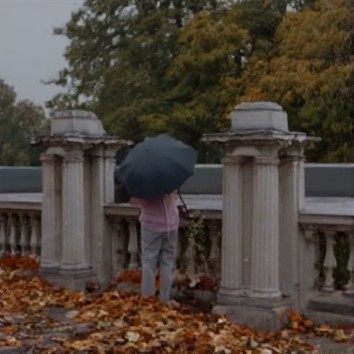}\par\vspace{2pt}
    \textbf{Q:} What season is it?\newline
    \textbf{A:} Fall/Winter\par\vspace{8pt}
    \textbf{SIMIT-ICL:} Umbrella\ \textcolor{green!60!black}{\ding{51}} {\scriptsize(exact\_match=1.00)}\par\vspace{3pt}
    \textbf{SIMIT-FT (SQ):} Umbrella\ \textcolor{green!60!black}{\ding{51}} {\scriptsize(exact\_match=1.00)}
    \\

    \midrule
    \includegraphics[width=\linewidth,height=4.2cm,keepaspectratio]{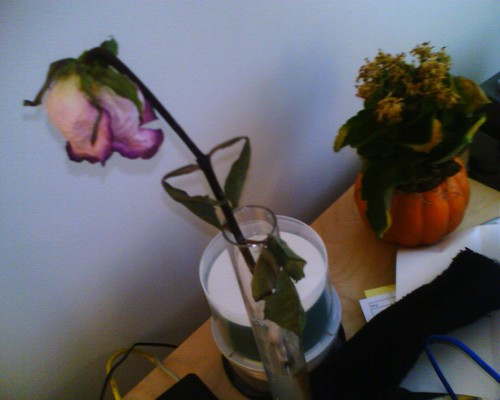}\par\vspace{4pt}
    \textbf{Prompt:} \textit{What type of flower is this? \newline
     When the provided information is insufficient, respond with `Unanswerable'. \newline
     Answer the question using a single word or phrase.}\par\vspace{4pt}
    \textbf{GT:} rose\par\vspace{4pt}
    \textbf{Pretrained:} Unanswerable\ \textcolor{red!70!black}{\ding{55}} {\scriptsize(exact\_match=0.00)}
    &
    \includegraphics[width=\linewidth,height=3cm,keepaspectratio]{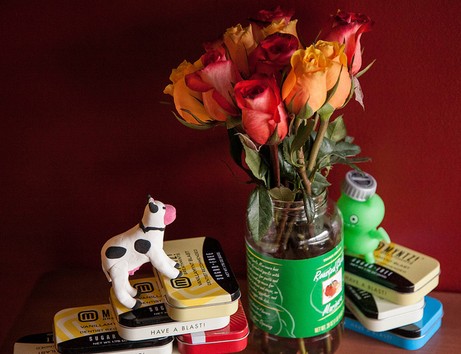}\par\vspace{2pt}
    \textbf{Q:} What type of flower is this?\newline
    \textbf{A:} rose\par\vspace{6pt}
    \includegraphics[width=\linewidth,height=3cm,keepaspectratio]{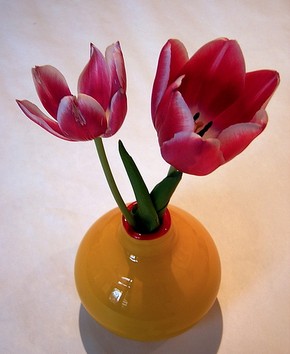}\par\vspace{2pt}
    \textbf{Q:} What kind of flowers are these?\newline
    \textbf{A:} tulip\par\vspace{8pt}
    \textbf{RICES:} rose\ \textcolor{green!60!black}{\ding{51}} {\scriptsize(exact\_match=1.00)}\par\vspace{3pt}
    \textbf{TTT-NN:} rose\ \textcolor{green!60!black}{\ding{51}} {\scriptsize(exact\_match=1.00)}
    &
    \includegraphics[width=\linewidth,height=3cm,keepaspectratio]{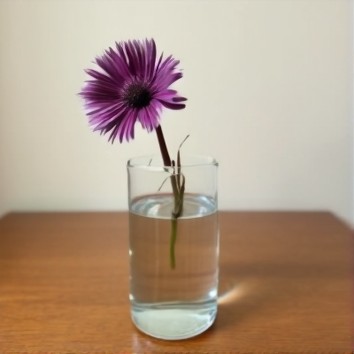}\par\vspace{2pt}
    \textbf{Q:} How many flowers are in the vase?\newline
    \textbf{A:} One\par\vspace{6pt}
    \includegraphics[width=\linewidth,height=3cm,keepaspectratio]{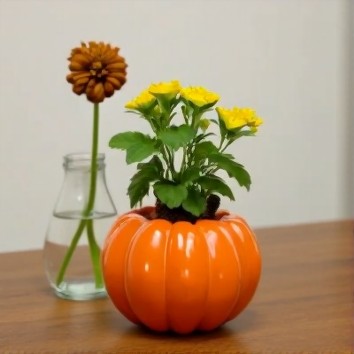}\par\vspace{2pt}
    \textbf{Q:} What color is the pumpkin-shaped container holding the plant?\newline
    \textbf{A:} Orange\par\vspace{8pt}
    \textbf{SIMIT-ICL:} Rose\ \textcolor{green!60!black}{\ding{51}} {\scriptsize(exact\_match=1.00)}\par\vspace{3pt}
    \textbf{SIMIT-FT (SQ):} Rose\ \textcolor{green!60!black}{\ding{51}} {\scriptsize(exact\_match=1.00)}
    \\
    \end{tabularx}
    \caption{Qualitative comparison of adaptation using retrieved real samples and self-generated synthetic samples. Both SIMIT-ICL and SIMIT-FT (SQ) improve on the pretrained model.}
    \label{fig:qual_ok_vqa_val2014_210}
\end{figure}

\begin{figure}[tbh]
    \centering
    \footnotesize
    \setlength{\tabcolsep}{4pt}
    \renewcommand{\arraystretch}{1.2}
    \renewcommand{\tabularxcolumn}[1]{m{#1}}
    \begin{tabularx}{\textwidth}{
        @{}
        >{\hsize=0.9375\hsize\linewidth=\hsize\raggedright\arraybackslash}X
        >{\hsize=1.03125\hsize\linewidth=\hsize\raggedright\arraybackslash}X
        >{\hsize=1.03125\hsize\linewidth=\hsize\raggedright\arraybackslash}X
        @{}
    }
    \textbf{Unlabeled test query} & \textbf{Retrieved real samples} & \textbf{Self-generated synthetic samples} \\
    \midrule
    \includegraphics[width=\linewidth,height=4.2cm,keepaspectratio]{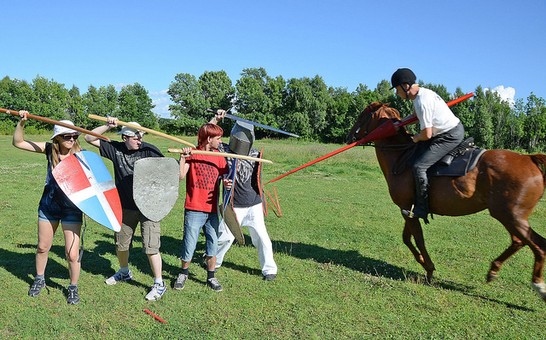}\par\vspace{4pt}
    \textbf{Prompt:} \textit{What time period is the action pictured here based off of? \newline
     When the provided information is insufficient, respond with `Unanswerable'. \newline
     Answer the question using a single word or phrase.}\par\vspace{4pt}
    \textbf{GT:} medieval, 1400's\par\vspace{4pt}
    \textbf{Pretrained:} Unanswerable\ \textcolor{red!70!black}{\ding{55}} {\scriptsize(exact\_match=0.00)}
    &
    \includegraphics[width=\linewidth,height=3cm,keepaspectratio]{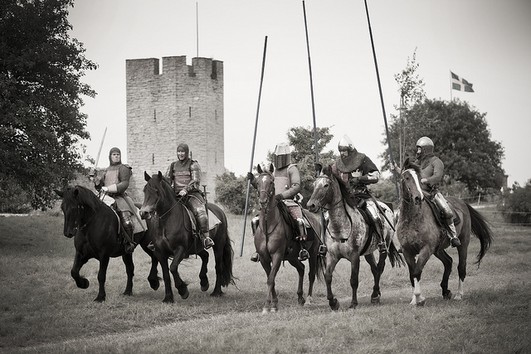}\par\vspace{2pt}
    \textbf{Q:} What time period are these people from?\newline
    \textbf{A:} medieval\par\vspace{6pt}
    \includegraphics[width=\linewidth,height=3cm,keepaspectratio]{figures/samples_images/ok_vqa_val2014_253_real2}\par\vspace{2pt}
    \textbf{Q:} What kind of event is this?\newline
    \textbf{A:} jousting\par\vspace{8pt}
    \textbf{RICES:} medieval\ \textcolor{green!60!black}{\ding{51}} {\scriptsize(exact\_match=1.00)}\par\vspace{3pt}
    \textbf{TTT-NN:} medieval\ \textcolor{green!60!black}{\ding{51}} {\scriptsize(exact\_match=1.00)}
    &
    \includegraphics[width=\linewidth,height=3cm,keepaspectratio]{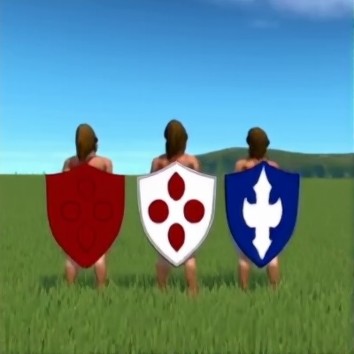}\par\vspace{2pt}
    \textbf{Q:} How many people are holding shields?\newline
    \textbf{A:} Three\par\vspace{6pt}
    \includegraphics[width=\linewidth,height=3cm,keepaspectratio]{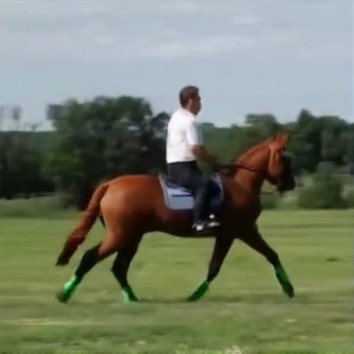}\par\vspace{2pt}
    \textbf{Q:} What is the person riding?\newline
    \textbf{A:} Horse\par\vspace{8pt}
    \textbf{SIMIT-ICL:} Medieval.\ \textcolor{green!60!black}{\ding{51}} {\scriptsize(exact\_match=1.00)}\par\vspace{3pt}
    \textbf{SIMIT-FT (SQ):} Medieval\ \textcolor{green!60!black}{\ding{51}} {\scriptsize(exact\_match=1.00)}
    \\

    \midrule
    \includegraphics[width=\linewidth,height=4.2cm,keepaspectratio]{figures/samples_images/ok_vqa_val2014_272_query}\par\vspace{4pt}
    \textbf{Prompt:} \textit{What toy is this? \newline
     When the provided information is insufficient, respond with `Unanswerable'. \newline
     Answer the question using a single word or phrase.}\par\vspace{4pt}
    \textbf{GT:} teddy bear\par\vspace{4pt}
    \textbf{Pretrained:} Unanswerable\ \textcolor{red!70!black}{\ding{55}} {\scriptsize(exact\_match=0.00)}
    &
    \includegraphics[width=\linewidth,height=3cm,keepaspectratio]{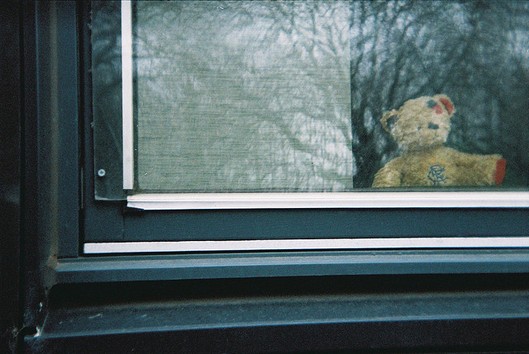}\par\vspace{2pt}
    \textbf{Q:} What kind of stuffed toy is in the picture?\newline
    \textbf{A:} bear\par\vspace{6pt}
    \includegraphics[width=\linewidth,height=3cm,keepaspectratio]{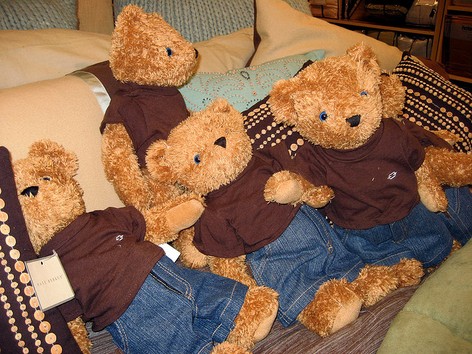}\par\vspace{2pt}
    \textbf{Q:} What animal is this?\newline
    \textbf{A:} bear\par\vspace{8pt}
    \textbf{RICES:} teddy bear\ \textcolor{green!60!black}{\ding{51}} {\scriptsize(exact\_match=1.00)}\par\vspace{3pt}
    \textbf{TTT-NN:} bear\ \textcolor{red!70!black}{\ding{55}} {\scriptsize(exact\_match=0.60)}
    &
    \includegraphics[width=\linewidth,height=3cm,keepaspectratio]{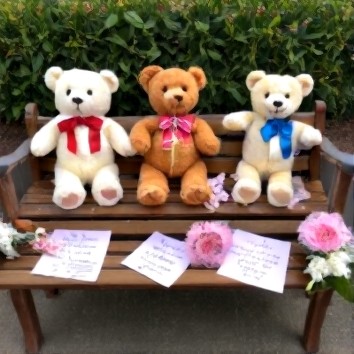}\par\vspace{2pt}
    \textbf{Q:} How many teddy bears are on the bench?\newline
    \textbf{A:} Three\par\vspace{6pt}
    \includegraphics[width=\linewidth,height=3cm,keepaspectratio]{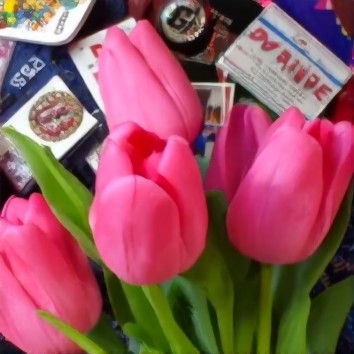}\par\vspace{2pt}
    \textbf{Q:} What color is the flower bouquet?\newline
    \textbf{A:} Pink\par\vspace{8pt}
    \textbf{SIMIT-ICL:} Teddy bear\ \textcolor{green!60!black}{\ding{51}} {\scriptsize(exact\_match=1.00)}\par\vspace{3pt}
    \textbf{SIMIT-FT (SQ):} Teddy bear\ \textcolor{green!60!black}{\ding{51}} {\scriptsize(exact\_match=1.00)}
    \\
    \end{tabularx}
    \caption{Qualitative comparison of adaptation using retrieved real samples and self-generated synthetic samples. Both SIMIT-ICL and SIMIT-FT (SQ) improve on the pretrained model.}
    \label{fig:qual_ok_vqa_val2014_272}
\end{figure}

\begin{figure}[tbh]
    \centering
    \footnotesize
    \setlength{\tabcolsep}{4pt}
    \renewcommand{\arraystretch}{1.2}
    \renewcommand{\tabularxcolumn}[1]{m{#1}}
    \begin{tabularx}{\textwidth}{
        @{}
        >{\hsize=0.9375\hsize\linewidth=\hsize\raggedright\arraybackslash}X
        >{\hsize=1.03125\hsize\linewidth=\hsize\raggedright\arraybackslash}X
        >{\hsize=1.03125\hsize\linewidth=\hsize\raggedright\arraybackslash}X
        @{}
    }
    \textbf{Unlabeled test query} & \textbf{Retrieved real samples} & \textbf{Self-generated synthetic samples} \\
    \midrule
    \includegraphics[width=\linewidth,height=4.2cm,keepaspectratio]{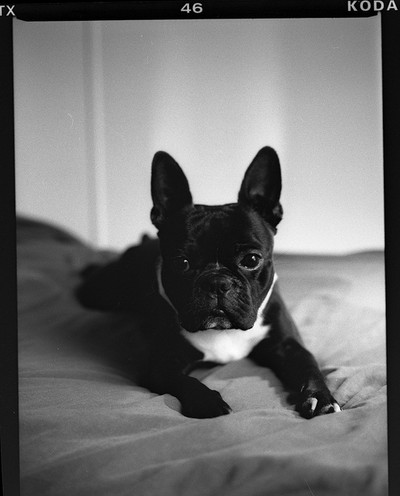}\par\vspace{4pt}
    \textbf{Prompt:} \textit{Can you guess the family or breed to which the dog belongs to? \newline
     When the provided information is insufficient, respond with `Unanswerable'. \newline
     Answer the question using a single word or phrase.}\par\vspace{4pt}
    \textbf{GT:} boston terrier, french bulldog\par\vspace{4pt}
    \textbf{Pretrained:} Unanswerable\ \textcolor{red!70!black}{\ding{55}} {\scriptsize(exact\_match=0.00)}
    &
    \includegraphics[width=\linewidth,height=3cm,keepaspectratio]{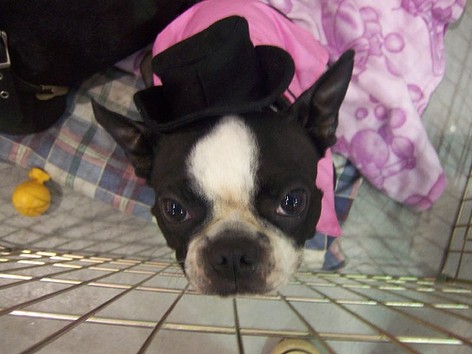}\par\vspace{2pt}
    \textbf{Q:} Name the family or bread to which this dog belongs to?\newline
    \textbf{A:} boston terrier\par\vspace{6pt}
    \includegraphics[width=\linewidth,height=3cm,keepaspectratio]{figures/samples_images/ok_vqa_val2014_291_real2}\par\vspace{2pt}
    \textbf{Q:} What breed of dog is seen in this picture? 
    \textbf{A:} bulldog\par\vspace{8pt}
    \textbf{RICES:} french bulldog\ \textcolor{green!60!black}{\ding{51}} {\scriptsize(exact\_match=1.00)}\par\vspace{3pt}
    \textbf{TTT-NN:} boston terrier\ \textcolor{green!60!black}{\ding{51}} {\scriptsize(exact\_match=1.00)}
    &
    \includegraphics[width=\linewidth,height=3cm,keepaspectratio]{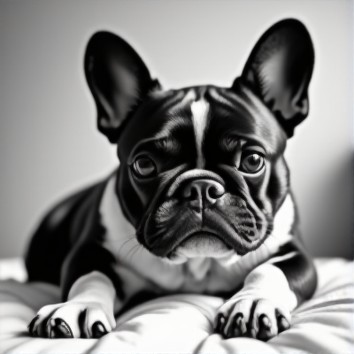}\par\vspace{2pt}
    \textbf{Q:} How many legs does the dog have?\newline
    \textbf{A:} Four\par\vspace{6pt}
    \includegraphics[width=\linewidth,height=3cm,keepaspectratio]{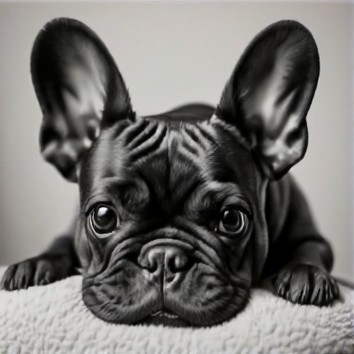}\par\vspace{2pt}
    \textbf{Q:} What kind of breed is the dog?\newline
    \textbf{A:} French Bulldog\par\vspace{8pt}
    \textbf{SIMIT-ICL:} French Bulldog\ \textcolor{green!60!black}{\ding{51}} {\scriptsize(exact\_match=1.00)}\par\vspace{3pt}
    \textbf{SIMIT-FT (SQ):} French bulldog\ \textcolor{green!60!black}{\ding{51}} {\scriptsize(exact\_match=1.00)}
    \\



    \midrule
    \includegraphics[width=\linewidth,height=4.2cm,keepaspectratio]{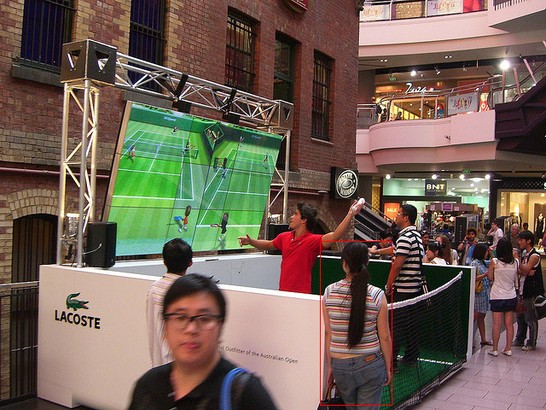}\par\vspace{4pt}
    \textbf{Prompt:} \textit{Provide a short description for this region.}\par\vspace{4pt}
    \textbf{GT:} woman in striped shirt, Girl with long hair in striped shirt, girl with long ponytail and striped shirt\par\vspace{4pt}
    \textbf{Pretrained:} A person is standing up.\ \textcolor{red!70!black}{\ding{55}} {\scriptsize(CIDEr=0.000)}
    &
    \includegraphics[width=\linewidth,height=3cm,keepaspectratio]{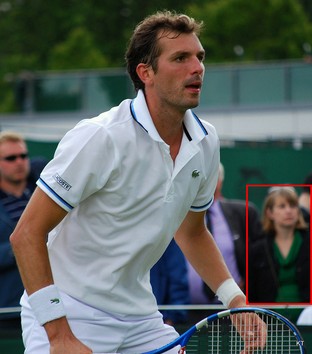}\par\vspace{2pt}
    \textbf{Q:} Please carefully observe the area circled in the image and come up with a caption for the area.\newline
    \textbf{A:} lady to right green shirt\par\vspace{6pt}
    \includegraphics[width=\linewidth,height=3cm,keepaspectratio]{figures/samples_images/refcoco_bbox_val_90_real2}\par\vspace{2pt}
    \textbf{Q:} Please carefully observe the area circled in the image and come up with a caption for the area.\newline
    \textbf{A:} man holding tennis racket\par\vspace{8pt}
    \textbf{RICES:} Girl in striped shirt.\ \textcolor{green!60!black}{\ding{51}} {\scriptsize(CIDEr=3.458)}\par\vspace{3pt}
    \textbf{TTT-NN:} lady with long ponytail\ \textcolor{green!60!black}{\ding{51}} {\scriptsize(CIDEr=1.618)}
    &
    \includegraphics[width=\linewidth,height=3cm,keepaspectratio]{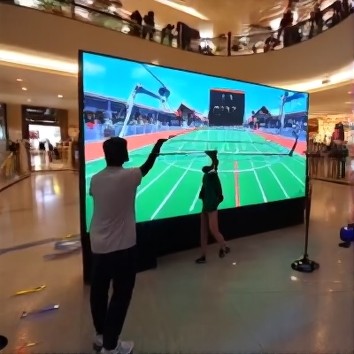}\par\vspace{2pt}
    \textbf{Q:} How many people are actively participating in the activity?\newline
    \textbf{A:} Two\par\vspace{6pt}
    \includegraphics[width=\linewidth,height=3cm,keepaspectratio]{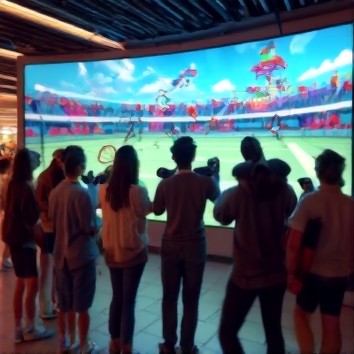}\par\vspace{2pt}
    \textbf{Q:} What type of activity is taking place in the image?\newline
    \textbf{A:} Virtual sports gaming\par\vspace{8pt}
    \textbf{SIMIT-ICL:} A woman with a long ponytail.\ \textcolor{green!60!black}{\ding{51}} {\scriptsize(CIDEr=0.938)}\par\vspace{3pt}
    \textbf{SIMIT-FT (SQ):} Girl with long ponytail.\ \textcolor{green!60!black}{\ding{51}} {\scriptsize(CIDEr=2.837)}
    \\
    \end{tabularx}
    \caption{Qualitative comparison of adaptation using retrieved real samples and self-generated synthetic samples. Both SIMIT-ICL and SIMIT-FT (SQ) improve on the pretrained model.}
    \label{fig:qual_refcoco_bbox_val_90}
\end{figure}

\begin{figure}[t!]
    \centering
    \footnotesize
    \setlength{\tabcolsep}{4pt}
    \renewcommand{\arraystretch}{1.2}
    \renewcommand{\tabularxcolumn}[1]{m{#1}}
    \begin{tabularx}{\textwidth}{
        @{}
        >{\hsize=0.9375\hsize\linewidth=\hsize\raggedright\arraybackslash}X
        >{\hsize=1.03125\hsize\linewidth=\hsize\raggedright\arraybackslash}X
        >{\hsize=1.03125\hsize\linewidth=\hsize\raggedright\arraybackslash}X
        @{}
    }
    \textbf{Unlabeled test query} & \textbf{Retrieved real samples} & \textbf{Self-generated synthetic samples} \\
    

    \midrule
    \includegraphics[width=\linewidth,height=4.2cm,keepaspectratio]{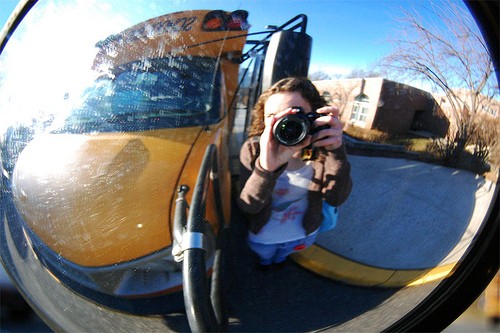}\par\vspace{4pt}
    \textbf{Prompt:} \textit{What kind of lens distorts the picture in this way? \newline
     When the provided information is insufficient, respond with `Unanswerable'. \newline
     Answer the question using a single word or phrase.}\par\vspace{4pt}
    \textbf{GT:} fisheye, fish eye lens\par\vspace{4pt}
    \textbf{Pretrained:} Unanswerable\ \textcolor{red!70!black}{\ding{55}} {\scriptsize(exact\_match=0.00)}
    &
    \includegraphics[width=\linewidth,height=3cm,keepaspectratio]{figures/samples_images/ok_vqa_val2014_22_real1}\par\vspace{2pt}
    \textbf{Q:} What type of lens is used for this picture?\newline
    \textbf{A:} fisheye\par\vspace{6pt}
    \includegraphics[width=\linewidth,height=3cm,keepaspectratio]{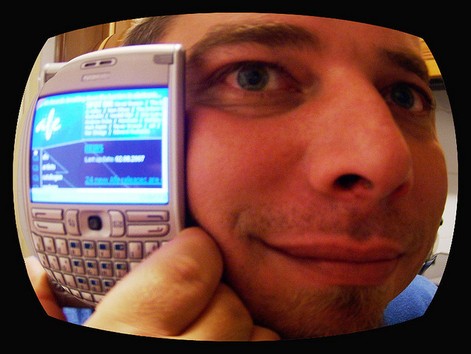}\par\vspace{2pt}
    \textbf{Q:} What is the term for the effect used in this photo?\newline
    \textbf{A:} fisheye\par\vspace{8pt}
    \textbf{RICES:} fisheye\ \textcolor{green!60!black}{\ding{51}} {\scriptsize(exact\_match=1.00)}\par\vspace{3pt}
    \textbf{TTT-NN:} fisheye\ \textcolor{green!60!black}{\ding{51}} {\scriptsize(exact\_match=1.00)}
    &
    \includegraphics[width=\linewidth,height=3cm,keepaspectratio]{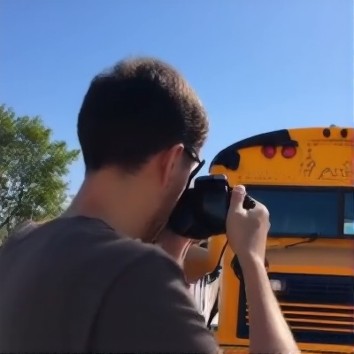}\par\vspace{2pt}
    \textbf{Q:} How many people are in the image? 
    \textbf{A:} One\par\vspace{6pt}
    \includegraphics[width=\linewidth,height=3cm,keepaspectratio]{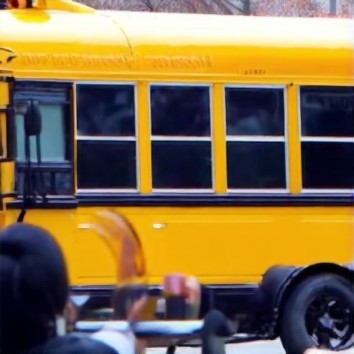}\par\vspace{2pt}
    \textbf{Q:} What color is the vehicle in the image? 
    \textbf{A:} Yellow\par\vspace{8pt}
    \textbf{SIMIT-ICL:} Fisheye.\ \textcolor{green!60!black}{\ding{51}} {\scriptsize(exact\_match=1.00)}\par\vspace{3pt}
    \textbf{SIMIT-FT (SQ):} Fisheye\ \textcolor{green!60!black}{\ding{51}} {\scriptsize(exact\_match=1.00)}
    \\
    \end{tabularx}
    \caption{Qualitative comparison of adaptation using retrieved real samples and self-generated synthetic samples. Both SIMIT-ICL and SIMIT-FT (SQ) improve on the pretrained model.}
    \label{fig:qual_refcoco_bbox_val_205}
\end{figure}

Figs.~\ref{fig:qual_ok_vqa_val2014_381}
--\ref{fig:qual_refcoco_bbox_val_205} present selected qualitative examples with ABA and DF disabled and a fixed sample budget of \(K=2\) for compact visualization. We compare adaptation using retrieved real samples (RICES and TTT-NN) with self-generated synthetic samples (SIMIT-ICL and SIMIT-FT (SQ)). These examples illustrate how self-generated samples can yield more accurate answers. 


The examples also suggest greater sensitivity to mismatched supervision under fine-tuning. In Fig. \ref{fig:qual_ok_vqa_val2014_22} (bottom), TTT-NN output adds the unnecessary attributes ``tan and white'', 
echoing ``tan'' in the first retrieved sample, whereas RICES answers ``dog.''
Similarly, in Fig. \ref{fig:qual_ok_vqa_val2014_477} (top), TTT-NN repeats the retrieved material label ``porcelain'' when asked to identify an art form.
These cases suggest that fine-tuning can sometimes absorb irrelevant details from retrieved supervision, while ICL can disregard them.

Synthetic supervision nevertheless remains imperfect. In Fig. \ref{fig:qual_ok_vqa_val2014_22} (bottom), the first synthetic answer labels the animal’s belly ``pink,'' although its belly is not clearly visible. This unsupported attribute differs from the target task of recognizing the animal, and both SIMIT variants still answer correctly. The example illustrates that a noisy synthetic label need not reinforce the target prediction error or prevent successful adaptation. Sec. \ref{sec:label_noise} examines this distinction quantitatively by measuring the alignment between supervision errors and the model’s existing failures.

\FloatBarrier

\FloatBarrier

\FloatBarrier

\FloatBarrier

  \section{Limitations and Future Works}
\label{sec:appendix_limitations}

\begin{itemize}[leftmargin=*]
    \item SIMIT-FT$\to$ICL just uses the SIMIT-FT (Test-Set) checkpoint in place of the pretrained baseline model in the SIMIT-ICL procedure. We apply this combination without additional hyperparameter tuning and any dedicated training on synthetic samples formatted for in-context learning. Both directions can improve the accuracy even more.
    \item SIMIT is orthogonal to LFRL methods, suggesting that combining both approaches could yield further gains in suitable domains. 
    \item SIMIT could also be used to augment real training data with similar synthetic samples. Our self-generated synthetic samples are tailored for the expected target distribution and the model's learning capacity. 
    CRAFT~\citep{ziegler2025craft} demonstrates that a small set of human-curated examples can guide the construction of a much larger task-specific synthetic dataset for finetuning.
    SEAL~\citep{zweiger2025selfadapting} also supports this paradigm by providing a human-learning analogy: students rewrite raw material from lectures, textbooks, and online sources into their own notes to internalize the content and better answer exam questions.
    \item SIMIT performs query-conditioned, \textit{transductive} test-time adaptation through constructive self-supervision, producing local-optimal models specialized for a given target test set. Such specialization can reduce performance on other benchmarks, as illustrated in Fig.~\ref{fig:cross_dataset_full}. Transferring these capabilities to a generalist model could therefore benefit from multi-teacher on-policy distillation across specialists~\citep{ma2026mopd}, a strategy popularized by DeepSeek-V4~\citep{xu2026deepseek}.
    \item Future work could explore jointly updating the solver and synthesizer. We keep the synthesizer fixed to preserve broad applicability and enable rigorous controlled ablations of the proposed design choices. In particular, SIMIT-ICL requires no training and can improve fixed models, including closed-source proprietary models accessible only through an API. 
\end{itemize}


\end{document}